\documentclass[a4paper]{article} % 一般的なスタイルの書き方
\usepackage[margin=1in]{geometry}

\usepackage{microtype}
\usepackage{graphicx}
\usepackage{subcaption}
\usepackage{booktabs} % for professional tables
\usepackage{here}
\usepackage{hyperref}

\usepackage{amsmath,amsthm,amsfonts,amssymb}
\usepackage{mathtools}
\usepackage{xcolor}
\usepackage{natbib}
\usepackage{algorithm} % for algorithm
\usepackage{algpseudocode} % for algorithm
\usepackage{url}
\usepackage{multirow}

\usepackage[T1]{fontenc}
\usepackage[utf8]{inputenc}
\usepackage{bbm}

\def\*#1{\boldsymbol{#1}}

\theoremstyle{plain}

\newtheorem{theorem}{Theorem}[section]
\newtheorem{proposition}[theorem]{Proposition}
\newtheorem{lemma}[theorem]{Lemma}
\newtheorem{corollary}[theorem]{Corollary}
\theoremstyle{definition}
\newtheorem{definition}[theorem]{Definition}
\newtheorem{assumption}[theorem]{Assumption}
\theoremstyle{remark}

\DeclareMathOperator*{\argmin}{arg\,min}
\DeclareMathOperator*{\argmax}{arg\,max}

\newcommand{\lw}[1]{\smash{\lower2.ex\hbox{#1}}}

\newcommand{\RR}{\mathbb{R}}
\newcommand{\NN}{\mathbb{N}}

\newcommand{\EE}{\mathbb{E}}

\newcommand{\cD}{{\cal D}}

\newcommand{\cF}{{\cal F}}
\newcommand{\cG}{{\cal G}}
\newcommand{\cH}{{\cal H}}

\newcommand{\cN}{{\cal N}}
\newcommand{\cO}{{\cal O}}
\newcommand{\cP}{{\cal P}}

\newcommand{\cX}{{\cal X}}

\mathtoolsset{showonlyrefs}
\usepackage{authblk}

\title{Distributionally Robust Active Learning \\for Gaussian Process Regression}
\date{}

\author[1,2]{Shion Takeno}

\author[1]{Yoshito Okura}
\author[3]{Yu Inatsu}
\author[1]{Tatsuya Aoyama}
\author[1]{Tomonari Tanaka}
\author[1]{Satoshi Akahane}
\author[2]{Hiroyuki Hanada}
\author[2]{Noriaki Hashimoto}
\author[4]{Taro Murayama}
\author[4]{Hanju Lee}
\author[4]{Shinya Kojima}
\author[1,2]{Ichiro Takeuchi}
\affil[1]{Nagoya University}
\affil[2]{RIKEN AIP}
\affil[3]{Nagoya Institute of Technology}
\affil[4]{DENSO CORPORATION}

\affil[ ]{\texttt{{takeno.s.mllab.nit@gmail.com}}}
\affil[ ]{\texttt{{takeuchi.ichiro.n6@f.mail.nagoya-u.ac.jp}}}

\begin{document}
\maketitle

\begin{abstract}
     Gaussian process regression (GPR) or kernel ridge regression is a widely used and powerful tool for nonlinear prediction.
     Therefore, active learning (AL) for GPR, which actively collects data labels to achieve an accurate prediction with fewer data labels, is an important problem.
     However, existing AL methods do not theoretically guarantee prediction accuracy for target distribution.
     Furthermore, as discussed in the distributionally robust learning literature, specifying the target distribution is often difficult.
     Thus, this paper proposes two AL methods that effectively reduce the worst-case expected error for GPR, which is the worst-case expectation in target distribution candidates.
     We show an upper bound of the worst-case expected squared error, which suggests that the error will be arbitrarily small by a finite number of data labels under mild conditions.
     Finally, we demonstrate the effectiveness of the proposed methods through synthetic and real-world datasets.
\end{abstract}

\section{Introduction}
\label{sec:intro}

% active learning
Active learning (AL) \citep{settles2009-active} is a framework for achieving high prediction performance with fewer data when labeling new data is expensive. 
For this purpose, AL algorithms actively acquire the label of data that improves the prediction performance of some statistical model based on \emph{acquisition functions} (AFs).
Many types of AFs have been proposed, such as uncertainty sampling (US), random sampling (RS), variance reduction, and information gain, as summarized in \citep{settles2009-active}.

% Gaussian process-based AL
Gaussian process regression (GPR) model \citep{Rasmussen2005-Gaussian} is often used as a base statistical model for AL algorithms due to its flexible prediction capability \citep{Seo2000gaussian,yu2006active,Guestrin2005-near,krause2008-near,hoang2014nonmyopic,hubotter2024-transductive}.
Standard AL methods for the GPR are based on information gain~\citep{Guestrin2005-near,krause2008-near,kirsch2021test,kirsch2022unifying,bickford2023-prediction,hubotter2024-transductive}.
Most information gain-based approaches are heuristics without theoretical guarantees.
A notable exception is the work by \citep{Guestrin2005-near,krause2008-near}, which shows that the US for the GPR model is optimal to maximize the information gain from the obtained data labels regarding the GP prior.
Furthermore, from the analysis of kernelized bandits~\citep[e.g., ][]{Srinivas2010-Gaussian,salgia2024random}, we can see that the US and RS guarantee the convergence of the maximum of posterior variance (See Proposition~\ref{prop:us_rs} for details).
Another commonly used AF is variance reduction~\citep{Seo2000gaussian,yu2006active,Shoham2023experimental,hubotter2024-transductive}, which can be computed efficiently in the GPR.
However, these AFs do not incorporate the importance of the unlabelled dataset, that is, the prior information regarding the target distribution.
In addition, to our knowledge, except for the worst-case analysis in Proposition~\ref{prop:us_rs}, there are no theoretical guarantees for the target prediction error.

% Target distribution-aware AL
Several studies have tackled the development of the target distribution-aware AL~\citep{kirsch2021test,kirsch2022unifying,bickford2023-prediction}.
In particular, as an extension of the distributionally robust learning~\citep{chen2020distributionally}, \citet{frogner2021incorporating} proposed \emph{distributionally robust AL} (DRAL), which aims to minimize the worst-case error in the set of target distributions to obtain a robust model.
However, since these studies employed the heuristic AL methods based on, e.g., information gain and expected model change~\citep{settles2009-active}, the theoretical guarantee has not been shown.

This paper develops a DRAL framework for the GPR model.
We aim to minimize the worst-case expected error, where the worst-case scenario and the expectation are taken regarding the target distribution candidates and chosen target distributions, respectively.
Note that our formulation generalizes target distribution-aware AL since it includes the case in which the unique target distribution can be specified.
We perform the theoretical analysis under two conditions called Bayesian and frequentist assumptions~\citep{Srinivas2010-Gaussian}, in which we leverage several useful lemmas in kernelized bandit literature~\citep{Srinivas2010-Gaussian,vakili2021-optimal,vakili2021-information,Kusakawa2022-bayesian}.
%
% First, we provide several common properties of this worst-case expected error in the GPR model.
% %
% Based on these properties, we propose simple stochastic and deterministic algorithms.
% %
% We derive the upper bound of the worst-case expected error of the proposed methods, which suggests that, under mild conditions, the error can be arbitrarily small by a finite number of data labels.

Our contributions are summarized as follows:
\begin{enumerate}
    \item We show several properties of the worst-case squared error for the GPR model, which suggests that the error can be bounded from above using the posterior variance even if the input domain is continuous. Along the way to proving the error properties, we show the Lipschitz constant of the posterior mean of GPs in Lemmas~\ref{lem:RKHS_lipschitz} and \ref{lem:bayesian_lipschitz_posterior_mean}, which may be of independent interest.
    \item We propose two DRAL methods for the GPR model, inspired by the RS and the greedy algorithm. Our proposed methods are designed to guarantee the convergence of the (expected) posterior variance.
    \item We show the probabilistic upper bounds of the error incurred by the proposed algorithm, which suggests that under mild conditions, the error can be arbitrarily small by a finite number of data labels.
\end{enumerate}
Finally, we demonstrate the effectiveness of the proposed methods via synthetic and real-world regression problems.

\section{Background}
\label{sec:background}

This section provides the known properties of the GPR.

%%%%%%%%%%%%%%%%%%%%%%%%%%%%%%%%%%%%%%%%%%%%%%%%%%%%%%%%%%%%%%%%%%%%%%%%%%%%
\subsection{GPR model}

The GPR model \citep{Rasmussen2005-Gaussian} is a kernel-based regression model.
Let us consider that we have already obtained the training dataset of input-output pair $\cD_t = \{ (\*x_i, y_{\*x_i}) \}_{i=1}^t$, where $\forall i, \*x_i \in \cX \subset \RR^d$, $y_{\*x_i} \in \RR$, and $d$ is an input dimension.
The GPR model assumes that, without loss of generality, $f$ follows zero-mean GP, that is, $f \sim \cG \cP (0, k)$, where $k: \cX \times \cX \mapsto \RR$ is a predefined positive semidefinite kernel function.
In addition, the $i$-th observation $y_{\*x_i}$ is assumed to be contaminated by i.i.d. Gaussian noise $\epsilon_i \sim \cN(0, \sigma^2)$ as $y_{\*x_i} = f(\*x_i) + \epsilon_i$.
Then, the posterior distribution of $f$ becomes again a GP, whose mean and variance are analytically derived as follows:
\begin{equation}
    \begin{split}
        \mu_{t}(\*x) &= \*k_{t}(\*x)^\top \bigl(\*K + \sigma^2 \*I_{t} \bigr)^{-1} \*y_{t}, \\
        \sigma_{t}^2 (\*x) &= k(\*x, \*x) - \*k_{t}(\*x) ^\top \bigl(\*K + \sigma^2 \*I_{t} \bigr)^{-1} \*k_{t}(\*x),
    \end{split}
    \label{eq:GP}
\end{equation}
where $\*k_{t}(\*x) \coloneqq \bigl( k(\*x, \*x_1), \dots, k(\*x, \*x_{t}) \bigr)^\top \in \RR^{t}$, $\*K \in \RR^{t \times t}$ is the kernel matrix whose $(i, j)$-element is $k(\*x_i, \*x_j)$, $\*I_{t} \in \RR^{t \times t}$ is the identity matrix, and $\*y_{t} \coloneqq (y_{\*x_1}, \dots, y_{\*x_t})^\top \in \RR^{t}$.
Finally, for later use, let the posterior variance at $\*x^*$ when $\*x_t = \*x$ be $\sigma_{t}^2 (\*x^* \mid \*x) \coloneqq \sigma_{t}^2 (\*x^* \mid \*x_t = \*x)$.
Note that the posterior variance calculation does not require $\*y_t$.
Furthermore, it is known that $\mu_t(\*x)$ is equivalent to the kernel ridge regression estimator with regularization parameter $\lambda = \sigma^2$~\citep{kanagawa2018gaussian}.

\paragraph{Maximum Information Gain (MIG):}
Further, we define MIG \citep{Srinivas2010-Gaussian,vakili2021-information}:
\begin{definition}[Maximum information gain]
    Let $f \sim \cG \cP (0, k)$ over $\cX \subset [0, r]^d$.
    Let $A = \{ \*a_i \}_{i=1}^T \subset \cX$.
    Let $\*f_A = \bigl(f(\*a_i) \bigr)_{i=1}^T$, $\*\epsilon_A = \bigl(\epsilon_i \bigr)_{i=1}^T$, where $\forall i, \epsilon_i \sim \cN(0, \sigma^2)$, and $\*y_A = \*f_A + \*\epsilon_A \in \RR^T$.
    Then, MIG $\gamma_T$ is defined as follows:
    \begin{align*}
        \gamma_T \coloneqq \max_{A \subset \cX; |A| = T} I(\*y_A ; \*f_A),
    \end{align*}
    where $I$ is the Shannon mutual information.
    \label{def:MIG}
\end{definition}
\noindent
It is known that MIG is sublinear for commonly used kernel functions, for example, $\gamma_T = \cO\bigl( d \log T \bigr)$ for linear kernels, $\gamma_T = \cO\bigl( (\log T)^{d+1} \bigr)$ for squared exponential (SE) kernels $k_{\rm SE} (\*x, \*x^\prime) = \exp\left( - \| \*x - \*x^\prime \|_2^2 / (2 \ell^2) \right)$, and $\gamma_T = \cO\bigl( T^{\frac{d}{2\nu + d}} (\log T)^{\frac{2\nu}{2\nu + d}} \bigr)$ for Mat\'{e}rn-$\nu$ kernels $k_{\rm Mat} = \frac{2^{1 - \nu}}{\Gamma(\nu)} \left( \frac{\sqrt{2\nu} \| \*x - \*x^\prime \|_2 }{\ell} \right)^{\nu} J_{\nu} \left( \frac{\sqrt{2\nu} \| \*x - \*x^\prime \|_2 }{\ell} \right) $, where $\ell, \nu > 0$ are the lengthscale and smoothness parameter, respectively, and $\Gamma(\cdot)$ and $J_{\nu}$ are Gamma and modified Bessel functions, respectively \citep{Srinivas2010-Gaussian,vakili2021-information}.

\paragraph{Lipschitz Consatant of $\sigma_t(\*x)$:}
We will use the following useful result from Theorem~E.4 in~\citep{Kusakawa2022-bayesian}:
\begin{lemma}[Lipschitz constant for posterior standard deviation]
    Let $k(\*x, \*x^\prime): \RR^d \times \RR^d \to \RR$ be linear, SE, or Mat\'{e}rn-$\nu$ kernel and $k(\*x, \*x) \leq 1$. 
    Moreover, assume that a noise variance $\sigma^2$ is positive.
    Then, for any $t \geq 1$ and $\cD_{t}$, the posterior standard deviation $\sigma_{t} (\*x )$ satisfies that 
    \begin{align*}
        \forall \*x,\*x^\prime \in \RR^d, \ | \sigma_{t} (\*x ) - \sigma_{t} (\*x^\prime ) | \leq L_{\sigma} \| \*x - \*x^\prime \| _1,
    % \label{eq:ineq_sigma_Lip}
    \end{align*}
    where $L_{\sigma}$ is a positive constant given by 
    \begin{align*}
        L_{\sigma} = \left\{
        \begin{array}{ll}
            1 & \text{if $k(\*x,\*x^\prime)$ is the linear kernel}, \\
            \frac{\sqrt{2}}{\ell} & \text{if $k(\*x,\*x^\prime)$ is the SE kernel}, \\
            \frac{\sqrt{2}}{\ell}
            \sqrt{\frac{\nu}{\nu-1} } & \text{if $k(\*x,\*x^\prime)$ is the Mat\'{e}rn kernel},
        \end{array}\right.
    \end{align*}
    where $\nu > 1$.
    % where $\ell$ is length scale parameters in Gaussian and Mat\'{e}rn kernels, and $\nu > 1$ is a degree of freedom with $\nu > 1$.
    \label{lem:Lipschitz_posterior_var}
\end{lemma}
% \begin{proof}
%     We can see that
%     \begin{align*}
%         \frac{\partial \sigma^2_t(\*u)}{\partial u_j} \bigg|_{\*u = \*x} =  2 \sigma_t(\*x) \frac{\partial \sigma_t(\*u)}{\partial u_j} \bigg|_{\*u = \*x} \leq 2 \frac{\partial \sigma_t(\*u)}{\partial u_j} \bigg|_{\*u = \*x}.
%     \end{align*}
%     %
%     From Theorem~E.4 in~\citet{Kusakawa2022-bayesian}, we see that $2\frac{\partial \sigma_t(\*u)}{\partial u_j} \bigg|_{\*u = \*x} \leq L_{\sigma_t^2}$ for all $\*x \in \cX$ and $j \in [d]$.
% \end{proof}

%%%%%%%%%%%%%%%%%%%%%%%%%%%%%%%%%%%%%%%%%%%%%%%%%%%%%%%%%%%%%%%%%%%%%%%%%%%%
\subsection{Uncertainty Sampling and Random Sampling}
% The US and RS are common AL methods.
%
For the GPR model, the US selects the most uncertain input as $t$-th input:
\begin{align*}
    \*x_t = \argmax_{\*x \in \cX} \sigma^2_{t-1}(\*x).
\end{align*}
The RS randomly selects a $t$-th input by a fixed probability distribution $p(\*x)$ over $\cX$:
\begin{align*}
    \*x_t \sim p(\*x).
\end{align*}
For both algorithms, the upper bound of the maximum variance is known:
\begin{proposition}
    Assume $\cX$ is a compact subset of $\RR^d$.
    If we run the US, the following inequality holds:
    \begin{align*}
        \max_{\*x \in \cX} \sigma^2_T (\*x) \leq \frac{C_1 \gamma_T}{T},
    \end{align*}
    where $C_1 = 2 / \log(1 + \sigma^{-2})$.
    Furthermore, if we run the RS, the following inequality holds with probability at least $1 - \delta$, where $\delta \in (0, 1)$, under several conditions:
    \begin{align*}
        \max_{\*x \in \cX} \sigma^2_T (\*x) = \cO \left(\frac{\sigma^2\gamma_T}{T}\right).
    \end{align*}
    \label{prop:us_rs}
\end{proposition}
\begin{proof}
    For the US, see, e.g., Eq.~(16) in \citep{vakili2021-optimal} and Lemma~5.4 in \citep{Srinivas2010-Gaussian}.
    For the RS, see Theorem~3.1 in \citep{salgia2024random}.
\end{proof}
When $\gamma_T$ is sublinear, the above upper bounds suggest that the maximum variance will be arbitrarily small within the finite time horizons.

%%%%%%%%%%%%%%%%%%%%%%%%%%%%%%%%%%%%%%%%%%%%%%%%%%%%%%%%%%%%%%%%%%%%%%%%%%%%
\subsection{Regularity Assumptions and Predictive Guarantees}

% We consider two conditions called Bayesian and frequentist assumptions.
%
Here, we provide the details of Bayesian and frequentist assumptions and predictive guarantees for both assumptions.

% \paragraph{Bayesian Assumption and Credible Interval}
\subsubsection{Bayesian Assumption}

We consider the following assumption:
\begin{assumption}
    The function $f$ is a sample path $f \sim \cG \cP (0, k)$ and the $i$-th observation $y_{\*x_i}$ is contaminated by i.i.d. Gaussian noise $\epsilon_i \sim \cN(0, \sigma^2)$ as $y_{\*x_i} = f(\*x_i) + \epsilon_i$.
    In addition, the kernel function is normalized as $k(\*x, \*x^\prime) \leq 1$ for all $\*x, \*x^\prime \in \cX$.
    \label{assump:Bayesian}
\end{assumption}
Furthermore, for continuous $\cX$, we assume the following smoothness condition:
\begin{assumption}
    Let $\cX \subset [0, r]^d$ be a compact set, where $r > 0$.
    Assume that the kernel $k$ satisfies the following condition on the derivatives of a sample path $f$.
    There exist the constants $a, b > 0$ such that,
    \begin{align*}
        \Pr \left( \sup_{\*x \in \cX} \left| \frac{\partial f(\*u)}{\partial u_j} \Big|_{\*u=\*x} \right| > L \right) \leq a \exp \left( - \frac{L^2}{b^2} \right),
    \end{align*}
    for all $j \in [d]$.
    \label{assump:Bayesian_continuous}
\end{assumption}
This assumption holds for stationary and four times differentiable kernels \citep[Theorem~5 of][]{Ghosal2006-posterior}, such as SE kernel and Mat\'{e}rn-$\nu$ kernels with $\nu > 2$~\citep[Section~4 of][]{Srinivas2010-Gaussian}.
These assumptions are commonly used \citep{Srinivas2010-Gaussian,Kandasamy2018-Parallelised,paria2020-flexible,Takeno2023-randomized,takeno2024-posterior}.

As with Lemma~5.1 in \citep{Srinivas2010-Gaussian}, the credible interval can be obtained as follows:
\begin{lemma}
    Suppose that $\cX$ is finite and Assumption~\ref{assump:Bayesian} holds.
    Pick $\delta \in (0, 1)$ and $t \in \NN$.
    Then, for any given $\cD_{t}$,
    \begin{align*}
        \Pr \left( |f(\*x) - \mu_{t}(\*x) | \leq \beta^{1/2}_{\delta} \sigma_{t}(\*x), \forall \*x \in \cX \mid \cD_{t} \right)
        \geq 1 - \delta,
    \end{align*}
    where $\beta_{\delta} = 2 \log (|\cX| / \delta)$.
    \label{lem:bound_srinivas}
\end{lemma}

% \paragraph{Frequentist Assumption and Confidence Interval}
\subsubsection{Frequentist Assumption}

We assume that $f$ is an element of the reproducing kernel Hilbert space (RKHS) specified by the kernel $k$ as with \citep{Srinivas2010-Gaussian,Chowdhury2017-on,vakili2021-optimal,vakili2022improved,li2022gaussian}:
\begin{assumption}
    Let $f$ be an element of RKHS $\cH_k$ specified by the kernel $k$ used in the GPR model.
    Furthermore, the RKHS norm of $f$ is bounded as $\| f \|_{\cH_k} \leq B < \infty$ for some $B > 0$, where $\| \cdot \|_{\cH_k}$ denotes the RKHS norm of $\cH_k$.
    In addition, the $i$-th observation $y_{\*x_i}$ is contaminated by independent sub-Gaussian noises $\{ \epsilon_i \}_{i \in \NN}$ as $y_{\*x_i} = f(\*x_i) + \epsilon_i$.
    That is, for all $i \in \NN$, for all $\eta \in \RR$, and for some $R > 0$, the moment generating function of $\epsilon_i$ satisfies $\EE [\exp (\eta \epsilon_i)] \leq \exp\bigl( \frac{\eta^2 R^2}{2} \bigr) $.
    Finally, the kernel function is normalized as $k(\*x, \*x^\prime) \leq 1$ for all $\*x, \*x^\prime \in \cX$.
    \label{assump:frequentist}
\end{assumption}

Furthermore, for continuous $\cX$, we assume the following smoothness condition as with \citep{Chowdhury2017-on,vakili2021-optimal,vakili2022improved}:
\begin{assumption}
    The kernel function $k$ satisfies the following condition on the derivatives.
    There exists a constant $L_k$ such that,
    \begin{align*}
        \sup_{\*x \in \cX} \sup_{j \in [d]} \left| \frac{\partial^2 k(\*u, \*v)}{\partial u_j \partial v_j}\bigg|_{\*u=\*v=\*x} \right|^{1/2} \leq L_k.
    \end{align*}
    \label{assump:frequentist_continuous}
\end{assumption}
This assumption provides the Lipschitz constant of $f$:
\begin{lemma}[Lemma~5.1 in \citep{freitas2012exponential}]
    Suppose that Assumption~\ref{assump:frequentist_continuous} holds.
    Then, any $g \in \cH_k$ is Lipschitz continuous with respect to $\|g\|_{\cH_k} L_k$.
    \label{lem:RKHS_lipschitz}
\end{lemma}

We rely on the confidence bounds for non-adaptive sampling methods, which is a direct consequence of Theorem~1 in \citep{vakili2021-optimal} and the union bound:
\begin{lemma}
    Suppose that $\cX$ is finite and Assumption~\ref{assump:frequentist} holds.
    Pick $\delta \in (0, 1)$ and $t \in \NN$.
    Assume that $(\*x_i)_{i \in [t]}$ is independent of $(\epsilon_i)_{i \in [t]}$.
    Then, the following holds:
    \begin{align*}
        \Pr \left( |f(\*x) - \mu_{t}(\*x) | \leq \beta^{1/2}_{\delta} \sigma_{t}(\*x), \forall \*x \in \cX \right)
        \geq 1 - \delta,
    \end{align*}
    where $\beta_{\delta} = \left( B + \frac{R}{\sigma} \sqrt{2 \log ( \frac{2 |\cX|}{\delta})} \right)^2$.
    \label{lem:bound_vakili}
\end{lemma}

\section{Problem Statement and Its Property}
\label{sec:problem}

This section provides details on our problem setup and its properties.

%%%%%%%%%%%%%%%%%%%%%%%%%%%%%%%%%%%%%%%%%%%%%%%%%%%%%%%%%%%%%%%%%%%%%%%%%%%%
\subsection{Problem Statement}

We aim to minimize the worst-case expected errors regarding the GP prediction $\mu_T (\*x)$ after $T$-th function evaluations:
\begin{align}
    % {\rm DRAE}_T &\coloneqq \max_{p \in \cP} \EE_{p(\*x)} \left[ | f(\*x) - \mu_T(\*x) | \right] \\
    E_T &\coloneqq \max_{p \in \cP} \EE_{p(\*x^{*})} \left[ ( f(\*x^{*}) - \mu_T(\*x^{*}) )^2 \right],
    \label{eq:target_error}
\end{align}
where $\cP$ is a set of target distributions over the input space $\cX$ called ambiguity set~\citep{chen2020distributionally}.
We assume that $\max_{p \in \cP} \EE_{p(\*x^*)} \left[ g(\*x^*) \right]$ exists for any continuous function $g: \cX \rightarrow \RR$.
This paper concentrates on the setting where the training input space from which we can obtain labels includes the test input space.

Our problem setup can be seen as the generalization of the target distribution-aware AL and the AL for the worst-case error $\max_{\*x \in \cX} ( f(\*x) - \mu_T(\*x) )^2$.
This is because our problem is equivalent to the target distribution-aware AL if we set $|\cP| = 1$ and to the worst-case error minimization if $\cP$ includes $\{p \in \cP_{\cX} \mid \exists \*x \in \cX, p(\*x) = 1 \}$, where $\cP_{\rm \cX}$ is the set of the distributions over $\cX$.

%%%%%%%%%%%%%%%%%%%%%%%%%%%%%%%%%%%%%%%%%%%%%%%%%%%%%%%%%%%%%%%%%%%%%%%%%%%%
\subsection{High Probability Bound of Error}

% First, we provide the upper bound by the posterior variance.
%
If the input space $\cX$ is finite, we can obtain the upper bound of Eq.~\eqref{eq:target_error} as the direct consequence of Lemmas~\ref{lem:bound_srinivas} and \ref{lem:bound_vakili}:
\begin{lemma}
    Fix $\delta \in (0, 1)$ and $T \in \NN$.
    Suppose that Assumption~\ref{assump:Bayesian} holds and $\beta_\delta$ is set as in Lemma~\ref{lem:bound_srinivas}, or Assumption~\ref{assump:frequentist} holds and $\beta_\delta$ is set as in Lemma~\ref{lem:bound_vakili}.
    Then, the following holds with probability at least $1 - \delta$:
    \begin{align*}
        E_T &\leq \beta_{\delta} \max_{p \in \cP} \EE_{p(\*x^{*})}\left[ \sigma^2_{T}(\*x^{*}) \right].
    \end{align*}
    \label{lem:UB_error_discrete}
\end{lemma}

% Next, let us consider the case that $\cX = [0, r]^d$.
%
For continuous $\cX$, the confidence parameter $\beta_\delta \propto \log |\cX|$ diverges if we apply Lemmas~\ref{lem:bound_srinivas} and \ref{lem:bound_vakili} directly.
Therefore, in this case, the Lipschitz property is often leveraged~\citep{Chowdhury2017-on,vakili2021-optimal}.
The Lipschitz constant of $f$ can be directly derived from the Assumption~\ref{assump:Bayesian_continuous}, or Assumption~\ref{assump:frequentist_continuous} and Lemma~\ref{lem:RKHS_lipschitz}~\citep{Srinivas2010-Gaussian,freitas2012exponential}.

Furthermore, we need the Lipschitz constant of $\mu_T$.
In the frequentist setting, the Lipschitz constant for $\mu_T$ can be derived as $\cO(L_k \sqrt{t \log t})$ by Lemma~4 in \citep{vakili2021-optimal} and Lemma~\ref{lem:RKHS_lipschitz}.
To obtain a slightly tighter upper bound, we show the following lemma:
% \begin{lemma}[Modified from Lemma~F.1 of \citet{vakili2022improved}]
%     Fix $\delta \in (0, 1)$ and $t \in [T]$.
%     %
%     Suppose that Assumptions~\ref{assump:frequentist} and ~\ref{assump:frequentist_continuous} hold.
%     %
%     Then, the RKHS norm of $\mu_t(\cdot)$ satisfies the following with probability at least $1 - \delta$:
%     \begin{align*}
%         \| \mu_t \|_{\cH_k} \leq B + \frac{R}{\sigma} \sqrt{ 2t \log \left( \frac{2t}{\delta} \right)}.
%     \end{align*}
%     %
%     Thus, $\mu_T$ is $L_k \bigl( B + \frac{R}{\sigma} \sqrt{ 2t \log \left( 2t / \delta \right)} \bigr)$ Lipschitz continuous.
%     \label{lem:RKHS_norm_posterior_mean}
% \end{lemma}
\begin{lemma}
    Fix $\delta \in (0, 1)$ and $t \in [T]$.
    Suppose that Assumptions~\ref{assump:frequentist} and ~\ref{assump:frequentist_continuous} hold.
    Then, $\mu_t(\cdot)$ is Lipschitz continuous with the Lipschitz constant,
    \begin{align*}
        L_k \left( B + \frac{R}{\sigma} \sqrt{ 2 \gamma_t + 2 \log \left( \frac{d}{\delta} \right)} \right)
    \end{align*}
    with probability at least $1 - \delta$.
    \label{lem:lipschitz_posterior_mean}
\end{lemma}
We show the proof in Appendix~\ref{sec:proof_lipschitz_posterior_mean}.
Since the MIG $\gamma_T$ is sublinear for the kernels on which we mainly focus, the upper bound $\cO(L_k \sqrt{\gamma_t})$ is tighter than $\cO(L_k \sqrt{t \log t})$.

In the Bayesian setting, the upper bound of the Lipschitz constant for $\mu_T$ has not been shown to our knowledge.
Therefore, we show the following lemma:
\begin{lemma}
    Fix $\delta \in (0, 1)$ and $t \in [T]$.
    Suppose that Assumptions~\ref{assump:Bayesian} and \ref{assump:Bayesian_continuous} hold and the kernel has mixed partial derivative $\frac{\partial^2 k(\*x, \*z)}{ \partial x_j \partial z_j}$ for all $j \in [d]$.
    Set $a$ and $b$ as in Lemma~\ref{assump:Bayesian_continuous}.
    Assume that $(\*x_i)_{i \in [t]}$ is independent of $(\epsilon_i)_{i \in [t]}$ and $f$.
    Then, $\mu_t$ and $r_t(\*x) \coloneqq f(\*x) - \mu_t(\*x)$ satisfies the following:
    \begin{align*}
        \Pr \left( \sup_{\*x \in \cX} \left| \frac{\partial \mu_t(\*u)}{\partial u_j} \Big|_{\*u = \*x} \right| > L \right) \leq 2a \exp \left( - \frac{L^2}{b^2} \right), \\
        \Pr \left( \sup_{\*x \in \cX} \left| \frac{\partial r_t(\*u)}{\partial u_j} \Big|_{\*u = \*x} \right| > L \right) \leq 2a \exp \left( - \frac{L^2}{b^2} \right), 
    \end{align*}
    for all $j \in [d]$.
    \label{lem:bayesian_lipschitz_posterior_mean}
\end{lemma}
See Appendix~\ref{sec:proof_bayesian_lipschitz_posterior_mean} for the proof, in which we leverage Slepian's inequality~\citep[Proposition~A.2.6 in][]{van1996weak} and the fact that the derivative of the sample path follows GP jointly when the kernel is differentiable.

By leveraging the above results, even if $\cX$ is continuous, we can obtain the following upper bound of Eq.~\eqref{eq:target_error}:
\begin{lemma}
    Suppose that Assumptions~\ref{assump:frequentist} and ~\ref{assump:frequentist_continuous} hold.
    Fix $\delta \in (0, 1)$ and $T \in \NN$.
    Then, the following holds with probability at least $1 - \delta$:
    \begin{align*}
        E_T 
        &\leq 2 \beta_{\delta, T} \max_{p \in \cP} \EE_{p(\*x^*)} \left[  \sigma_T^2(\*x^*) \right] 
        + \cO \left( \frac{\max\{\gamma_T, \log(\frac{T}{\delta})\}}{T^2} \right).
    \end{align*}
    where $\beta_{\delta, T} = \left( B + \frac{R}{\sigma} \sqrt{ 2 d \log \left( T d r + 1 \right) + 2 \log \left( \frac{4}{\delta} \right)} \right)^2$.
    \label{lem:UB_error_frequentist_continuous}
\end{lemma}
\begin{lemma}
    Suppose that Assumptions~\ref{assump:Bayesian} and \ref{assump:Bayesian_continuous} hold.
    Fix $\delta \in (0, 1)$ and $T \in \NN$.
    Then, the following holds with probability at least $1 - \delta$:
    \begin{align*}
        E_T 
        &\leq 2 \beta_{\delta, T} \max_{p \in \cP} \EE_{p(\*x^*)} \left[  \sigma_T^2(\*x^*) \right] 
        + \cO\left( \frac{\log(\frac{T}{\delta})}{T^2} \right),
    \end{align*}
    where $\beta_{\delta, T} = 2d \log (T d r + 1) + 2 \log (2 / \delta)$.
    \label{lem:UB_error_bayesian_continuous}
\end{lemma}
See Appendices~\ref{sec:proof_UB_error_frequentist_continuous} and ~\ref{sec:proof_UB_error_bayesian_continuous} for the proof.

Consequently, we can  minimize Eq.~\eqref{eq:target_error} by minimizing $\max_{p \in \cP} \EE_{p(\*x^{*})}\left[ \sigma^2_{T}(\*x^{*}) \right]$.
In this perspective, the US and RS are theoretically guaranteed because of $\max_{p \in \cP} \EE_{p(\*x^{*})}\left[ \sigma^2_{T}(\*x^{*}) \right] \leq \max_{\*x \in \cX} \sigma^2_T (\*x)$ and Proposition~\ref{prop:us_rs}.
However, the US and RS do not incorporate the information of $\cP$.
Therefore, the practical effectiveness of the US and RS is limited.

\subsection{Other Performance Mesuares}

Although we mainly discuss the squared error, other measures can also be bounded from above:
\begin{lemma}
    The worst-case expected absolute error for any $T \in \NN$ is bounded from above as follows:
    \begin{align*}
        \max_{p \in \cP} \EE_{p(\*x^{*})} \left[ |f(\*x^{*}) - \mu_T(\*x^{*})| \right]
        \leq \sqrt{E_T},
        % &\coloneqq \max_{p \in \cP} \EE_{p(\*x^{*})} \left[ ( f(\*x^{*}) - \mu_T(\*x^{*}) )^2 \right]
    \end{align*}
    where $E_T$ is defined as in Eq.~\eqref{eq:target_error}.
    \label{lem:UB_absolute_error}
\end{lemma}
\begin{lemma}
    The worst-case expectation of entropy for any $T \in \NN$ is bounded from above as follows:
    \begin{align*}
        \max_{p \in \cP} \EE_{p(\*x^{*})} \left[ H\left[ f(\*x^*) \mid \cD_T \right] \right]
        % &= \max_{p \in \cP} \EE_{p(\*x^{*})} \left[ \frac{1}{2} \log \left(2 \pi e \sigma_T^2(\*x^*) \right) \right] \\
        &\leq \frac{1}{2} \log \left(2 \pi e \tilde{E}_T \right),
        % &\leq \frac{1}{2} \log \left(2 \pi e \max_{p \in \cP} \EE_{p(\*x^{*})} \left[ \sigma_T^2(\*x^*) \right] \right),
        % &= \cO\left( \log \left( \max_{p \in \cP} \EE_{p(\*x^{*})} \left[ \sigma_T^2(\*x^*) \right] \right)\right)
    \end{align*}
    where $\tilde{E}_T = \max_{p \in \cP} \EE_{p(\*x^{*})}\left[ \sigma^2_{T}(\*x^{*}) \right]$ and $H[f(\*x) \mid \cD_T] = \log \left(\sqrt{2 \pi e} \sigma_T(\*x) \right)$ is Shannon entropy.
    \label{lem:UB_entropy}
\end{lemma}
See Appendices~\ref{sec:UB_absolute_error_proof} and \ref{sec:UB_entropy_proof} for the proof.
Therefore, minimizing $\max_{p \in \cP} \EE_{p(\*x^{*})}\left[ \sigma^2_{T}(\*x^{*}) \right]$ also provides the convergence of the absolute error and the entropy\footnote{For the absolute error, we can design algorithms that directly reduce $\sigma_t$, not $\sigma_t^2$, and achieves the similar theoretical guarantee.}.

% \subsection{Discussion}

% Our problem setup can be seen as the generalization of the target distribution-aware AL and the AL for the worst-case error, i.e., $\max_{\*x \in \cX} ( f(\*x) - \mu_T(\*x) )^2$.
% %
% This is because our problem is equivalent to the target distribution-aware AL if we set $|\cP| = 1$ and to the worst-case error minimization if $\cP$ includes $\{p \in \cP_{\cX} \mid \exist \*x \in \cX, p(\*x) = 1 \}$, where $\cP_{\rm \cX}$ is the set of the distributions over $\cX$.
% %
% Clearly, for the worst-case analysis for $\max_{\*x \in \cX} ( f(\*x) - \mu_T(\*x) )^2$, we must use the method that reduce the largest variance $\max_{\*x \in \cX} \sigma_t^2(\*x)$.
% %
% This is satisfied by the US and RS, as shown in Proposition~\ref{prop:us_rs}.
\section{Proposed Methods and Analysis}
\label{sec:proposed}

We aim to design algorithms that enjoy both a similar convergence guarantee as the US and RS and practical effectiveness, incorporating the information of $\cP$.
In particular, we consider two algorithms inspired by the greedy algorithm and the RS and show theoretical guarantees.
Algorithm~\ref{alg:proposed} shows the pseudo-code of the proposed algorithms.

\begin{algorithm}[!t]
    \caption{Proposed DRAL methods}\label{alg:proposed}
    \begin{algorithmic}[1]
        \Require Domain $\cX$, GP prior $\mu$ and $k$, ambiguity set $\cP$
        \State $\cD_{0} \gets \emptyset$
        \For{$t = 1, \dots, T$}
            \State Update $\sigma_{t-1}^2 (\cdot)$ according to Eq.~\eqref{eq:GP}
            \State Compute $\*x_t$ according to Eq.~\eqref{eq:RS} or Eq.~\eqref{eq:greedy}
        \EndFor
        \State Observe $y_1, \dots, y_T$ 
        \State Update $\mu_{T} (\cdot)$ and $\sigma_{T}^2 (\cdot)$ according to Eq.~\eqref{eq:GP}
        \State \Return $\mu_{T} (\cdot)$ and $\sigma_{T}^2 (\cdot)$
    \end{algorithmic}
\end{algorithm}

\subsection{Algorithms}

First, we consider the RS-based algorithm.
The algorithm is straightforward as follows:
\begin{align}
    \*x_t \sim p_t(\*x),
    \label{eq:RS}
\end{align}
where $p_t(\*x) = \argmax_{p \in \cP} \EE_{p(\*x^{*})}[\sigma_{t-1}^2 (\*x^{*})]$ and we assume that we can generate the sample from $p_t$.
By using the worst-case distribution $p_t$ for each iteration, this algorithm incorporates the information of $\cP$.

% \begin{itemize}
%     \item 貪欲法は一般に強いためそれに基づく方法を考える.
%     \item しかし, $\max_{p \in \cP}$を次ステップの全候補に対し計算するのに多大な計算量が必要なため, 貪欲法すら計算不可能
%     \item そこで, $p_t(\*x) = \argmax_{p \in \cP} \EE_{p(\*x)}[\sigma_t^2 (\*x)]$を固定した貪欲法を考える.
%     \item しかし, もはや貪欲法ですらないこの方法の近似保証は我々には難しかった.
%     \item そこで, 保守的なUSに理論保証があることから, 少し保守的になる (uncertainな候補を選択する) ように候補を制限することを考えた.
%     \item 最終的なアルゴリズムをXXXに示す.
% \end{itemize}

Second, we consider the greedy algorithm since its practical efficiency has often been reported~\citep[e.g., ][]{bian2017guarantees}.
However, in our setup, the algorithm that greedily decreases the expected posterior variance should be
\begin{align*}
    \argmin_{\*x \in \cX} \max_{p \in \cP} \EE_{p(\*x^{*})}[\sigma_{t}^2 (\*x^{*} \mid \*x)],
\end{align*}
which requires huge computational time in general due to min-max optimization.
Thus, we consider an approximately greedy algorithm as follows:
\begin{align*}
    \argmin_{\*x \in \cX} \EE_{p_t(\*x^{*})}[\sigma_{t}^2 (\*x^{*} \mid \*x)],
\end{align*}
where $p_t(\*x) = \argmax_{p \in \cP} \EE_{p(\*x^{*})}[\sigma_{t-1}^2 (\*x^{*})]$ is the worst-case distribution defined by $(\*x_i)_{i \in [t-1]}$.
On the other hand, the theoretical guarantee for this algorithm is challenging for us.
Hence, inspired by the fact that the US has a theoretical guarantee, we set the constraint so that the chosen input is uncertain than $\EE_{p_t(\*x^{*})}[\sigma_{t-1}^2 (\*x^{*})]$:
\begin{align}
    \*x_t = \argmin_{\*x \in \cX_t} \EE_{p_t(\*x^{*})}[\sigma_t^2 (\*x^{*} \mid \*x)],
    \label{eq:greedy}
\end{align}
where $\cX_t \coloneqq \{ \*x \in \cX \mid \sigma^2_{t-1}(\*x) \geq \EE_{p_t(\*x^{*})}[\sigma_{t-1}^2 (\*x^{*})] \}$.
Note that $|\cX_t| \geq 1$ holds due to the definition.

\paragraph{Necessity of Constraints:}
We considered that the constraint regarding $\sigma^2_{t-1}(\*x)$ makes the analysis easy since the US that maximize $\sigma_{t-1}(\*x_t)$ achieves the error convergence, as shown in Proposition~\ref{prop:us_rs}.
Therefore, we employ the constraint on $\cX_t$.
We set the threshold of the constraint as $\EE_{p_t(\*x^{*})}[\sigma_{t-1}^2 (\*x^{*})]$ sake of the analysis.
On the other hand, our experimental results suggest that the greedy (approximated) expected error reduction algorithm without the constraint shows superior performance.
Therefore, removing or alleviating the constraint can be important future work.

\paragraph{Computational Complexity:}
The computation of the GP has $O(T^3)$ computational complexity, which can be alleviated by scalable GP learning approaches~\citep{liu2020when}.
However, more careful proofs incorporating an approximation error in GP learning are required to derive similar theoretical analyses as ours, as with \citep{vakili2022improved}.
On the other hand, the computational complexity of the maximization $\max_{p \in \cP} \EE_{p(\*x^*)} [\sigma^2_{t-1} (\*x^*)]$, the expectation $\EE_{p_t(\*x^*)} [\sigma^2_{t-1} (\*x^*)]$ and the sampling from $p_t(\*x)$ depends on the ambiguity set $\cP$ and may increase in proportion to $d$.
Although our experiments focus on the set of discrete probability distributions with a moderate size of $|\cX|$, the above computations may be complicated if $\cP$ is the set of continuous probability distributions or $|\cX|$ is huge.
Extensions to such more computationally intractable ambiguity sets $\cP$, e.g., the ball defined by Wasserstein distance and Kullback--Leibler divergence~\citep{hu2013kullback,frogner2021incorporating}, is crucial future work.

%%%%%%%%%%%%%%%%%%%%%%%%%%%%%%%%%%%%%%%%%%%%%%%%%%%%%%%%%%%%%%%%%%%%%%%%%%%%%%%%%%%%%%%%%%%%%%%%%%
\subsection{Analysis}
\label{sec:analysis}

Here, we show the error convergence by Eqs.~\eqref{eq:RS} and \eqref{eq:greedy}:
\begin{theorem}
    Fix $\delta \in (0, 1)$.
    Assume that $\cX \subset \RR^d$ is a compact subset.
    If we run Algorithm~\ref{alg:proposed} with Eq.~\eqref{eq:RS}, the following holds with probability at least $1 - \delta$:
    \begin{align*}
        \max_{p \in \cP} \EE_{p(\*x^{*})}[\sigma_{T}^2 (\*x^{*})] 
        &\leq \frac{2 C_1 \gamma_T}{T} + \cO \left( \frac{\log (1 / \delta)}{T} \right),
    \end{align*}
    where $C_1 = 2 / \log(1 + \sigma^{-2})$.
    \label{theo:error_convergence_RS}
\end{theorem}
\begin{theorem}
    Assume that $\cX \subset \RR^d$ is a compact subset.
    If we run Algorithm~\ref{alg:proposed} with Eq.~\eqref{eq:greedy}, the following holds:
    \begin{align*}
        \max_{p \in \cP} \EE_{p(\*x^{*})}[\sigma_{T}^2 (\*x^{*})] 
        &\leq \frac{C_1 \gamma_T}{T},
    \end{align*}
    where $C_1 = 2 / \log(1 + \sigma^{-2})$.
    \label{theo:error_convergence_greedy}
\end{theorem}
See Appendix~\ref{sec:proposed_proof} for the proof, in which Lemma~3 in \citep{kirschner2018-information} is used to show Theorem~\ref{theo:error_convergence_RS}.

Consequently, our proposed methods achieve almost the same convergence as those of the US and RS shown in Proposition~\ref{prop:us_rs}.
Furthermore, by combining Lemmas~\ref{lem:UB_error_discrete}, \ref{lem:UB_error_frequentist_continuous}, and \ref{lem:UB_error_bayesian_continuous}, we can see that the upper bound of $E_T$:
\begin{corollary}
    Fix $\delta \in (0, 1)$ and $T \in \NN$.
    Then, if we run Algorithm~\ref{alg:proposed}, the following hold with probability at least $1 - \delta$:
    \begin{enumerate}
        \item When Assumption~\ref{assump:Bayesian} or Assumptions~\ref{assump:frequentist} holds, 
        \begin{align*}
            E_T = \cO\left( \frac{\log (|\cX| / \delta) \gamma_T}{T} \right);
        \end{align*}
        \item When Assumptions~\ref{assump:Bayesian} and \ref{assump:Bayesian_continuous} or Assumptions~\ref{assump:frequentist} and \ref{assump:frequentist_continuous} hold, 
        \begin{align*}
            E_T = \cO\left( \frac{\log (T / \delta) \gamma_T}{T} \right),
        \end{align*}
    \end{enumerate}
\end{corollary}
\begin{proof}
    We can obtain the result by combining Lemmas~\ref{lem:UB_error_discrete}, \ref{lem:UB_error_frequentist_continuous}, and \ref{lem:UB_error_bayesian_continuous}, Theorems~\ref{theo:error_convergence_RS} and \ref{theo:error_convergence_greedy}, and the union bound.
    Note that we assume $|\cX| > T$.
\end{proof}
Thus, the error incurred by the proposed algorithms converges to $0$ with high probability for discrete and continuous input domains, at least with linear, SE, and Mat\'ern kernels.

\begin{figure*}[t]
    \centering
    \includegraphics[width=0.95\linewidth]{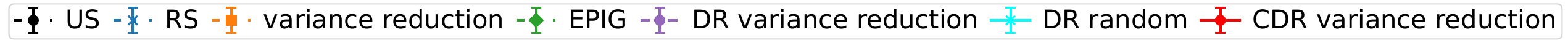}\\
    \includegraphics[width=0.24\linewidth]{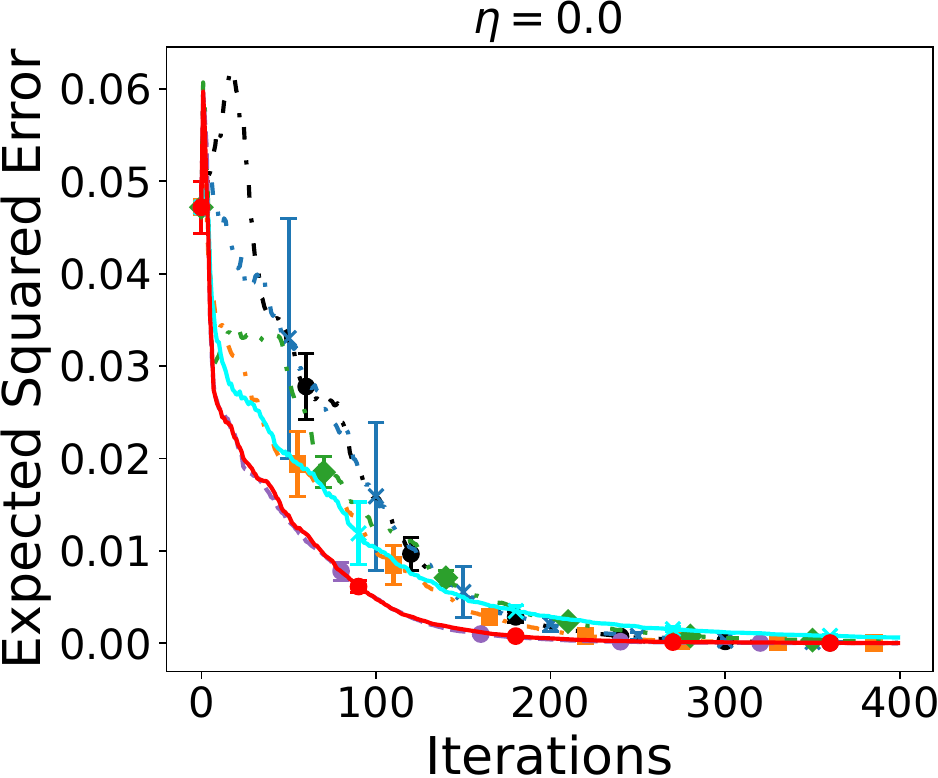}
    \includegraphics[width=0.24\linewidth]{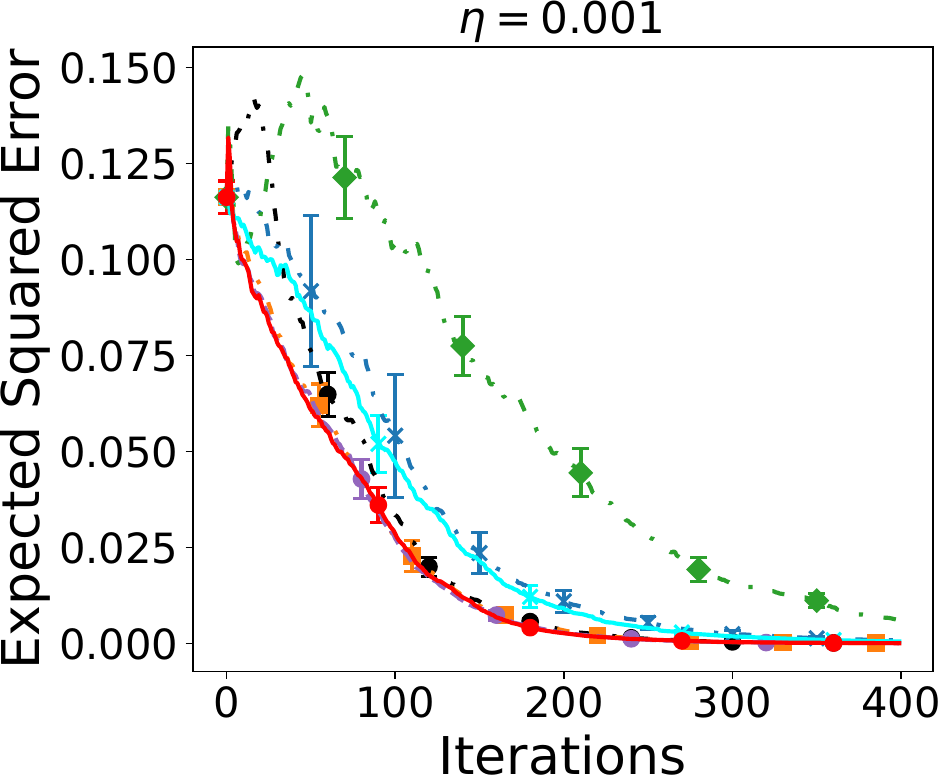}
    \includegraphics[width=0.24\linewidth]{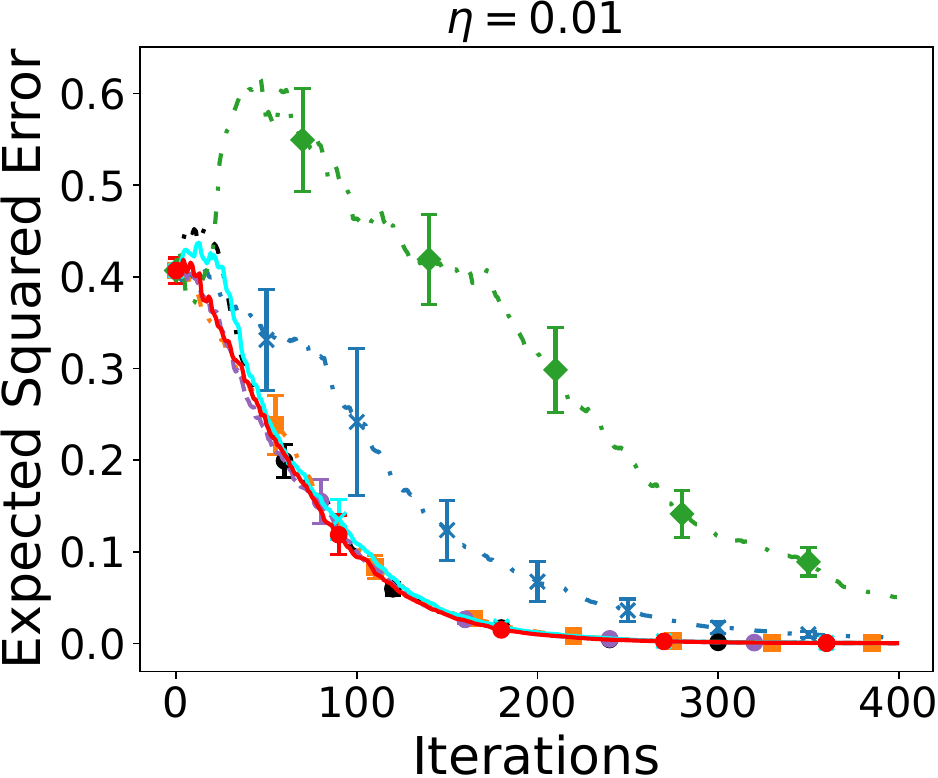}
    \includegraphics[width=0.24\linewidth]{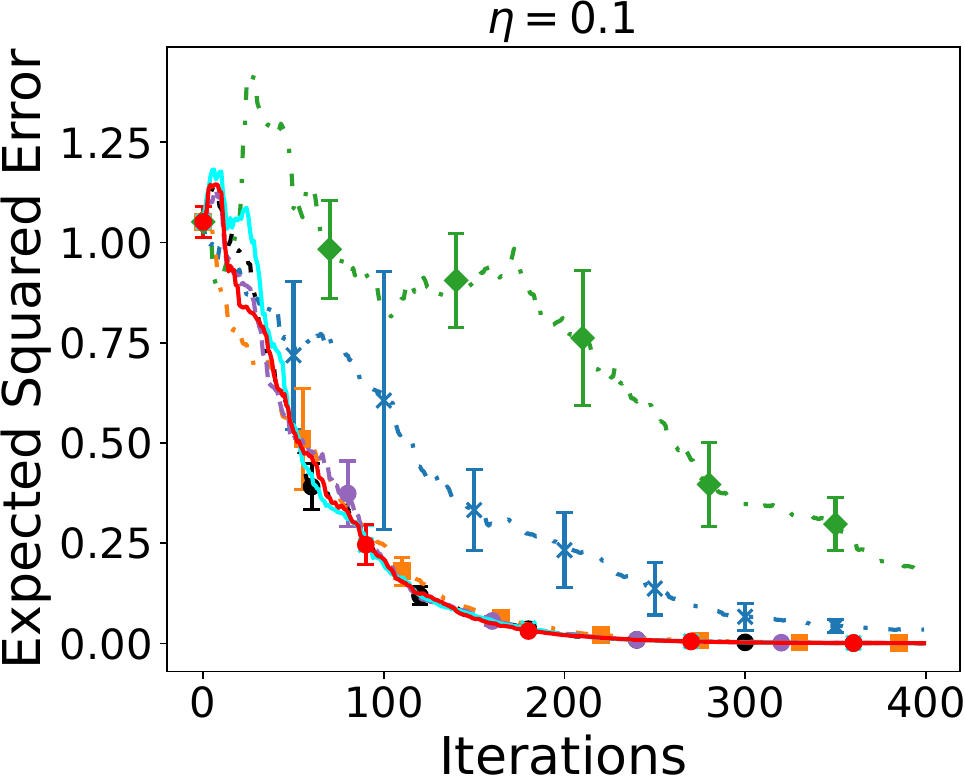}\\
    \includegraphics[width=0.24\linewidth]{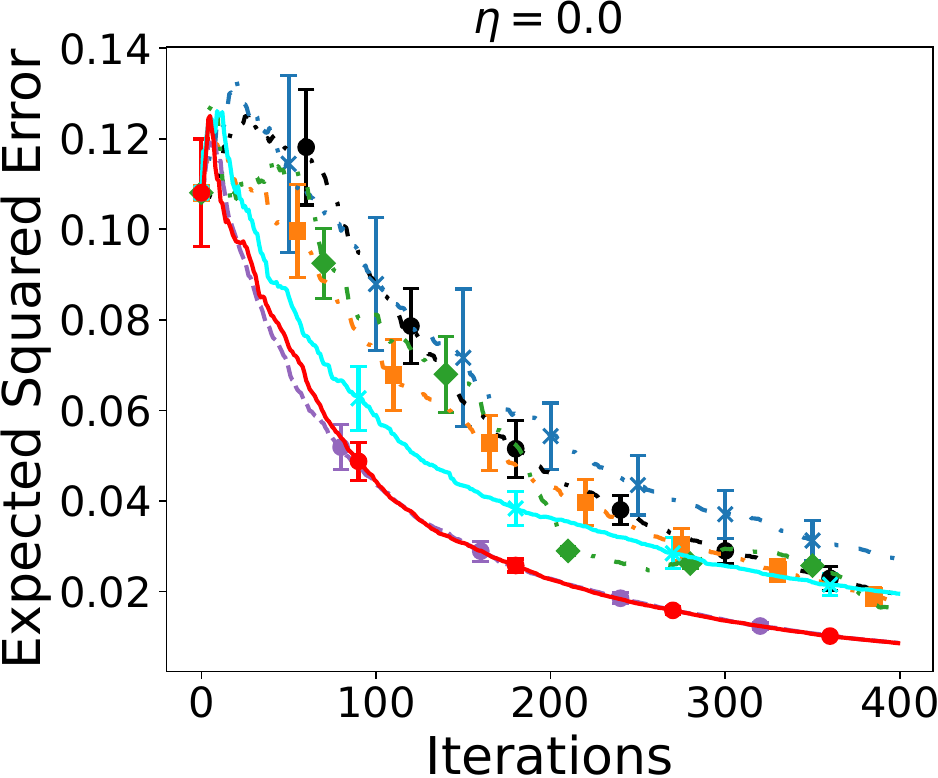}
    \includegraphics[width=0.24\linewidth]{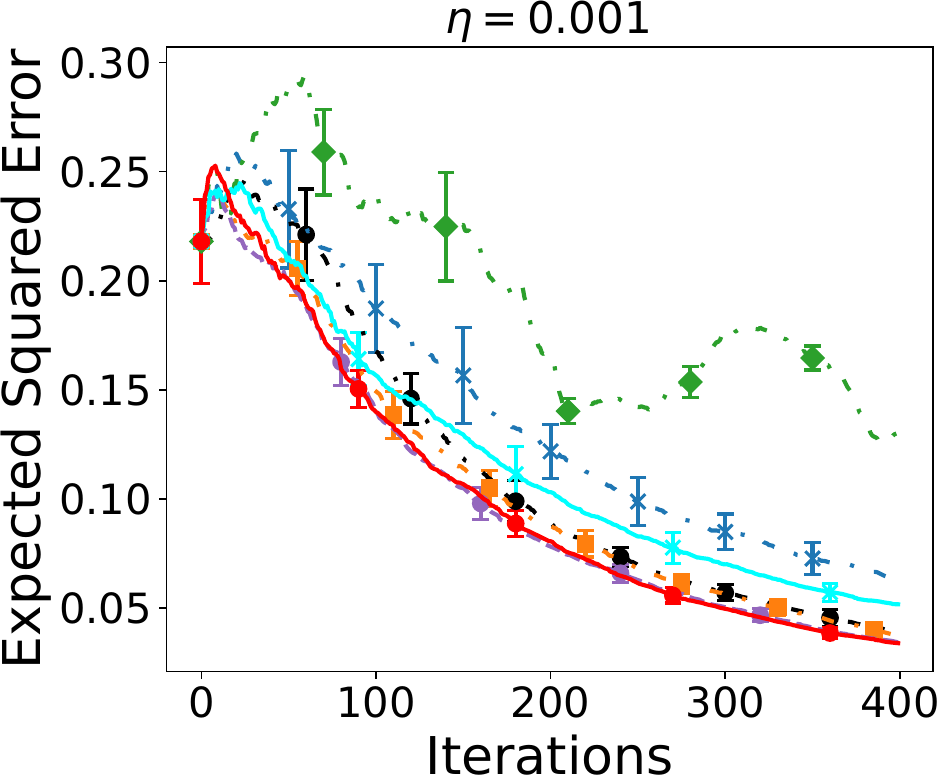}
    \includegraphics[width=0.24\linewidth]{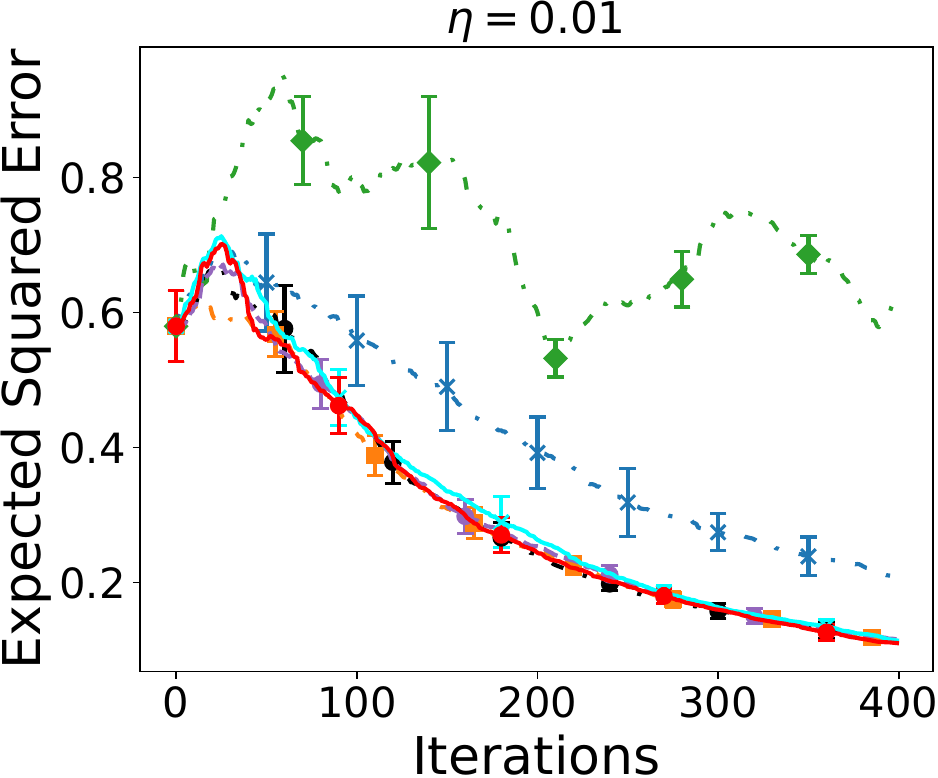}
    \includegraphics[width=0.24\linewidth]{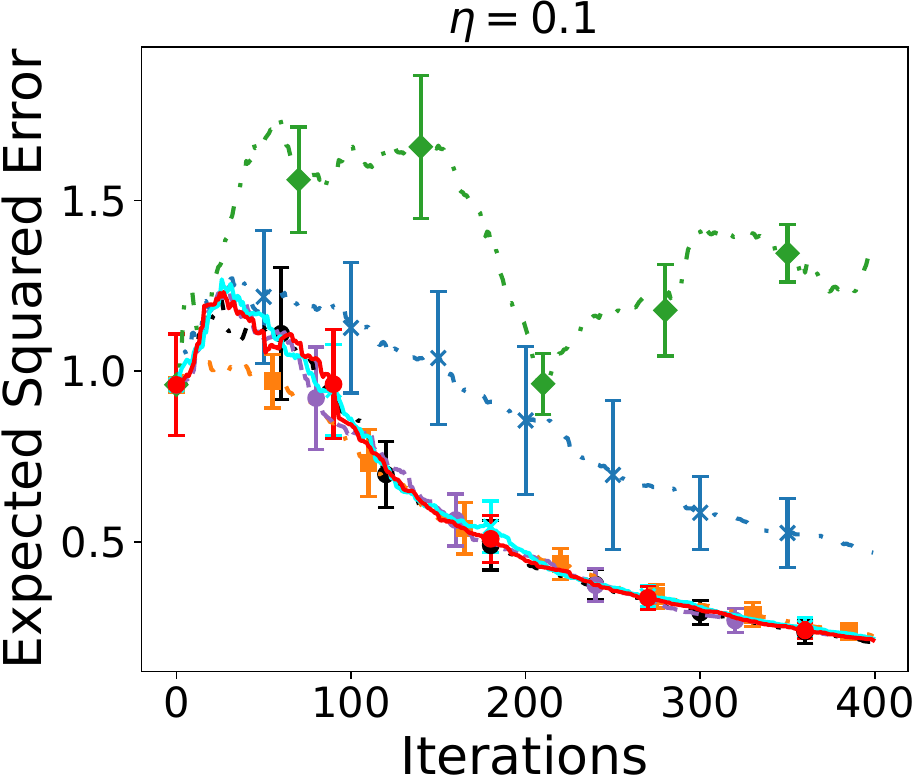}
    \caption{
        Result of the error $E_T$ in the synthetic data experiments with $\eta=0, 0.001, 0.01, 0.1$.
        The horizontal and vertical axes show the number of iterations and $E_T$, respectively.
        The error bar shows mean and standard errors for 20 random trials regarding the random initial point (and the algorithm's randomness).
        The top and bottom columns represent the results of the SE and Mat\'ern kernels, respectively.
        %
        % 最悪期待二乗誤差を用いて評価を行った場合の人工データ実験結果.
        % %
        % 横軸は反復数, 縦軸は最悪期待二乗誤差を示す.
        % %
        % ランダムに初期点を変更したことによる20回試行の実験結果の平均と標準誤差を示す.
        % %
        % 上段はRBFカーネル, 下段はMaternカーネルをGP回帰のカーネルとして用いた場合の結果を示す.
        % %
        % 参照分布は平均が${\* 0}$, 共分散行列が$0.2*{\*I}_3$の三次元正規分布とする. 
        % %
        % ここで, ${\*I}_3$は三次元の単位行列である.
        % %
        % また, 参照分布からの最大距離$\eta=0, 0.001, 0.01, 0.1$とする.
    }
    \label{fig:syn_results}
\end{figure*}

%%%%%%%%%%%%%%%%%%%%%%%%%%%%%%%%%%%%%%%%%%%%%%%%%%%%%%%%%%%%%%%%%%%%%%%%%%%%%%%%%%%%%%%
\section{Related Work}
\label{sec:related}

% AL algorithms
As discussed in Section~\ref{sec:intro}, many AL algorithms have been developed~\citep{settles2009-active}.
The AL algorithms for classification problems are heavily discussed compared with the regression problem~\citep[for example, ][]{houlsby2011bayesian,zhao2021active,bickford2023-prediction}.
In particular, theoretical properties for binary classification problems are well-investigated~\citep{hanneke2014theory}.
On the other hand, the theoretical analysis of AL for the regression problem is relatively limited.

% Experimental design methods
AL is often referred to as optimal experimental design (OED)~\citep{lindley1956-on,cohn1993neural,Chaloner1995bayesian,cohn1996active,ryan2007modern}.
The OED frameworks aim to reduce the uncertainty of target parameters or statistical models.
For this purpose, many measures for the optimality have been proposed, such as A-optimality (average), D-optimality (determinant), and V-optimality (variance)~\citep{pukelsheim2006optimal,zhu2017near}.
The OED methods for various models, such as the linear model~\citep[e.g., ][]{zhu2017near}, neural network~\citep[e.g., ][]{cohn1993neural}, and GPs~\citep[e.g., ][]{yu2006active}, have been proposed.
Our analysis concentrates on the V-optimality of the GPR~\citep{Seo2000gaussian,yu2006active,Shoham2023experimental} and its DR variant, for which, to our knowledge, a theoretical guarantee has not been shown.

% Several examples
% For example, \citet{krause2008-near,Guestrin2005-near} proposed that the greedy algorithm for the mutual information, which corresponds to the D-optimality, is the near-optimal algorithm for the GP model.
% %
% Many other studies have concentrated on the information-based criteria for Bayesian models~\citep{mackay1992information,houlsby2011bayesian,kirsch2021test,kirsch2022unifying,bickford2023-prediction}.
% %
% However, the relationship between the prediction error and the mutual information is not obvious.
%
% On the other hand, to our knowledge, theoretical analysis of classical OED methods mainly concentrated on the linear models~\citep{pukelsheim2006optimal}.

% subset selection
In OED or AL, subset selection algorithms~\citep{das2008algorithms} are often leveraged.
The subset selection is a general problem whose goal is to find the subset that maximizes some set function.
Therefore, the AL can be seen as the subset selection of $\*x_1, \dots, \*x_t \in \cX$.
In this literature, the submodular property of the set function, for which the greedy algorithm can be optimal, is commonly investigated~\citep{das2008algorithms,krause2008-near,Guestrin2005-near,bian2017guarantees}.
The criteria for the AL, such as the D-optimality of the GPR~\citep{krause2008-near,Guestrin2005-near}, sometimes satisfy the submodular property.
%
% On the other hand, the V-optimality of the GPR does not satisfy the submodularity in general.
%
Furthermore, \citet{das2008algorithms} have shown sufficient conditions for the greedy algorithms to be optimal in Theorem~3.4 (an assumption can be rephrased as $k(\*x, \*x^\prime) \leq \frac{1}{4T}$ in our problem) and Section~8.
However, even if we consider minimizing $\EE_{p(\*x^*)}\left[ \sigma_T^2 (\*x^*) \right]$ with $|\cP| = 1$, these conditions and the submodularity do not hold in general.
Therefore, the DR maximization of submodular function~\citep[e.g., ][]{krause2008robust,staib2019distributionally} also cannot be applied directly.

% information gain-based approach
% Most AL methods based on the Bayesian models are based on mutual information, also called information gain~\citep{mackay1992information,houlsby2011bayesian}.
% %
% The information-based approach can apply to the GPR model due to its versatility.
% %
% When the mutual information between the target function we want to learn about and the label we can obtain is considered, the US can be seen as a greedy algorithm for maximizing mutual information.
% %
% Then, the optimality of the information gain-based approach is theoretically shown using the sub-modularity of mutual information~\citep{Guestrin2005-near,krause2008-near}.
% %
% Furthermore, transductive information-based AL algorithms have been proposed~\citep{kirsch2021test,kirsch2022unifying,bickford2023-prediction}.
% %
% However, there are no theoretical guarantees since transductive variants do not satisfy the sub-modularity.

% target distribution-aware AL
Several studies have discussed the target distribution-aware AL.
At least from \citep{sugiyama2005active}, the effectiveness of AL incorporating the information of target distribution for misspecified models has been discussed.
Transductive AL~\citep{Seo2000gaussian,yu2006active,Shoham2023experimental} can be interpreted as the expected error minimization when the uniformly random target distribution $p(\*x) = 1 / |\cX|$ is specified.
\citet{kirsch2021test,kirsch2022unifying,bickford2023-prediction} extended this setting so that an arbitrary target distribution can be considered.
These existing methods are heuristic algorithms that do not guarantee the convergence of the error.
On the other hand, \citet{hubotter2024-transductive} show transductive AL methods with theoretical guarantees.
However, for the VTR algorithm in \citep{hubotter2024-transductive}, their analysis assumes an assumption of submodularity, which may not hold as discussed immediately after Assumption~3.2 of \citep{hubotter2024-transductive}.
Furthermore, their analysis for the VTR algorithm results in $\max_{\*x \in \cX} \sigma^2_T(\*x) = O(|\cX|\gamma_T / T)$, which is vacuous in the usual regime that $T < |\cX|$ (see Theorems 3.3 and C.13 and Section~C.6.2 of \citet{hubotter2024-transductive}).

\citet{frogner2021incorporating} further extended to DRAL using the AF called expected model change~\citep{settles2009-active} for the non-Bayesian model.
In addition, \citet{Liu2015shift} considers DRAL for non-Bayesian classification models.
However, these methods are heuristic greedy algorithms and are not theoretically guaranteed for the prediction error.
% Relation to transductive learning, target distribution-aware AL, V-optimal design
% On the other hand, since $|\cP| = 1$ with any target distribution in our problem setup, our study includes the target distribution-aware AL discussed above, where only one test function is specified.

The DRAL is inspired by the DR learning (DRL)~\citep{chen2018robust,chen2020distributionally}.
DRL considers learning a robust statistical model by optimizing the model parameter so that the worst-case expected loss is minimized, where the worst-case is taken regarding the target distribution candidates called an \emph{ambiguity set}.
Therefore, DRAL is an intuitive extension of DRL to AL.

% core-set selection
Another related literature is core-set selection~\citep{sener2018active}, which selects the subset of the training dataset to maintain prediction accuracy while reducing the computational cost.
Our proposed methods can be applied to the core-set selection for the GPR.
However, its effectiveness may be limited since the information on training labels is not leveraged.

% BO, kernel bandit
Other highly relevant literature is kernelized bandits, also called Bayesian optimization (BO)~\citep{Kushner1964-new,Srinivas2010-Gaussian,Shahriari2016-Taking}.
BO aims for efficient black-box optimization using the GPR model.
For this purpose, several properties of GPs, such as the confidence intervals and the MIG, have been analyzed~\citep{Srinivas2010-Gaussian,vakili2021-optimal,vakili2021-information}.
Our analyses heavily depended on the existing results in this field.

%LSE
In addition, level set estimation (LSE)~\citep{gotovos2013active,bogunovic2016truncated,inatsu2024active} is an AL framework using the GPR model, which aims to classify the test input set by whether or not a black-box function value exceeds a given threshold.
In particular, \citet{inatsu2021active} consider the variant of LSE, which aims to classify by whether or not the DR measure defined by the black-box function exceeds a given threshold.
Thus, our problem setup differs from the problem of \citep{inatsu2021active}.

\begin{figure*}[t]
    \centering
    \includegraphics[width=0.95\linewidth]{fig/experiments/legend.pdf}\\
    \includegraphics[width=0.24\linewidth]{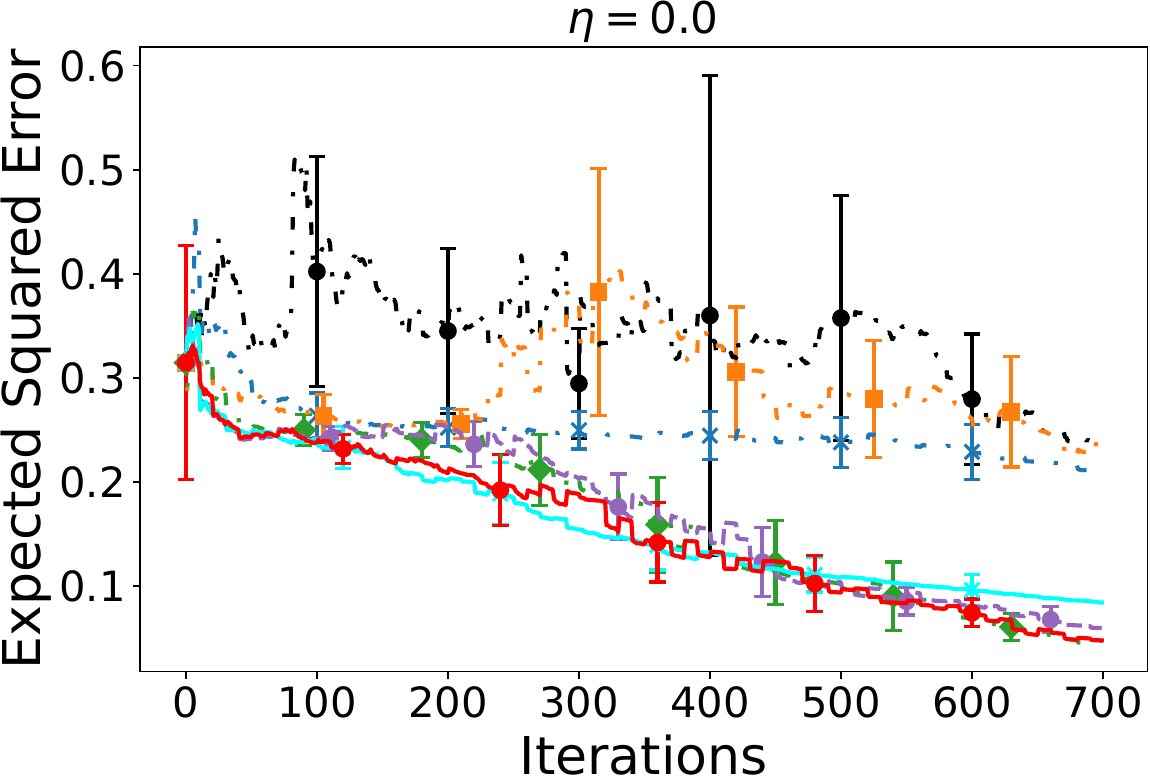}
    \includegraphics[width=0.24\linewidth]{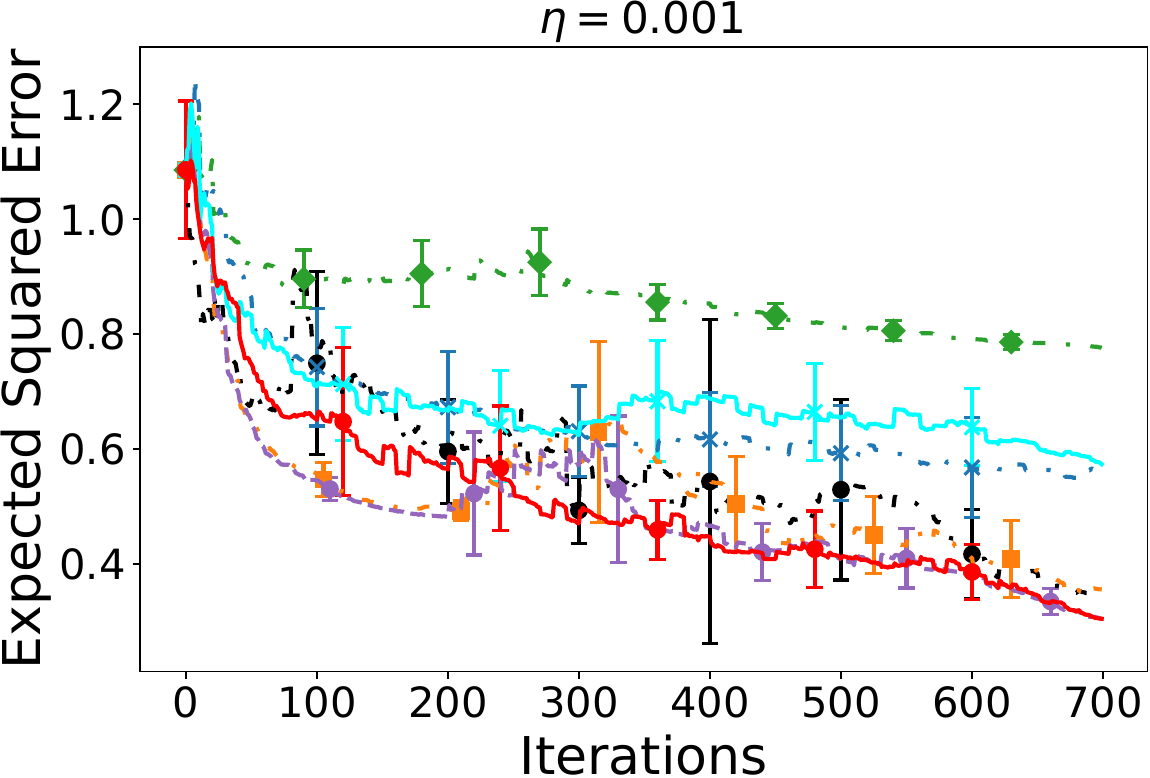}
    \includegraphics[width=0.24\linewidth]{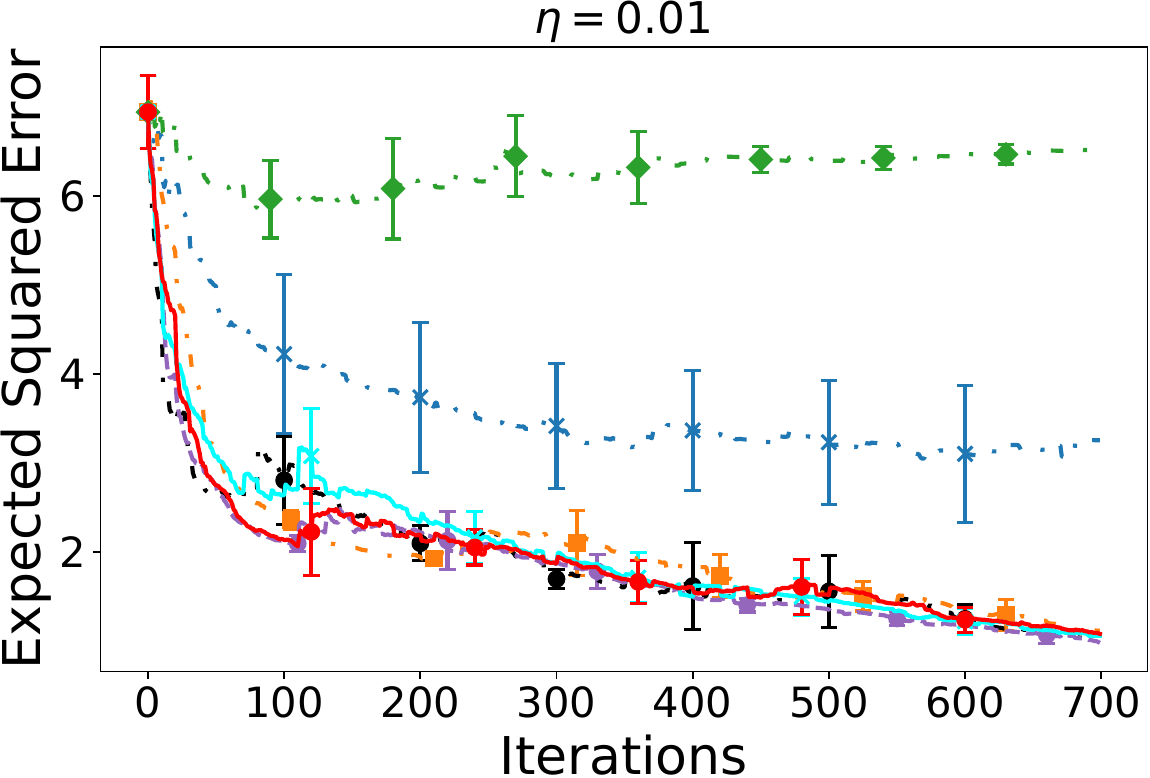}
    \includegraphics[width=0.24\linewidth]{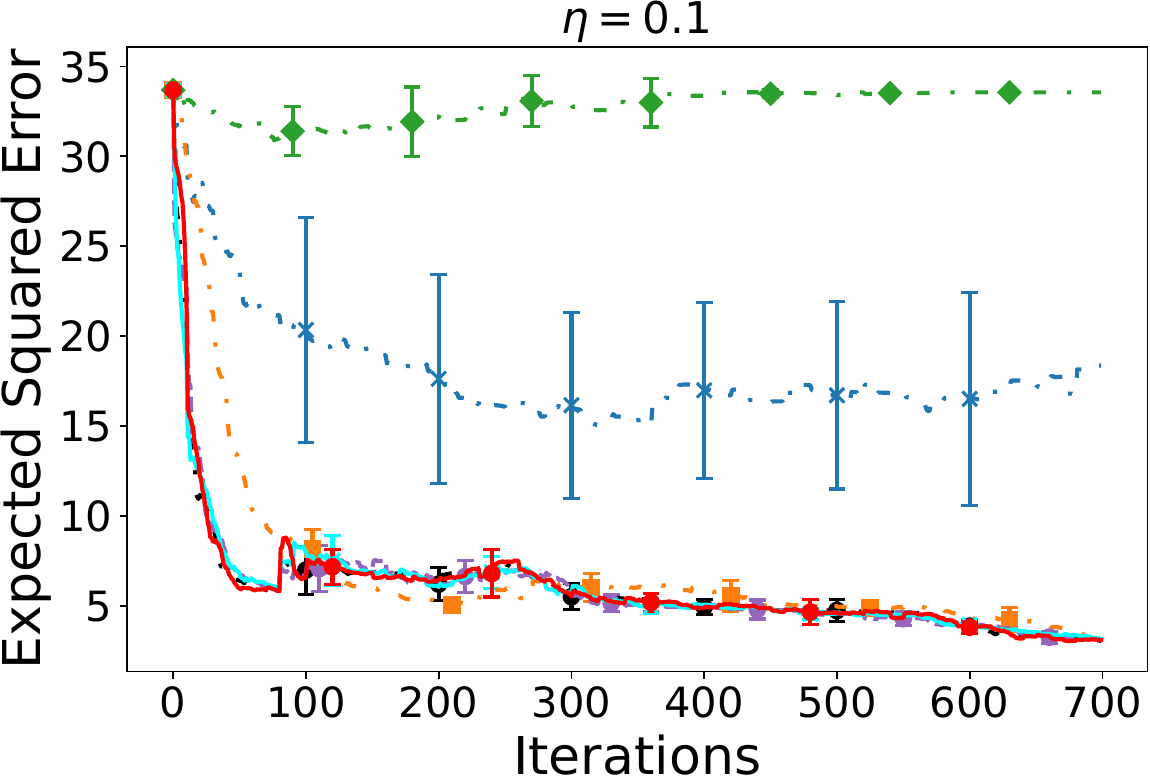}\\
    \includegraphics[width=0.24\linewidth]{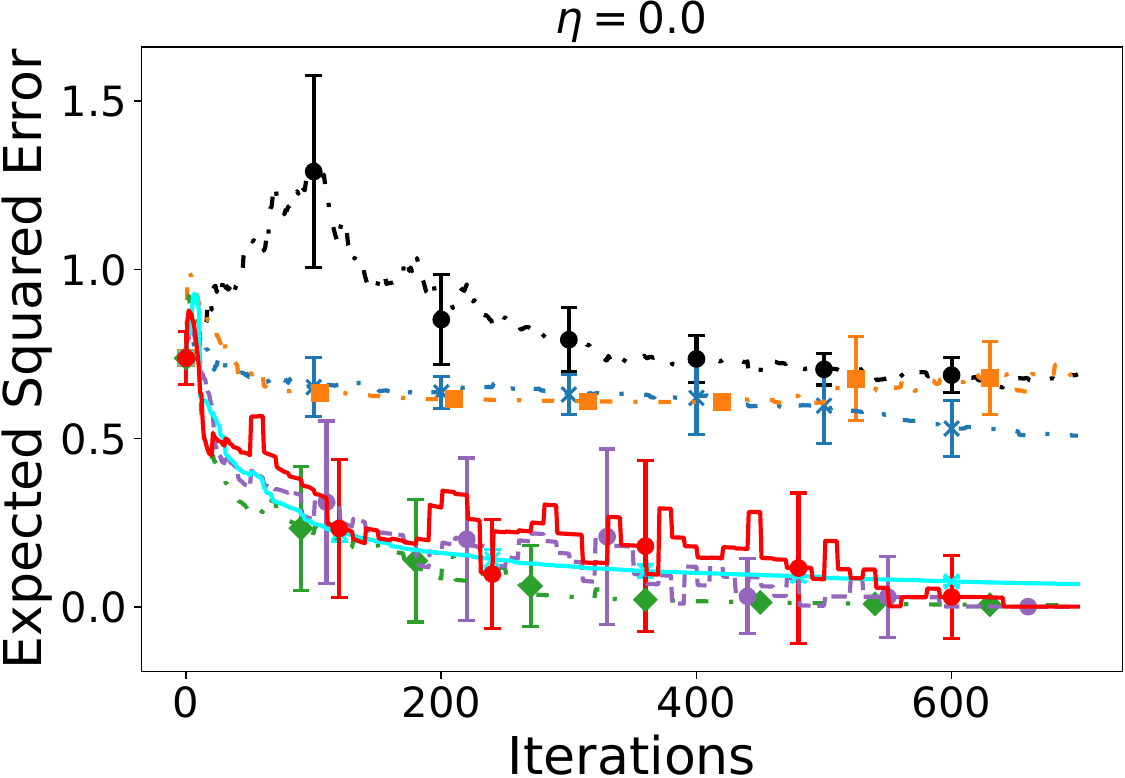}
    \includegraphics[width=0.24\linewidth]{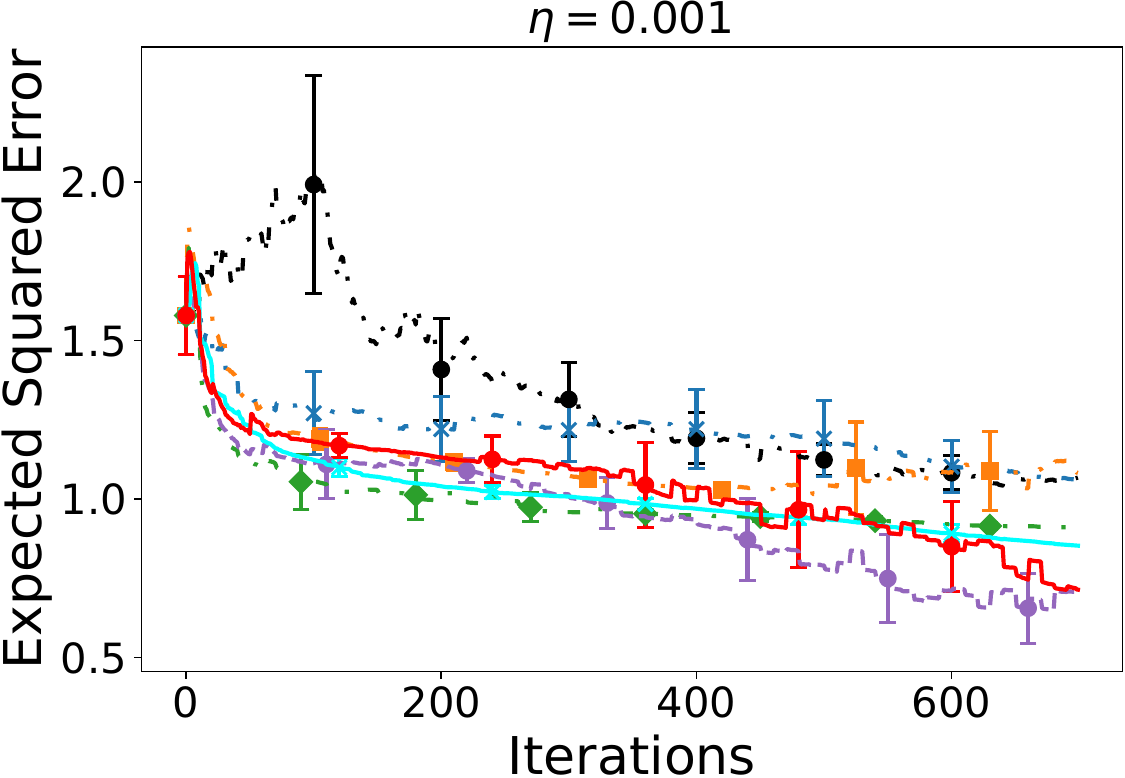}
    \includegraphics[width=0.24\linewidth]{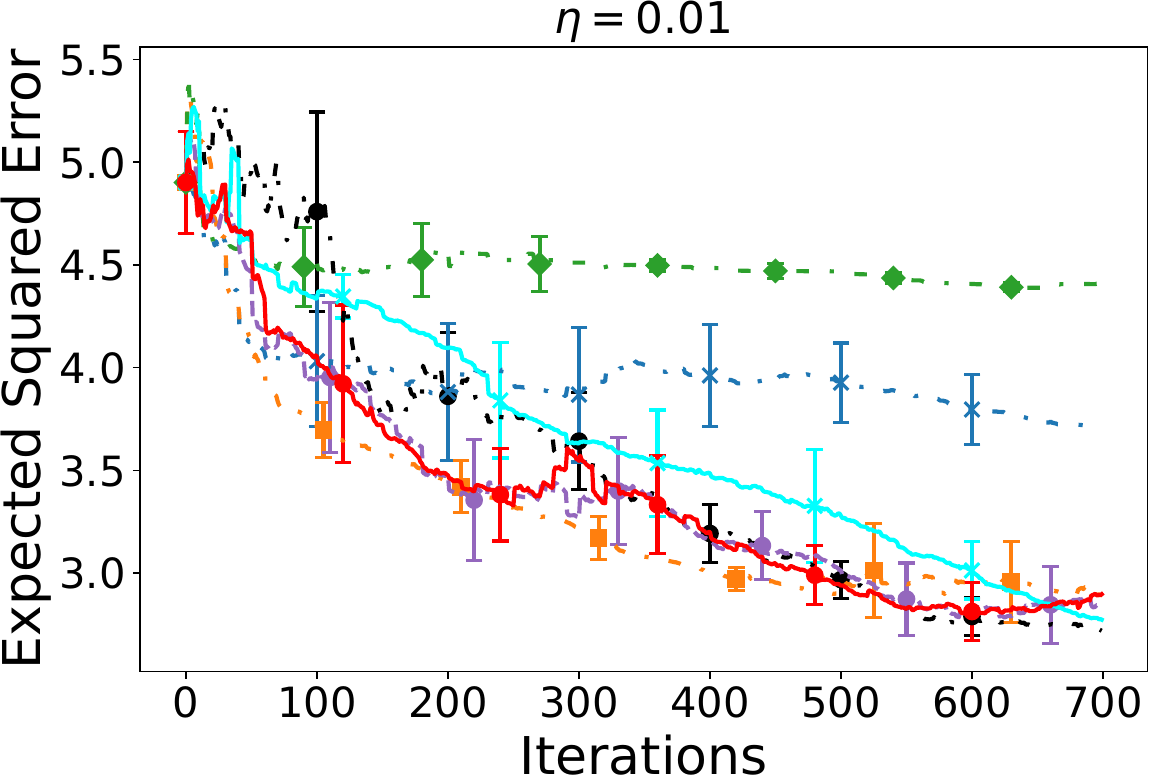}
    \includegraphics[width=0.24\linewidth]{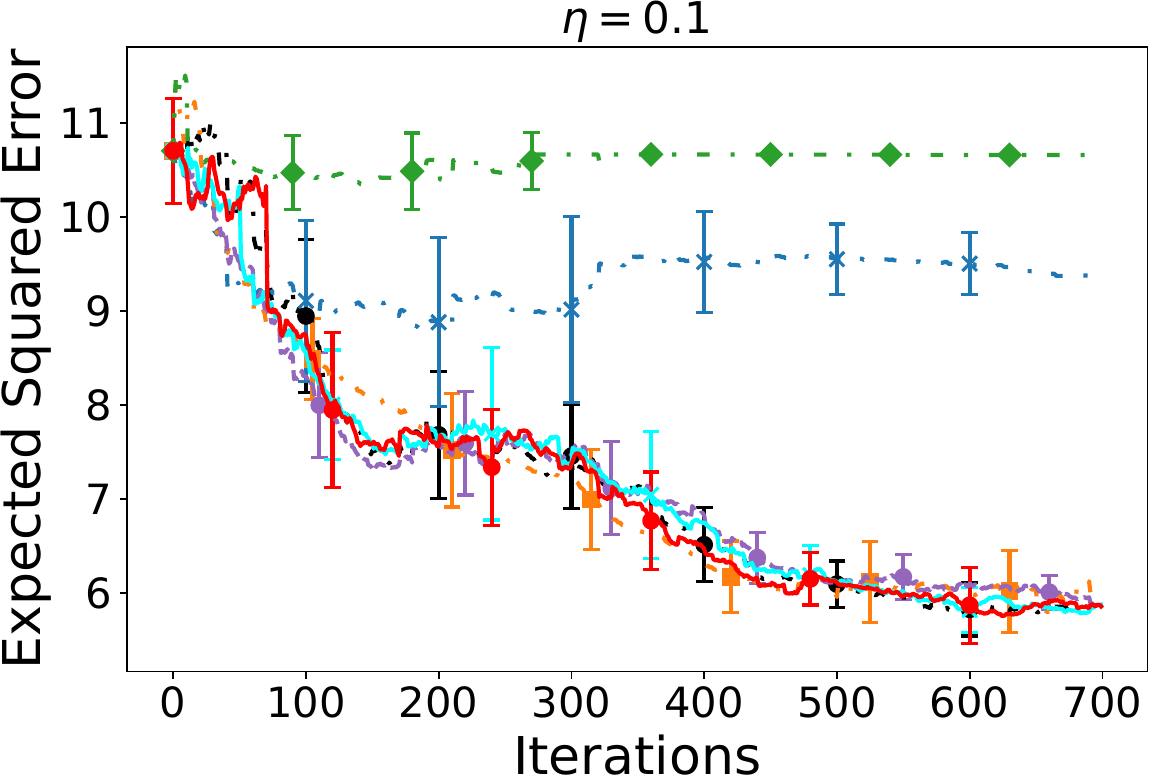}\\
    \includegraphics[width=0.24\linewidth]{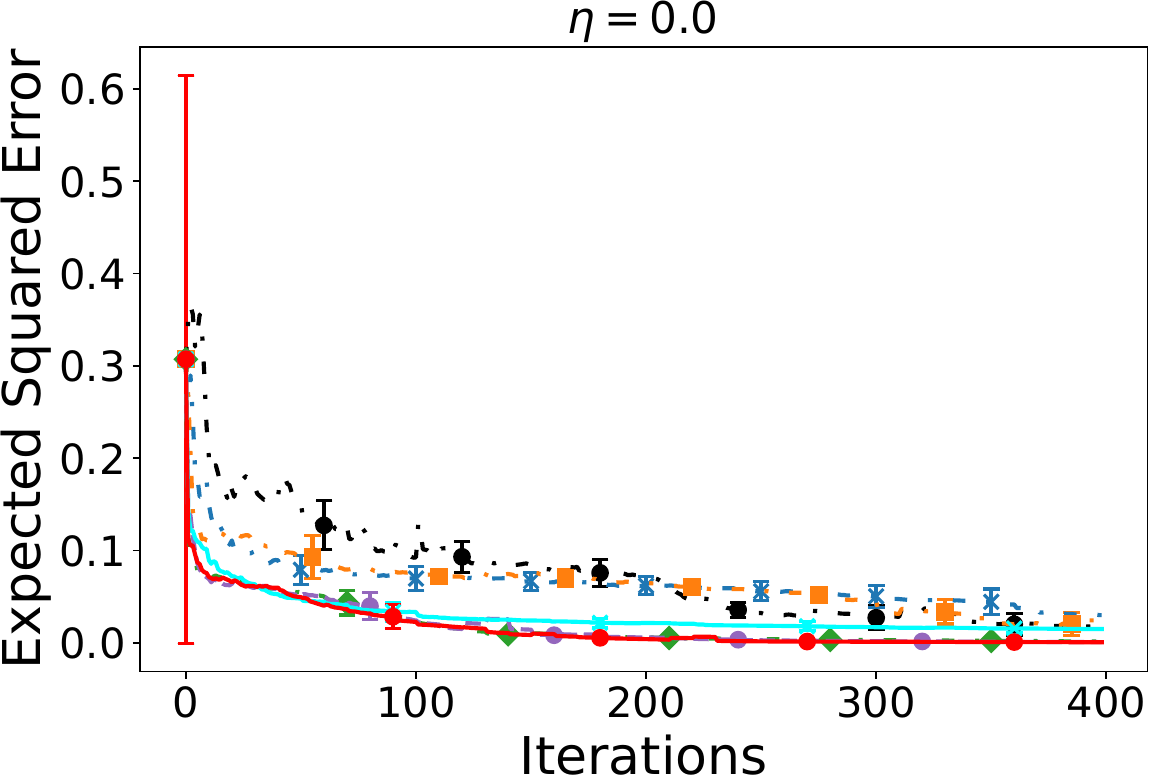}
    \includegraphics[width=0.24\linewidth]{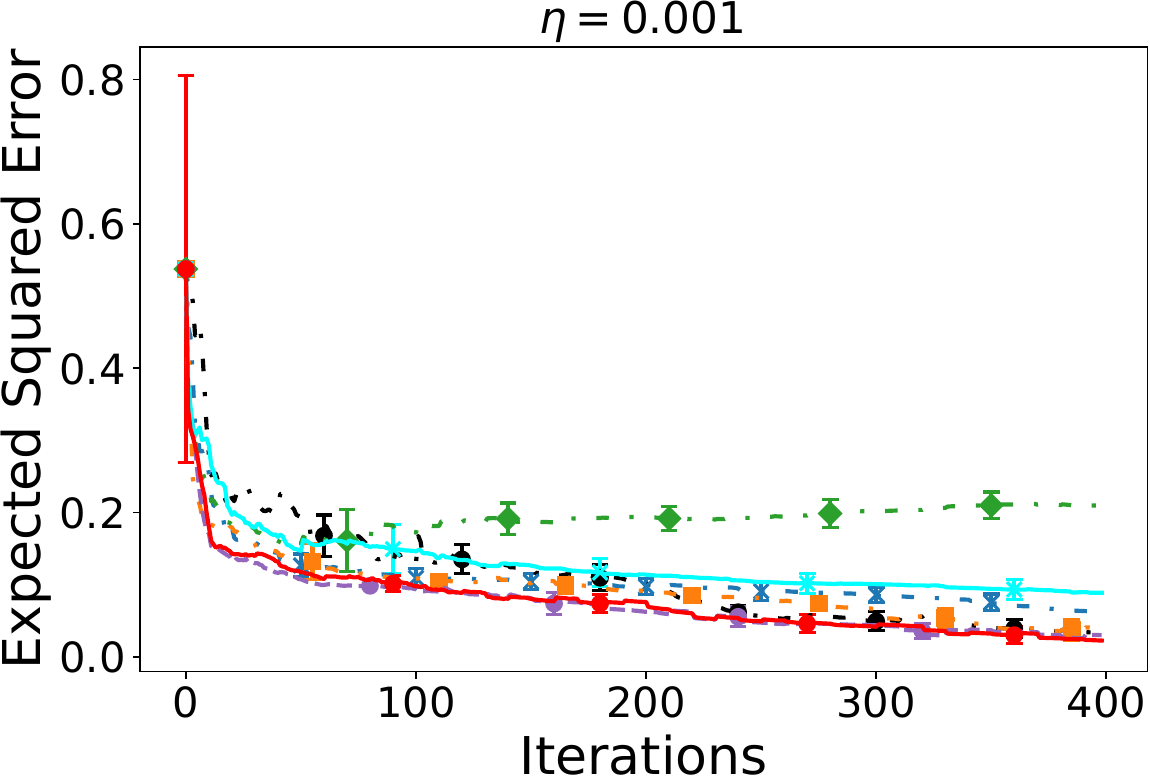}
    \includegraphics[width=0.24\linewidth]{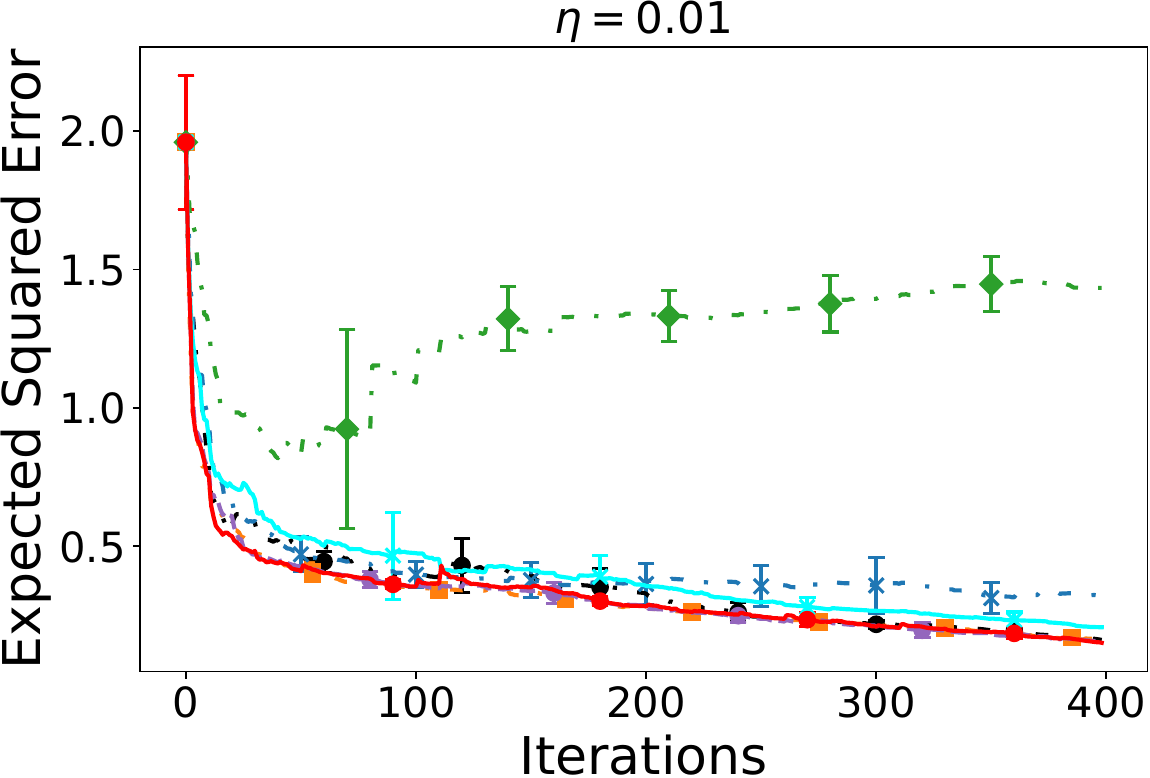}
    \includegraphics[width=0.24\linewidth]{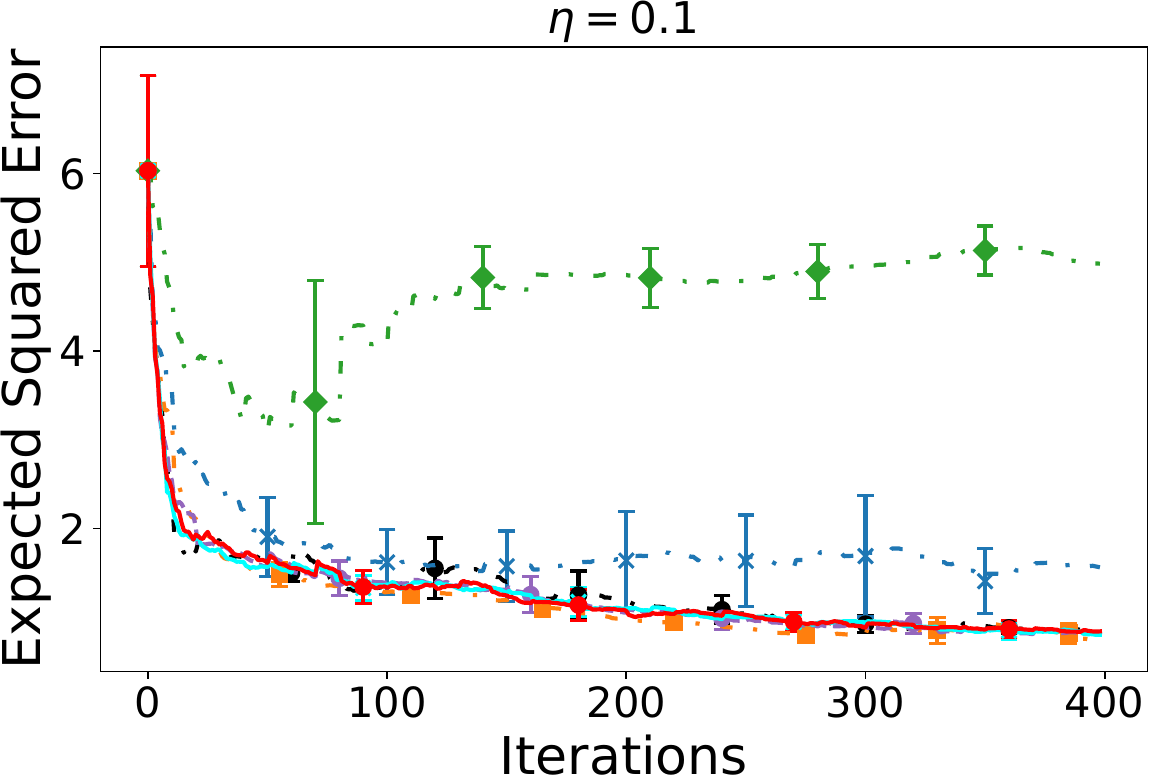}    
    \caption{
        Result of the expected squared error $E_T$ in the real-world data experiments with $\eta=0, 0.001, 0.01, 0.1$.
        The horizontal and vertical axes show the number of iterations and $E_T$, respectively.
        The error bar shows mean and standard errors for 20 random trials regarding the random initial point (and the algorithm's randomness).
        The top, middle, and bottom rows represent the results of the King County house sales dataset, red wine quality dataset, and auto MPG dataset, respectively.
        %
        % 実データ実験結果.
        % %
        % 横軸は反復数, 縦軸は最悪期待二乗誤差を示す.
        % %
        % ランダムに初期点を変更したことによる20回試行の実験結果の平均と標準誤差を示す.
        % %
        % 各段はそれぞれKing County House Sales Dataset, Red Wine Quality Dataset, Auto MPG Datasetを用いた実験結果を示す.
        % %
        % 参照分布は平均が${\* 0}$, 共分散行列が$0.3*{\*I}_d$の$d$次元正規分布とする. 
        % %
        % ここで, $d$は各データセットの入力次元を表し, ${\*I}_d$は$d$次元の単位行列である.
        % %
        % また, 参照分布からの最大距離$\eta=0, 0.001, 0.01, 0.1$とする. 
    }
    \label{fig:real_results}
\end{figure*}

\section{Experiments}
\label{sec:experiment}

In this section, we demonstrate the effectiveness of the proposed methods via synthetic and real-world datasets.
We employ RS, US, variance reduction~\citep{yu2006active}, and expected predictive information gain (EPIG)~\citep{bickford2023-prediction} as the baseline.
Note that we do not employ the method of \citep{frogner2021incorporating} since adapting their method to GPR models is not apparent, and they focus on the ambiguity sets defined by Wasserstein distance over a continuous input domain.
We show the implementation details of EPIG in Appendix~\ref{sec:EPIG}.
Furthermore, as the ablation study, we performed the method, referred to as DR variance reduction, that greedily minimizes $E_{p_t(\*x^*)}\left[ \sigma^2_{t}(\*x^* \mid \*x) \right]$, that is, the unconstrained version of Eq.~\eqref{eq:greedy}.
We referred to the proposed methods as DR random and constrained DR variance reduction (CDR variance reduction).
We evaluate the performance by the error $E_T$ defined in Eq.~\eqref{eq:target_error}.
Furthermore, for the synthetic dataset, we show the result of $\max_{p \in \cP} \EE_{p(\*x)} \left[ \sigma_t^2 (\*x) \mid \cD_t \right]$ in Appendix~\ref{sec:exp_variance}.

In these experiments, by some $\eta \geq 0$, we define the ambiguity set as follows:
\begin{align*}
  \cP = \{p \in \cP_{\cX} \mid \|p_{\rm ref} - p \|_{\infty} \leq \eta \},
\end{align*}
where $\cP_{\cX}$ and $p_{\rm ref}$ are sets of all distributions over $\cX$ and some reference distribution, respectively, and $\|\cdot\|_{\infty}$ denotes $L_{\infty}$ norm.
Note that $\eta = 0$ matches the case that the unique target distribution $p_{\rm ref}$ is specified.
Since we consider the case of discrete $\cX$, maximization over $\cP$ can be written as linear programming for which we used CVXPY~\citep{diamond2016cvxpy,agrawal2018rewriting}.

The aim of our experiments is to show that the proposed algorithms are consistently well-performing for any size of $\cP$, i.e., $\eta$, in contrast to the baselines, which can deteriorate for some $\eta$.
The same parameter $\eta$ is used for the computation of the performance measure $E_T$ and the actual algorithm.
Therefore, we expect that our proposed algorithms show good performance among all $\eta = 0, 0.001, 0.01, 0.1$.
On the other hand, if $\eta$ is large, then $\cP$ starts to include more distributions.
Therefore, the proposed algorithms behave similarly to the US when $\eta$ is large.
Hence, the proposed algorithms have comparable performance to the usual AL methods when $\eta$ is large, though the proposed algorithms are superior to those if $\eta$ is small.
On the other hand, EPIG is designed for the case of $\eta = 0$, that is, the test distribution is explicitly specified.
However, EPIG aims to decrease the entropy $\EE_{p(\*x)} [\log(\sigma_{t-1}(\*x))]$, not $\EE_{p(\*x)} [\sigma^2_{t-1}(\*x)]$.
Therefore, EPIG is not suitable for decreasing $E_T$ and can be inefficient in our experiments even for the case of $\eta = 0$.

% 本章では, 人工データと実データを用いた実験を通じて提案手法の有効性を示す.
% %
% どちらの実験においても比較手法としては, 既存の能動学習手法であるRS, US, variance reduction, EPIGと提案法における推薦候補に対する制限をなくした手法であるvariance reductionを用いる.
% %
% 評価指標として, 人工データ実験では最悪期待損失と最悪期待二乗誤差, 実データ実験では最悪期待二乗誤差のみを用いる.
% %
% ここで, 最悪期待損失は, 以下に定義される$t$反復目における最悪ケースの分布に対する予測分散の期待値とする.
% %
% \begin{align*}
%     {\max_{p^* \in \cP^*}} \EE_{p^*(\*x)} \left[ \sigma_t (\*x) \mid \cD_t \right]
% \end{align*}
% また，最悪期待二乗誤差は, 以下に定義される$t$反復目における最悪ケースの分布に対する二乗誤差の期待値とする．
% %
% \begin{align*}
%     {\max_{p^* \in \cP^*}} \EE_{p^*(\*x)} \left[ (\mu_t(\*x) - f(\*x))^2 \mid \cD_t \right]
% \end{align*}
% %
% このとき, 最悪ケースの分布が含まれる候補分布族$\cP^*$は任意の参照分布から一定の距離$\eta$以内に存在する分布とする. 
% %
% $\cP^*$の構成は次の式で表される. 
% \begin{align*}
%   \cP^* = \{p \in \cP_{\cX} \mid \|p_{\rm ref} - p \|_{\infty} \leq \eta \}
% \end{align*}
% ここで, $\cP_{\cX}$は$\cX$上の任意の分布, $p_{\rm ref}$は参照分布, $\|\cdot\|_{\infty}$は$L^{\infty}$距離を表す. 

\subsection{Synthetic Data Experiemnts}

We set $\cX=\{-1, -0.8, \ldots,  1\}^3$, where $|\cX| = 11^3 = 1331$.
The target function $f$ is the sample path from GPs, where we use SE and Mat\'ern-$\nu$ kernels with $\nu = 5/2$.
We use the fixed hyperparameters of the kernel function in the GPR model, which is used to generate $f$, and fix $\sigma^2 = 10^{-4}$.
The first input $\*x_1$ is selected uniformly at random, and $T$ is set to $400$.
Furthermore, we set $p_{\rm ref} = \cN(\*0, 0.2 \*I_3)$.

% 人工データ実験における候補集合は$\cX=\{-1, -0.8, \ldots,  1\}$とし, $|\cX| = 11^3 = 1331$である.
% %
% 出力にはGP回帰からのサンプルパスを用いる.
% %
% GP回帰のカーネルとして, RBF (Radial Basis Function) カーネルとMat\`ernカーネルを用いる.
% %
% RBFカーネルは$k({\*x}, {\* x}')=\exp(-r)$, Mat\`ernカーネルは$k({\*x}, {\* x}')=(1 + \sqrt{5}r + 5/3r^2)\exp(-\sqrt{5}r)$とし, $r=\| {\*x} - {\*x}' \|^2_2/(2 \times 0.5^2)$とする.
% %
% ノイズ分散は$\sigma^2_n = (10^{-4})^2$とし, カーネルのハイパーパラメータとノイズ分散は固定する.
% %
% 初期点はランダムに1点とり, 実験は400反復まで行う. 

Figure~\ref{fig:syn_results} shows the result.
We can see that DR and CDR variance reductions show superior performance consistently for all the kernel functions and $\eta$, although the DR random is often inferior to those due to the randomness.
This result suggests that the DR and CDR variance reductions effectively incorporate the information of $\cP$.
Furthermore, although the constraint by $\cX_t$ is required for the theoretical analysis in CDR variance reduction, we can confirm that it does not sacrifice the practical effectiveness.
On the other hand, the usual AL methods, such as US and variance reduction, deteriorate when $\eta$ is small since they do not incorporate the information of $\cP$.
When $\eta$ is large, since our problem approaches the worst-case error minimization, the US and variance reduction result in relatively good results.
On the other hand, the EPIG designed for the case $\eta = 0$ is inferior for all $\eta$ since the EPIG is based on the entropy $\cO\left(\log (\sigma_t(\*x))\right)$, not the squared error.
%
% That is, since $\frac{\partial \log a}{\partial a} = \frac{1}{a}$ is large when $a \rightarrow 0$, the EPIG tends to make a small $\sigma_t(\*x)$ more small than making a large $\sigma_t(\*x)$ small.
% %

% 人工データ実験の結果を図\ref{fig:syn_results}と図\ref{fig:syn_loss_results}に示す.
% %
% 図\ref{fig:syn_results}と図\ref{fig:syn_loss_results}では, 縦軸にそれぞれ最悪期待二乗誤差と最悪期待損失, 横軸に反復数を示す.
% %
% 実験結果はランダムに初期点を変更したことによる20回試行の平均と標準誤差を示す.
% %
% 各図の上段はRBFカーネル, 下段はMaternカーネルをGP回帰のカーネルとして用いた場合の結果を示す.
% %
% 参照分布は平均が${\* 0}$, 共分散行列が$0.2*{\*I}_3$の三次元正規分布とする.
% %
% ここで, ${\*I}$は三次元の単位行列である.
% %
% また, 参照分布からの最大距離$\eta=0, 0.001, 0.01, 0.1$とする.

% 図\ref{fig:syn_results}より, カーネルの種類や$\eta$の値に関わらず, 提案法とvariance reductionが最も少ない反復数で最悪期待二乗誤差を小さくしていることが確認できる.
% %
% 提案法とvariance reductionはほぼ同等の性能を示すが, 提案法には理論保証が示されている.
% %
% EPIGは$\eta$が小さい時は最悪期待二乗誤差を小さくすることができているが, $\eta$が大きくなると性能が悪化する.
% %
% これは, EPIGは一意なテスト分布をのみを考慮するためである.
% %
% また, USは$\eta$が小さい場合は提案法と比較して精度が悪いが, $\eta$が大きくなると提案法の性能に近づく.
% %
% これは, $\eta$が大きくなると, 最悪ケースの分布として不確実性の大きいデータに出現確率が偏った分布が計算されるため, 提案法はUSと同様に不確実性の大きなデータを観測しやすくなるからである.
% %

% 図\ref{fig:syn_loss_results}より, カーネルの種類や$\eta$の値に関わらず, 提案法とvariance reductionが最も少ない反復数で最悪期待損失を小さくしていることが確認できる.
% %
% 他の手法の性能に関しても最悪期待二乗誤差を用いて評価を行った場合と同様の傾向が確認できた.

\subsection{Real-World Dataset Experiments}
We use the King County house sales\footnote{\url{https://www.kaggle.com/datasets/harlfoxem/housesalesprediction}}, the red wine quality~\citep{wine_quality_186}, and the auto MPG datasets~\citep{auto_mpg_9} (See Appendix~\ref{sec:datasets_detail} for details).
For all experiments, we used SE kernels, where the hyperparameters $\ell$ and $\sigma^2$ are adaptively determined by the marginal likelihood maximization~\citep{Rasmussen2005-Gaussian} per 10 iterations.
The first input is selected uniformly at random.
Furthermore, we normalize the inputs and outputs of all datasets before the experiments and set $p_{\rm ref} = \cN(\*0, 0.3 \*I_d)$.

% 実データ実験ではKing County House Sales Dataset, Red Wine Quality Dataset, Auto MPG Datasetの3つのデータセットを用いて実験を行った.
% %
% 全ての実験において, GP回帰のカーネルとしてRBFカーネル$k({\*x}, {\* x}')=\sigma^2\exp(-\| {\*x} - {\*x}' \|^2_2/(2 \times l^2)$を用いる.
% %
% カーネルのハイパーパラメータ$\sigma^2$, $l$とノイズ分散$\sigma^2_n$は周辺尤度最大化を用いたハイパーパラメータ最適化によって決定する.
% %
% このハイパーパラメータ最適化は10反復に1回行う.
% %
% 初期点はランダムに一点とり, King County House Sales DatasetとRed Wine Quality Datasetは700反復まで, Auto MPG Datasetでは399反復まで実験を行う.
% %
% また, 扱いやすさのためデータの入力と出力は事前に全データを用いて平均0, 分散1に標準化する. 

Figure~\ref{fig:real_results} shows the result.
We can confirm the same tendency that DR and CDR variance reductions show superior performance consistently, as the synthetic data experiments shown in Figure~\ref{fig:syn_results}.
Note that the fluctuations come from the hyperparameter estimation.
%

% 実データ実験の結果を図\ref{fig:real_results}に示す.
% %
% 縦軸は最悪期待二乗誤差, 横軸は反復数を示す.
% %
% 実験結果はランダムに初期点を変更したことによる20回試行の平均と標準誤差を示す.
% %
% 図\ref{fig:real_results}の各段は上からKing County House Sales Dataset, Red Wine Quality Dataset, Auto MPG Datasetを用いて実験を行った場合の結果を示す.
% %
% 参照分布は平均が${\* 0}$, 共分散行列が$0.3*{\*I}_d$の$d$次元正規分布とする. 
% %
% ここで, $d$は各データセットの入力次元を表し, ${\*I}_d$は$d$次元の単位行列である.
% %
% また, 参照分布からの最大距離$\eta=0, 0.001, 0.01, 0.1$とする. 

% 図\ref{fig:real_results}より, データセットの種類や$\eta$の値に関わらず, 提案法とvariance reductionが最も少ない反復数で最悪期待二乗誤差を小さくしていることが確認できる.
% %
% 提案法とvariance reductionはほぼ同等の性能を示すが, 提案法には理論保証が示されている.
% %
% EPIGは$\eta$が小さい場合は提案法と同等の性能を示すが, $\eta$が大きくなると性能が悪化する.
% %
% また, 人工データと同様に, USは$\eta$が小さい場合は提案法と比較して精度が悪いが, $\eta$が大きくなると提案法の性能に近づく.

% \subsection{Discussion}

% \begin{itemize}
%     \item TODO: Discussion about unknown hyperparameter setting
%     \item DRAL is the middle framework between target distribution aware AL and worst case prediction $\max {\rm error}$.
%     \item TODO: EPIG aims to reduce $\log \sigma_t$. So, EPIG often concentrates on reducing the posterior variance of a certain point rather than reducing the posterior variance over the whole input space. This is just a difference in the goal of the algorithm. Ours is like V-optimal, but EPIG is like D-optimality.
% \end{itemize}
\section{Conclusion}
\label{sec:conclusion}

This paper investigated the DRAL problem for the GPR, in which we aim to reduce the worst-case error $E_T$.
We first showed several properties of this problem for the GPR, which implies that minimizing the variance guarantees a decrease in $E_T$. 
Therefore, we designed two algorithms that reduce the target variance and incorporate information about target distribution candidates for practical effectiveness.
Then, we proved the theoretical error convergence of the proposed methods, whose practical effectiveness is demonstrated via synthetic and real-world datasets.

\paragraph{Limitation and Future Work:}
We can consider several future research directions.
First, since we do not show the optimality of the convergence rate, developing a (near) optimal algorithm for $E_T$ is vital.
For this goal, the approximate submodularity~\citep{bian2017guarantees} may be relevant from the empirical superiority of DR variance reduction.
Second, since the expectation over $p(\*x^*)$ may be intractable, an analysis incorporating the approximation error or developing an efficient algorithm without expectation computation may be crucial (DR random does not require the expectation but is often inefficient).
Third, although our analyses only require the existence of the maximum over $\cP$, our experiments are limited to the discrete distribution set defined by the $L_{\infty}$ ball.
Thus, more general experiments regarding, e.g., the continuous probability distributions and the ambiguity sets defined by Kullback-Leibler divergence~\citep{hu2013kullback} and Wasserstein distance~\citep{frogner2021incorporating}, are interesting from the practical perspective.

% The following are important future work: (i) developing (near) optimal algorithm for $E_T$, in which approximate submodularity~\citep{bian2017guarantees} may be relevant from the empirical superiority of the approximate greedy algorithm, (ii) more general experiments regarding the ambiguity set $\cP$, such as the continuous probability distributions and the ball defined by Kullback-Leibler divergence~\citep{hu2013kullback} and Wasserstein distance~\citep{frogner2021incorporating}.

% Although we show the upper bounds of the error, we do not show the optimality of our analysis.
% %
% Developing (near) optimal algorithms is an important future direction.
% %
% In this direction, although our error is not submodular in general, the analysis for approximate submodular functions~\citep{bian2017guarantees} may be interesting, particularly from the empirical superiority of the approximate greedy algorithm in the experiments.
% %
% In addition, although our assumption on the ambiguity set is only that the maximum exists and can be computed, our empirical evaluation is limited to the ambiguity sets defined by $L_{\infty}$ norm.
% %
% Experiments with other ambiguity sets, e.g., using Kullback-Leibler divergence~\citep{hu2013kullback}, are of interest, although this paper mainly concentrates on the theoretical perspective.

\section*{Acknowkedgements}
This work was partially supported by 
JST ACT-X Grant Number (JPMJAX23CD and JPMJAX24C3), 
JST PRESTO Grant Number JPMJPR24J6, 
JST CREST Grant Numbers (JPMJCR21D3 including AIP challenge program and JPMJCR22N2),
JST Moonshot R\&D Grant Number JPMJMS2033-05, 
%%%%%%%
JSPS KAKENHI Grant Number (JP20H00601, JP23K16943, JP23K19967, JP24K15080, and JP24K20847), 
%%%%%%%
NEDO (JPNP20006),
and RIKEN Center for Advanced Intelligence Project.

% MEXT Program: Data Creation and Utilization-Type Material Research and Development Project Grant Number JPMXP1122712807,
% JSPS KAKENHI Grant Numbers JP20H00601,JP21H03498,JP22H00300,JP23K16943,JP23K19967,
% JST CREST Grant Numbers JPMJCR21D3,JPMJCR22N2, 

% JST ACT-X Grant Number JPMJAX23CD,
% NEDO Grant Numbers JPNP18002,JPNP20006, 
% and RIKEN Center for Advanced Intelligence Project.

%%%%%%%%%%%%%%%%%%%%%%%%%%%%%%%%%%%%%%%%%%%%%%%%%%%%%%%%%%%%%%%%%%%%%%%%%%%%%%%%%%%%%%%%%%%%%%
% \clearpage

% \section*{Impact Statements}

% This paper focuses on the theoretical and algorithmic aspects of the machine learning method.
% %
% We consider that some potential societal consequences of this paper need not necessarily be highlighted here.

%%%%%%%%%%%%%%%%%%%%%%%%%%%%%%%%%%%%%%%%%%%%%%%%%%%%%%%%%%%%%%%%%%%%%%%%%%%%%%%%%%%%%%%%%%%%%%
% % \subsubsection*{References}
\bibliography{ref}

\begin{thebibliography}{68}
\providecommand{\natexlab}[1]{#1}
\providecommand{\url}[1]{\texttt{#1}}
\expandafter\ifx\csname urlstyle\endcsname\relax
  \providecommand{\doi}[1]{doi: #1}\else
  \providecommand{\doi}{doi: \begingroup \urlstyle{rm}\Url}\fi

\bibitem[Abbasi-Yadkori(2013)]{abbasi2013online}
Yasin Abbasi-Yadkori.
\newblock \emph{Online learning for linearly parametrized control problems}.
\newblock PhD thesis, University of Alberta, 2013.

\bibitem[Adler(1981)]{adler1981geometry}
Robert~J Adler.
\newblock \emph{The Geometry of Random Fields}, volume~62.
\newblock SIAM, 1981.

\bibitem[Agrawal et~al.(2018)Agrawal, Verschueren, Diamond, and Boyd]{agrawal2018rewriting}
Akshay Agrawal, Robin Verschueren, Steven Diamond, and Stephen Boyd.
\newblock A rewriting system for convex optimization problems.
\newblock \emph{Journal of Control and Decision}, 5\penalty0 (1):\penalty0 42--60, 2018.

\bibitem[Allen-Zhu et~al.(2017)Allen-Zhu, Li, Singh, and Wang]{zhu2017near}
Zeyuan Allen-Zhu, Yuanzhi Li, Aarti Singh, and Yining Wang.
\newblock Near-optimal design of experiments via regret minimization.
\newblock In \emph{Proceedings of the 34th International Conference on Machine Learning}, volume~70 of \emph{Proceedings of Machine Learning Research}, pages 126--135. PMLR, 2017.

\bibitem[Bian et~al.(2017)Bian, Buhmann, Krause, and Tschiatschek]{bian2017guarantees}
Andrew~An Bian, Joachim~M. Buhmann, Andreas Krause, and Sebastian Tschiatschek.
\newblock Guarantees for greedy maximization of non-submodular functions with applications.
\newblock In \emph{Proceedings of the 34th International Conference on Machine Learning}, volume~70 of \emph{Proceedings of Machine Learning Research}, pages 498--507. PMLR, 2017.

\bibitem[Bickford~Smith et~al.(2023)Bickford~Smith, Kirsch, Farquhar, Gal, Foster, and Rainforth]{bickford2023-prediction}
Freddie Bickford~Smith, Andreas Kirsch, Sebastian Farquhar, Yarin Gal, Adam Foster, and Tom Rainforth.
\newblock Prediction-oriented {Bayesian} active learning.
\newblock In \emph{Proceedings of The 26th International Conference on Artificial Intelligence and Statistics}, volume 206 of \emph{Proceedings of Machine Learning Research}, pages 7331--7348. PMLR, 2023.

\bibitem[Bogunovic et~al.(2016)Bogunovic, Scarlett, Krause, and Cevher]{bogunovic2016truncated}
Ilija Bogunovic, Jonathan Scarlett, Andreas Krause, and Volkan Cevher.
\newblock Truncated variance reduction: A unified approach to {B}ayesian optimization and level-set estimation.
\newblock In \emph{Advances in neural information processing systems 29}, pages 1507--1515. Curran Associates, Inc., 2016.

\bibitem[Chaloner and Verdinelli(1995)]{Chaloner1995bayesian}
Kathryn Chaloner and Isabella Verdinelli.
\newblock {Bayesian Experimental Design: A Review}.
\newblock \emph{Statistical Science}, 10\penalty0 (3):\penalty0 273 -- 304, 1995.

\bibitem[Chen and Paschalidis(2018)]{chen2018robust}
Ruidi Chen and Ioannis~Ch. Paschalidis.
\newblock A robust learning approach for regression models based on distributionally robust optimization.
\newblock \emph{Journal of Machine Learning Research}, 19\penalty0 (13):\penalty0 1--48, 2018.

\bibitem[Chen et~al.(2020)Chen, Paschalidis, et~al.]{chen2020distributionally}
Ruidi Chen, Ioannis~Ch Paschalidis, et~al.
\newblock Distributionally robust learning.
\newblock \emph{Foundations and Trends{\textregistered} in Optimization}, 4\penalty0 (1-2):\penalty0 1--243, 2020.

\bibitem[Chowdhury and Gopalan(2017)]{Chowdhury2017-on}
Sayak~Ray Chowdhury and Aditya Gopalan.
\newblock On kernelized multi-armed bandits.
\newblock In \emph{Proceedings of the 34th International Conference on Machine Learning}, volume~70 of \emph{Proceedings of Machine Learning Research}, pages 844--853, 2017.

\bibitem[Cohn(1993)]{cohn1993neural}
David Cohn.
\newblock Neural network exploration using optimal experiment design.
\newblock \emph{Advances in neural information processing systems}, 6, 1993.

\bibitem[Cohn et~al.(1996)Cohn, Ghahramani, and Jordan]{cohn1996active}
David~A Cohn, Zoubin Ghahramani, and Michael~I Jordan.
\newblock Active learning with statistical models.
\newblock \emph{Journal of artificial intelligence research}, 4:\penalty0 129--145, 1996.

\bibitem[Cortez and Reis(2009)]{wine_quality_186}
Cerdeira A. Almeida F. Matos~T. Cortez, Paulo and J.~Reis.
\newblock {Wine Quality}.
\newblock UCI Machine Learning Repository, 2009.

\bibitem[Cortez et~al.(2009)Cortez, Teixeira, Cerdeira, Almeida, Matos, and Reis]{cortez2009using}
Paulo Cortez, Juliana Teixeira, Ant{\'o}nio Cerdeira, Fernando Almeida, Telmo Matos, and Jos{\'e} Reis.
\newblock Using data mining for wine quality assessment.
\newblock In \emph{Discovery Science: 12th International Conference}, pages 66--79. Springer, 2009.

\bibitem[Costa et~al.(2024)Costa, Pf^^c3^^b6rtner, Costa, and Hennig]{dacosta2024samplep}
Natha^^c3^^abl~Da Costa, Marvin Pf^^c3^^b6rtner, Lancelot~Da Costa, and Philipp Hennig.
\newblock Sample path regularity of {G}aussian processes from the covariance kernel, 2024.

\bibitem[Das and Kempe(2008)]{das2008algorithms}
Abhimanyu Das and David Kempe.
\newblock Algorithms for subset selection in linear regression.
\newblock In \emph{Proceedings of the Fortieth Annual ACM Symposium on Theory of Computing}, STOC '08, page 45^^e2^^80^^9354. Association for Computing Machinery, 2008.

\bibitem[De~Freitas et~al.(2012)De~Freitas, Smola, and Zoghi]{freitas2012exponential}
Nando De~Freitas, Alex~J. Smola, and Masrour Zoghi.
\newblock Exponential regret bounds for {G}aussian process bandits with deterministic observations.
\newblock In \emph{Proceedings of the 29th International Conference on International Conference on Machine Learning}, page 955^^e2^^80^^93962. Omnipress, 2012.

\bibitem[Diamond and Boyd(2016)]{diamond2016cvxpy}
Steven Diamond and Stephen Boyd.
\newblock {CVXPY}: {A} {P}ython-embedded modeling language for convex optimization.
\newblock \emph{Journal of Machine Learning Research}, 17\penalty0 (83):\penalty0 1--5, 2016.

\bibitem[Frogner et~al.(2021)Frogner, Claici, Chien, and Solomon]{frogner2021incorporating}
Charlie Frogner, Sebastian Claici, Edward Chien, and Justin Solomon.
\newblock Incorporating unlabeled data into distributionally robust learning.
\newblock \emph{Journal of Machine Learning Research}, 22\penalty0 (56):\penalty0 1--46, 2021.

\bibitem[Ghosal and Roy(2006)]{Ghosal2006-posterior}
Subhashis Ghosal and Anindya Roy.
\newblock {Posterior consistency of {G}aussian process prior for nonparametric binary regression}.
\newblock \emph{The Annals of Statistics}, 34\penalty0 (5):\penalty0 2413 -- 2429, 2006.

\bibitem[Gotovos et~al.(2013)Gotovos, Casati, Hitz, and Krause]{gotovos2013active}
Alkis Gotovos, Nathalie Casati, Gregory Hitz, and Andreas Krause.
\newblock Active learning for level set estimation.
\newblock In \emph{Proceedings of the Twenty-Third international joint conference on Artificial Intelligence}, pages 1344--1350, 2013.

\bibitem[Guestrin et~al.(2005)Guestrin, Krause, and Singh]{Guestrin2005-near}
Carlos Guestrin, Andreas Krause, and Ajit~Paul Singh.
\newblock Near-optimal sensor placements in {Gaussian} processes.
\newblock In \emph{Proceedings of the 22nd International Conference on Machine Learning}, page 265^^e2^^80^^93272. Association for Computing Machinery, 2005.

\bibitem[Handel(2016)]{Handel-HDP}
Ramon~Van Handel.
\newblock Probability in high dimension, 2016.
\newblock Lecture notes. Available in \url{https://web.math.princeton.edu/~rvan/APC550.pdf}.

\bibitem[Hanneke et~al.(2014)]{hanneke2014theory}
Steve Hanneke et~al.
\newblock Theory of disagreement-based active learning.
\newblock \emph{Foundations and Trends{\textregistered} in Machine Learning}, 7\penalty0 (2-3):\penalty0 131--309, 2014.

\bibitem[Hoang et~al.(2014)Hoang, Low, Jaillet, and Kankanhalli]{hoang2014nonmyopic}
Trong~Nghia Hoang, Bryan Kian~Hsiang Low, Patrick Jaillet, and Mohan Kankanhalli.
\newblock Nonmyopic $\varepsilon$-{B}ayes-optimal active learning of {G}aussian processes.
\newblock In \emph{International conference on machine learning}, pages 739--747. PMLR, 2014.

\bibitem[Houlsby et~al.(2011)Houlsby, Husz{\'a}r, Ghahramani, and Lengyel]{houlsby2011bayesian}
Neil Houlsby, Ferenc Husz{\'a}r, Zoubin Ghahramani, and M{\'a}t{\'e} Lengyel.
\newblock {B}ayesian active learning for classification and preference learning.
\newblock \emph{arXiv preprint arXiv:1112.5745}, 2011.

\bibitem[Hu and Hong(2013)]{hu2013kullback}
Zhaolin Hu and L~Jeff Hong.
\newblock Kullback-leibler divergence constrained distributionally robust optimization.
\newblock \emph{Available at Optimization Online}, 1\penalty0 (2):\penalty0 9, 2013.

\bibitem[H\"{u}botter et~al.(2024)H\"{u}botter, Sukhija, Treven, As, and Krause]{hubotter2024-transductive}
Jonas H\"{u}botter, Bhavya Sukhija, Lenart Treven, Yarden As, and Andreas Krause.
\newblock Transductive active learning: Theory and applications.
\newblock In \emph{Advances in Neural Information Processing Systems}, volume~37, pages 124686--124755. Curran Associates, Inc., 2024.

\bibitem[Inatsu et~al.(2021)Inatsu, Iwazaki, and Takeuchi]{inatsu2021active}
Yu~Inatsu, Shogo Iwazaki, and Ichiro Takeuchi.
\newblock Active learning for distributionally robust level-set estimation.
\newblock In \emph{Proceedings of the 38th International Conference on Machine Learning}, volume 139 of \emph{Proceedings of Machine Learning Research}, pages 4574--4584. PMLR, 2021.

\bibitem[Inatsu et~al.(2024)Inatsu, Takeno, Kutsukake, and Takeuchi]{inatsu2024active}
Yu~Inatsu, Shion Takeno, Kentaro Kutsukake, and Ichiro Takeuchi.
\newblock Active learning for level set estimation using randomized straddle algorithms.
\newblock \emph{Transactions on Machine Learning Research}, 2024.

\bibitem[Kanagawa et~al.(2018)Kanagawa, Hennig, Sejdinovic, and Sriperumbudur]{kanagawa2018gaussian}
Motonobu Kanagawa, Philipp Hennig, Dino Sejdinovic, and Bharath~K Sriperumbudur.
\newblock {G}aussian processes and kernel methods: A review on connections and equivalences.
\newblock \emph{arXiv:1807.02582}, 2018.

\bibitem[Kandasamy et~al.(2018)Kandasamy, Krishnamurthy, Schneider, and P{\'o}czos]{Kandasamy2018-Parallelised}
Kirthevasan Kandasamy, Akshay Krishnamurthy, Jeff Schneider, and Barnabas P{\'o}czos.
\newblock Parallelised {B}ayesian optimisation via {T}hompson sampling.
\newblock In \emph{Proceedings of the 21st International Conference on Artificial Intelligence and Statistics}, volume~84 of \emph{Proceedings of Machine Learning Research}, pages 133--142, 2018.

\bibitem[Kirsch and Gal(2022)]{kirsch2022unifying}
Andreas Kirsch and Yarin Gal.
\newblock Unifying approaches in active learning and active sampling via {Fisher} information and information-theoretic quantities.
\newblock \emph{Transactions on Machine Learning Research}, 2022.
\newblock Expert Certification.

\bibitem[Kirsch et~al.(2021)Kirsch, Rainforth, and Gal]{kirsch2021test}
Andreas Kirsch, Tom Rainforth, and Yarin Gal.
\newblock Test distribution-aware active learning: A principled approach against distribution shift and outliers.
\newblock \emph{arXiv:2106.11719}, 2021.

\bibitem[Kirschner and Krause(2018)]{kirschner2018-information}
Johannes Kirschner and Andreas Krause.
\newblock Information directed sampling and bandits with heteroscedastic noise.
\newblock In \emph{Proceedings of the 31st Conference On Learning Theory}, volume~75 of \emph{Proceedings of Machine Learning Research}, pages 358--384. PMLR, 2018.

\bibitem[Krause et~al.(2008{\natexlab{a}})Krause, McMahan, Guestrin, and Gupta]{krause2008robust}
Andreas Krause, H.~Brendan McMahan, Carlos Guestrin, and Anupam Gupta.
\newblock Robust submodular observation selection.
\newblock \emph{Journal of Machine Learning Research}, 9\penalty0 (93):\penalty0 2761--2801, 2008{\natexlab{a}}.

\bibitem[Krause et~al.(2008{\natexlab{b}})Krause, Singh, and Guestrin]{krause2008-near}
Andreas Krause, Ajit Singh, and Carlos Guestrin.
\newblock Near-optimal sensor placements in {Gaussian} processes: Theory, efficient algorithms and empirical studies.
\newblock \emph{J. Mach. Learn. Res.}, 9:\penalty0 235^^e2^^80^^93284, 2008{\natexlab{b}}.

\bibitem[Kusakawa et~al.(2022)Kusakawa, Takeno, Inatsu, Kutsukake, Iwazaki, Nakano, Ujihara, Karasuyama, and Takeuchi]{Kusakawa2022-bayesian}
Shunya Kusakawa, Shion Takeno, Yu~Inatsu, Kentaro Kutsukake, Shogo Iwazaki, Takashi Nakano, Toru Ujihara, Masayuki Karasuyama, and Ichiro Takeuchi.
\newblock {B}ayesian optimization for cascade-type multistage processes.
\newblock \emph{Neural Computation}, 34\penalty0 (12):\penalty0 2408--2431, 2022.

\bibitem[Kushner(1964)]{Kushner1964-new}
H.~J. Kushner.
\newblock {A New Method of Locating the Maximum Point of an Arbitrary Multipeak Curve in the Presence of Noise}.
\newblock \emph{Journal of Basic Engineering}, 86\penalty0 (1):\penalty0 97--106, 1964.

\bibitem[Li and Scarlett(2022)]{li2022gaussian}
Zihan Li and Jonathan Scarlett.
\newblock {G}aussian process bandit optimization with few batches.
\newblock In \emph{Proceedings of The 25th International Conference on Artificial Intelligence and Statistics}, volume 151 of \emph{Proceedings of Machine Learning Research}, pages 92--107. PMLR, 2022.

\bibitem[Lindley(1956)]{lindley1956-on}
D.~V. Lindley.
\newblock {On a Measure of the Information Provided by an Experiment}.
\newblock \emph{The Annals of Mathematical Statistics}, 27\penalty0 (4):\penalty0 986 -- 1005, 1956.

\bibitem[Liu et~al.(2015)Liu, Reyzin, and Ziebart]{Liu2015shift}
Anqi Liu, Lev Reyzin, and Brian Ziebart.
\newblock Shift-pessimistic active learning using robust bias-aware prediction.
\newblock \emph{Proceedings of the AAAI Conference on Artificial Intelligence}, 29\penalty0 (1), 2015.

\bibitem[Liu et~al.(2020)Liu, Ong, Shen, and Cai]{liu2020when}
Haitao Liu, Yew-Soon Ong, Xiaobo Shen, and Jianfei Cai.
\newblock When {G}aussian process meets big data: A review of scalable {GPs}.
\newblock \emph{IEEE Transactions on Neural Networks and Learning Systems}, 31\penalty0 (11):\penalty0 4405--4423, 2020.

\bibitem[Paria et~al.(2020)Paria, Kandasamy, and P{\'{o}}czos]{paria2020-flexible}
Biswajit Paria, Kirthevasan Kandasamy, and Barnab{\'{a}}s P{\'{o}}czos.
\newblock A flexible framework for multi-objective {B}ayesian optimization using random scalarizations.
\newblock In \emph{Proceedings of The 35th Uncertainty in Artificial Intelligence Conference}, volume 115 of \emph{Proceedings of Machine Learning Research}, pages 766--776, 2020.

\bibitem[Park et~al.(2020)Park, Lee, Hwang, and Kim]{park2020additive}
Minyoung Park, Seungyeon Lee, Sangheum Hwang, and Dohyun Kim.
\newblock Additive ensemble neural networks.
\newblock \emph{IEEE Access}, 8:\penalty0 113192--113199, 2020.

\bibitem[Park and Kim(2020)]{park2020robust}
Sung~Ho Park and Seoung~Bum Kim.
\newblock Robust expected model change for active learning in regression.
\newblock \emph{Applied Intelligence}, 50:\penalty0 296--313, 2020.

\bibitem[Pukelsheim(2006)]{pukelsheim2006optimal}
Friedrich Pukelsheim.
\newblock \emph{Optimal design of experiments}.
\newblock SIAM, 2006.

\bibitem[Quinlan(1993)]{auto_mpg_9}
R.~Quinlan.
\newblock {Auto MPG}.
\newblock UCI Machine Learning Repository, 1993.
\newblock {DOI}: https://doi.org/10.24432/C5859H.

\bibitem[Rasmussen and Williams(2005)]{Rasmussen2005-Gaussian}
Carl~Edward Rasmussen and Christopher K.~I. Williams.
\newblock \emph{Gaussian Processes for Machine Learning (Adaptive Computation and Machine Learning)}.
\newblock The MIT Press, 2005.

\bibitem[Ryan and Morgan(2007)]{ryan2007modern}
Thomas~P Ryan and JP~Morgan.
\newblock Modern experimental design.
\newblock \emph{Journal of Statistical Theory and Practice}, 1\penalty0 (3-4):\penalty0 501--506, 2007.

\bibitem[Salgia et~al.(2024)Salgia, Vakili, and Zhao]{salgia2024random}
Sudeep Salgia, Sattar Vakili, and Qing Zhao.
\newblock Random exploration in {B}ayesian optimization: Order-optimal regret and computational efficiency.
\newblock In \emph{Proceedings of the 41st International Conference on Machine Learning}, volume 235 of \emph{Proceedings of Machine Learning Research}, pages 43112--43141. PMLR, 2024.

\bibitem[Sener and Savarese(2018)]{sener2018active}
Ozan Sener and Silvio Savarese.
\newblock Active learning for convolutional neural networks: A core-set approach.
\newblock In \emph{International Conference on Learning Representations}, 2018.

\bibitem[Seo et~al.(2000)Seo, Wallat, Graepel, and Obermayer]{Seo2000gaussian}
Sambu Seo, M.~Wallat, T.~Graepel, and K.~Obermayer.
\newblock {G}aussian process regression: active data selection and test point rejection.
\newblock In \emph{Proceedings of the IEEE-INNS-ENNS International Joint Conference on Neural Networks. IJCNN 2000. Neural Computing: New Challenges and Perspectives for the New Millennium}, volume~3, pages 241--246, 2000.

\bibitem[Settles(2009)]{settles2009-active}
Burr Settles.
\newblock Active learning literature survey.
\newblock Computer Sciences Technical Report 1648, University of Wisconsin--Madison, 2009.

\bibitem[Shahriari et~al.(2016)Shahriari, Swersky, Wang, Adams, and {De Freitas}]{Shahriari2016-Taking}
Bobak Shahriari, Kevin Swersky, Ziyu Wang, {Ryan P.} Adams, and Nando {De Freitas}.
\newblock Taking the human out of the loop: A review of {B}ayesian optimization.
\newblock \emph{Proceedings of the IEEE}, 104\penalty0 (1):\penalty0 148--175, 2016.

\bibitem[Shoham and Avron(2023)]{Shoham2023experimental}
Neta Shoham and Haim Avron.
\newblock { Experimental Design for Overparameterized Learning With Application to Single Shot Deep Active Learning }.
\newblock \emph{IEEE Transactions on Pattern Analysis \& Machine Intelligence}, 45\penalty0 (10):\penalty0 11766--11777, 2023.

\bibitem[Srinivas et~al.(2010)Srinivas, Krause, Kakade, and Seeger]{Srinivas2010-Gaussian}
N.~Srinivas, A.~Krause, S.~Kakade, and M.~Seeger.
\newblock Gaussian process optimization in the bandit setting: No regret and experimental design.
\newblock In \emph{Proceedings of the 27th International Conference on Machine Learning}, pages 1015--1022. Omnipress, 2010.

\bibitem[Staib et~al.(2019)Staib, Wilder, and Jegelka]{staib2019distributionally}
Matthew Staib, Bryan Wilder, and Stefanie Jegelka.
\newblock Distributionally robust submodular maximization.
\newblock In \emph{Proceedings of the Twenty-Second International Conference on Artificial Intelligence and Statistics}, volume~89 of \emph{Proceedings of Machine Learning Research}, pages 506--516. PMLR, 2019.

\bibitem[Sugiyama(2005)]{sugiyama2005active}
Masashi Sugiyama.
\newblock Active learning for misspecified models.
\newblock In \emph{Advances in Neural Information Processing Systems}, volume~18, pages 1305--1312. MIT Press, 2005.

\bibitem[Takeno et~al.(2023)Takeno, Inatsu, and Karasuyama]{Takeno2023-randomized}
Shion Takeno, Yu~Inatsu, and Masayuki Karasuyama.
\newblock Randomized {G}aussian process upper confidence bound with tighter {B}ayesian regret bounds.
\newblock In \emph{Proceedings of the 40th International Conference on Machine Learning}, volume 202 of \emph{Proceedings of Machine Learning Research}, pages 33490--33515. PMLR, 2023.

\bibitem[Takeno et~al.(2024)Takeno, Inatsu, Karasuyama, and Takeuchi]{takeno2024-posterior}
Shion Takeno, Yu~Inatsu, Masayuki Karasuyama, and Ichiro Takeuchi.
\newblock Posterior sampling-based {B}ayesian optimization with tighter {B}ayesian regret bounds.
\newblock In \emph{Proceedings of the 41st International Conference on Machine Learning}, volume 235 of \emph{Proceedings of Machine Learning Research}, pages 47510--47534. PMLR, 2024.

\bibitem[Vakili et~al.(2021{\natexlab{a}})Vakili, Bouziani, Jalali, Bernacchia, and Shiu]{vakili2021-optimal}
Sattar Vakili, Nacime Bouziani, Sepehr Jalali, Alberto Bernacchia, and Da-shan Shiu.
\newblock Optimal order simple regret for {G}aussian process bandits.
\newblock In \emph{Advances in Neural Information Processing Systems}, volume~34, pages 21202--21215. Curran Associates, Inc., 2021{\natexlab{a}}.

\bibitem[Vakili et~al.(2021{\natexlab{b}})Vakili, Khezeli, and Picheny]{vakili2021-information}
Sattar Vakili, Kia Khezeli, and Victor Picheny.
\newblock On information gain and regret bounds in {G}aussian process bandits.
\newblock In \emph{Proceedings of The 24th International Conference on Artificial Intelligence and Statistics}, volume 130 of \emph{Proceedings of Machine Learning Research}, pages 82--90, 2021{\natexlab{b}}.

\bibitem[Vakili et~al.(2022)Vakili, Scarlett, Shiu, and Bernacchia]{vakili2022improved}
Sattar Vakili, Jonathan Scarlett, Da-shan Shiu, and Alberto Bernacchia.
\newblock Improved convergence rates for sparse approximation methods in kernel-based learning.
\newblock In \emph{International Conference on Machine Learning}, pages 21960--21983. PMLR, 2022.

\bibitem[van~der Vaart and Wellner(1996)]{van1996weak}
Aad van~der Vaart and Jon~A Wellner.
\newblock \emph{Weak Convergence and Empirical Processes: With Applications to Statistics}.
\newblock Springer Science \& Business Media, 1996.

\bibitem[Yu et~al.(2006)Yu, Bi, and Tresp]{yu2006active}
Kai Yu, Jinbo Bi, and Volker Tresp.
\newblock Active learning via transductive experimental design.
\newblock In \emph{Proceedings of the 23rd international conference on Machine learning}, pages 1081--1088, 2006.

\bibitem[Zhao et~al.(2021)Zhao, Liu, Anandkumar, and Yue]{zhao2021active}
Eric Zhao, Anqi Liu, Animashree Anandkumar, and Yisong Yue.
\newblock Active learning under label shift.
\newblock In \emph{International conference on artificial intelligence and statistics}, pages 3412--3420. PMLR, 2021.

\end{thebibliography}
\bibliographystyle{plainnat}
% %%%%%%%%%%%%%%%%%%%%%%%%%%%%%%%%%%%%%%%%%%%%%%%%%%%%%%%%%%%%%%%%%%%%%%%%%%%%%%%%%%%%%%%%%%%%%%

%%%%%%%%%%%%%%%%%%%%%%%%%%%%%%%%%%%%%%%%%%%%%%%%%%%%%%%%%%%%%%%%%%%%%%%%%%%%%%%%%%%%%%%%%%%%%%
%%%%%%%%%%%%%%%%%%%%%%%%%%%%%%%%%%%%%%%%%%%%%%%%%%%%%%%%%%%%%%%%%%%%%%%%%%%
\clearpage
\appendix
\onecolumn
%%%%%%%%%%%%%%%%%%%%%%%%%%%%%%%%%%%%%%%%%%%%%%%%%%%%%%%%%%%%%%%%%%%%%%%%%%%
\section{Proofs for Section~\ref{sec:problem}}
\label{sec:problem_proof}

%%%%%%%%%%%%%%%%%%%%%%%%%%%%%%%%%%%%%%%%%%%%%%%%%%%%%%%%%%%%%%%%%%%%%%%%%%%
\subsection{Proof of Lemma~\ref{lem:lipschitz_posterior_mean}}
\label{sec:proof_lipschitz_posterior_mean}

From the definition of $\mu_t$, we obtain
\begin{align*}
    \mu_t (\cdot)
    &\leq \*v_t^\top (\cdot) \*y_t  \\
    &\leq \*v_t^\top (\cdot) \*f_t  + \*v_t^\top (\cdot) \*\epsilon_t,
\end{align*}
where $\*v_t (\*x) = \left(\*k_t^\top(\*x) (\*K_t + \sigma^2 \*I_t)^{-1}\right)^\top$, $\*f_t = \bigl(f(\*x_1), \dots, f(\*x_t) \bigr)^\top$, and $\*\epsilon_t = (\epsilon_1, \dots, \epsilon_t)^\top$.
Therefore, the Lipschitz constant of $\mu_t$ is bounded from above by the Lipscthiz constants of $\*v_t^\top (\cdot) \*f_t$ and $\*v_t^\top (\cdot) \*\epsilon_t$.

For the first term $\*v_t^\top (\cdot) \*f_t$, we follow the proof of Lemma~4 of \citep{vakili2021-optimal}.
Recall the RKHS-based definition of kernel ridge estimator:
\begin{align*}
    \mu_t = \argmin_{\mu \in \cH_k} \sum_{i=1}^t \bigl( y_{\*x_i} - \mu(\*x_i) \bigr)^2 + \sigma^2 \| \mu \|_{\cH_k}.
\end{align*}
Therefore, we can derive
\begin{align*}
    \min_{\mu \in \cH_k} \sum_{i=1}^t \bigl( f(\*x_i) - \mu(\*x_i) \bigr)^2 + \sigma^2 \| \mu \|_{\cH_k} 
    &= \sum_{i=1}^t \bigl( f(\*x_i) - \*v_t^\top (\*x_i) \*f_t \bigr)^2 + \sigma^2 \| \*v_t^\top (\cdot) \*f_t \|_{\cH_k} \\
    &\leq \sum_{i=1}^t \bigl( f(\*x_i) - f (\*x_i) \bigr)^2 + \sigma^2 \| f \|_{\cH_k} && \left(\because f \in \cH_k \right) \\
    &= \sigma^2 \| f \|_{\cH_k}.
\end{align*}
Hence, we obtain $\| \*v_t^\top (\cdot) \*f_t \|_{\cH_k} \leq \| f \|_{\cH_k} \leq B$.
By combining Lemma~\ref{lem:RKHS_lipschitz}, $\*v_t^\top (\cdot) \*f_t$ is $B L_k$ Lipschitz continuous.

For the second term $\*v_t^\top (\cdot) \*\epsilon_t$, we leverage the confidence bounds of kernel ridge estimator~\citep[Theorem~3.11 in][]{abbasi2013online}.
Let $g: \cX \times \{0, 1\} \rightarrow \RR$ as $g(\*x, 0) = g(\*x, 1) = 0$ and fix $j \in [d]$.
Then, the zero function $g$ belongs to the RKHS with any kernel function $\overline{k}$.
Thus, we design the following kernel function $\overline{k}$:
\begin{align*}
    \overline{k}\left( (\*x, 0), (\*z, 0)  \right) &= k(\*x, \*z), \\
    \overline{k}\left( (\*x, 1), (\*z, 1)  \right) &= \frac{\partial^2 k(\*x, \*z)}{\partial x_j \partial z_j}, \\
    \overline{k}\left( (\*x, 0), (\*z, 1)  \right) &= \frac{\partial k(\*x, \*z)}{ \partial z_j}, 
\end{align*}
for all $\*x, \*z \in \cX$.
Note that since the kernel function, $k$ has partial derivatives due to Assumption~\ref{assump:frequentist_continuous}, the derivative of the kernel and the kernel itself are the kernels again as discussed in, e.g., Sec.~9.4 in \citep{Rasmussen2005-Gaussian} and Sec.~2.2 in \citep{adler1981geometry}.
Thus, we can interpret $\overline{\*v}_t^\top (\cdot) \*\epsilon_t$ as the kernel ridge estimator for $g(\*x, 1)$, where $\overline{\*v}_t (\*x) = \left( \frac{\partial \*k_t^\top(\*x)}{\partial x_j} (\*K_t + \sigma^2 \*I_t)^{-1}\right)^\top$.
In addition, $\| g \|_{\cH_{\overline{k}}} = 0$.
Therefore, from Theorem~3.11 in \citep{abbasi2013online} and $|g(\*x, 1) - \overline{\*v}_t^\top (\*x) \*\epsilon_t| = |\overline{\*v}_t^\top (\*x) \*\epsilon_t|$, we obtain
\begin{align*}
    \Pr \left( |\overline{\*v}_t^\top (\*x) \*\epsilon_t| \leq \overline{\sigma}_t(\*x) \frac{R}{\sigma} \sqrt{ 2 \gamma_t + 2 \log \left( 1 / \delta \right)}, \forall \*x \in \cX \right) \geq 1 - \delta,
\end{align*}
where $\delta \in (0, 1)$ and $\overline{\sigma}_t(\*x) = \overline{k}\left( (\*x, 1), (\*x, 1) \right) - \frac{\partial \*k_t^\top(\*x)}{\partial x_j} (\*K_t + \sigma^2 \*I_t)^{-1} \frac{\partial \*k_t(\*x)}{\partial x_j}$ is the posterior variance that corresponds to this kernel ridge estimation.
Note that since the kernel matrix $\*K_t$ is defined by $ k(\*x, \*z)$, the MIG is the usual one defined by $\cX$ and $t$.
In addition, due to the monotonic decreasing property of the posterior variance, $\overline{\sigma}_t(\*x) \leq L_k$, we obtain
\begin{align*}
    \Pr \left( |\overline{\*v}_t^\top (\*x) \*\epsilon_t| \leq \frac{L_k R}{\sigma} \sqrt{ 2 \gamma_t + 2 \log \left( 1 / \delta \right)}, \forall \*x \in \cX \right) \geq 1 - \delta,
\end{align*}
and thus,
\begin{align*}
    \Pr \left( \sup_{\*x \in \cX} |\overline{\*v}_t^\top (\*x) \*\epsilon_t| \leq \frac{L_k R}{\sigma} \sqrt{ 2 \gamma_t + 2 \log \left( 1 / \delta \right)} \right) \geq 1 - \delta.
\end{align*}
Consequently, by using the union bound for all $j \in [d]$, we derive
\begin{align*}
    \Pr \left( \sup_{j \in [d]} \sup_{\*x \in \cX} |\overline{\*v}_t^\top (\*x) \*\epsilon_t| \leq \frac{L_k R}{\sigma} \sqrt{ 2 \gamma_t + 2 \log \left( d / \delta \right)} \right) \geq 1 - \delta,
\end{align*}
which shows that $\*v_t^\top (\*x) \*\epsilon_t$ is $\frac{L_k R}{\sigma} \sqrt{ 2 \gamma_t + 2 \log \left( d / \delta \right)}$ Lipschitz continuous.

Combining the Lipschitz constants of $\*v_t^\top (\cdot) \*f_t$ and $\*v_t^\top (\cdot) \*\epsilon_t$, we can obtain the result.

%%%%%%%%%%%%%%%%%%%%%%%%%%%%%%%%%%%%%%%%%%%%%%%%%%%%%%%%%%%%%%%%%%%%%%%%%%%
\subsection{Proof of Lemma~\ref{lem:bayesian_lipschitz_posterior_mean}}
\label{sec:proof_bayesian_lipschitz_posterior_mean}

First, we fix $(\*x_i)_{i \in [t]}$ without loss of generality since
\begin{align*}
    \Pr \left( \sup_{\*x \in \cX} \left| \frac{\partial \mu_t(\*u)}{\partial u_j} \Big|_{\*u = \*x} \right| > L \right)
    &= \EE_{(\*x_i)_{i \in [t]}} \left[ \Pr \left( \sup_{\*x \in \cX} \left| \frac{\partial \mu_t(\*u)}{\partial u_j} \Big|_{\*u = \*x} \right| > L \bigg| (\*x_i)_{i \in [t]} \right) \right].
\end{align*}
That is, the upper bound of the conditional probability given any $(\*x_i)_{i \in [t]}$ directly suggests the upper bound of the target probability on the left-hand side.
Note that from the assumption $(\*x_i)_{i \in [t]}$ is independent of $(\epsilon_i)_{i \in [t]}$ and $f$, the observations $\*y_t$ follows Gaussian distribution even if $(\*x_i)_{i \in [t]}$ is fixed.

We leverage Slepian's inequality shown as Proposition~A.2.6 in \citep{van1996weak}:
\begin{lemma}[Slepian, Fernique, Marcus, and Shepp]
    Let $X$ and $Y$ be separable, mean-zero Gaussian processes indexed by a common index set $T$ such that
    \begin{align*}
        \EE[(X_s - X_t)^2] \leq \EE[(Y_s - Y_t)^2],
    \end{align*}
    for all $s, t \in T$. Then,
    \begin{align*}
        \Pr\left( \sup_{t \in T} X_t \geq \lambda \right) \leq \Pr\left( \sup_{t \in T} Y_t \geq \lambda \right),
    \end{align*}
    for all $\lambda > 0$.
    \label{lem:slepian}
\end{lemma}
The separability~\citep[Definition~5.22 in][]{Handel-HDP} holds commonly.
As discussed in  Remark~5.23 in \citep{Handel-HDP}, for example, if the sample path is almost surely continuous, then the separability holds.
Furthermore, the sample path defined by the commonly used kernel functions, such as linear, SE, and Mat\'ern-$\nu$ kernels with $\nu \geq 1$, is continuous almost surely~\citep{dacosta2024samplep}.
In addition, if the kernel function is continuous, the posterior mean function is also continuous, almost surely.

First, we provide the proof of the result regarding $\mu_t(\*x)$.
Since $\*y_t \mid (\*x_i)_{i \in [t]} \sim \cN(\*0, \*K_t + \sigma^2\*I_t)$, we can see that $\mu_t \mid (\*x_i)_{i \in [t]} \sim \cG \cP \bigl(0, k_{\mu_t}(\*x, \*z) \bigr)$, where $k_{\mu_t}(\*x, \*z) = \*k_{t}(\*x) ^\top \bigl(\*K + \sigma^2 \*I_{t} \bigr)^{-1} \*k_{t}(\*z)$.
Furthermore, it is known that if the kernel has mixed partial derivative $\frac{\partial^2 k(\*x, \*z)}{ \partial x_j \partial z_j}$, $f$ and its derivative $\partial f(\*x) / \partial x_j$ jointly follow GPs~\citep{Rasmussen2005-Gaussian,adler1981geometry}.
Specifically, the derivative is distributed as
\begin{align*}
    f^{(j)} \coloneqq \frac{\partial f(\*x)}{\partial x_j} &\sim \cG \cP\left(0, \tilde{k}(\*x, \*z) \coloneqq \frac{\partial^2 k(\*u, \*v)}{\partial u_j \partial v_j } \bigg|_{\*u=\*x, \*v=\*z} \right),
\end{align*}
for all $j \in [d]$.
Note that since the prior mean of $f$ is zero, the prior mean of $f^{(j)}$ is also zero.
As with $f$, the derivative of $\mu_t$ is distributed as 
\begin{align*}
    \mu_t^{(j)} \coloneqq \frac{\partial \mu_t(\*x)}{\partial x_j} \mid (\*x_i)_{i \in [t]} &\sim \cG \cP\left(0, \tilde{k}_{\mu_t}(\*x, \*z) \coloneqq \frac{\partial^2 k_{\mu_t}(\*u, \*v)}{\partial u_j \partial v_j } \bigg|_{\*u=\*x, \*v=\*z} \right),
\end{align*}
for all $j \in [d]$.
In addition, the covariance is given as
\begin{align*}
    {\rm Cov}\left( f(\*x), f^{(j)}(\*z) \right)
    &= \frac{\partial k(\*x, \*u)}{\partial u_j } \bigg|_{\*u=\*z}.
\end{align*}

Then, we see that the posterior variance of the derivative can be obtained in the same way as the usual GP calculation, as follows:
\begin{align*}
    {\rm Var}\left( f^{(j)}(\*x) \mid \cD_{t} \right) &= \tilde{k}(\*x, \*x) - \tilde{k}_{\mu_t}(\*x, \*x), \\
    {\rm Cov}\left( f^{(j)}(\*x), f^{(j)}(\*z) \mid \cD_{t} \right) &= \tilde{k}(\*x, \*z) - \tilde{k}_{\mu_t}(\*x, \*z).
\end{align*}
On the other hand, we can obtain that
\begin{align*}
    \EE\left[ \bigl( f^{(j)}(\*x) - f^{(j)}(\*z) \bigr)^2 \right] &= \tilde{k}(\*x, \*x) + \tilde{k}(\*z, \*z) - 2\tilde{k}(\*x, \*z), \\
    \EE\left[ \bigl( \mu_t^{(j)}(\*x) - \mu_t^{(j)}(\*z) \bigr)^2 \mid (\*x_i)_{i \in [t]} \right] &= \tilde{k}_{\mu_t}(\*x, \*x) + \tilde{k}_{\mu_t}(\*z, \*z) - 2\tilde{k}_{\mu_t}(\*x, \*z),
\end{align*}
for all $\*x, \*z \in \cX$.
Then, we obtain
\begin{align*}
    &\EE\left[ \bigl( f^{(j)}(\*x) - f^{(j)}(\*z) \bigr)^2 \right] - \EE\left[ \bigl( \mu_t^{(j)}(\*x) - \mu_t^{(j)}(\*z) \bigr)^2 \mid (\*x_i)_{i \in [t]}\right] \\
    &= \tilde{k}(\*x, \*x) + \tilde{k}(\*z, \*z) - 2\tilde{k}(\*x, \*z) - \left( \tilde{k}_{\mu_t}(\*x, \*x) + \tilde{k}_{\mu_t}(\*z, \*z) - 2\tilde{k}_{\mu_t}(\*x, \*z) \right) \\
    &= {\rm Var}\left( f^{(j)}(\*x) \mid \cD_{t} \right) + {\rm Var}\left( f^{(j)}(\*z) \mid \cD_{t} \right) - 2 {\rm Cov}\left( f^{(j)}(\*x), f^{(j)}(\*z) \mid \cD_t \right)
    \geq 0.
\end{align*}
Consequently, by applying Lemma~\ref{lem:slepian}, we obtain
\begin{align*}
    \Pr\left( \sup_{\*x \in \cX} \mu^{(j)}_t(\*x) \geq \lambda \biggm\vert (\*x_i)_{i \in [t]} \right) \leq \Pr\left( \sup_{\*x \in \cX} f^{(j)}_t(\*x) \geq \lambda \right).
\end{align*}
Since $\mu^{(j)}_t(\*x)$ and $f^{(j)}_t(\*x)$ follow centered GPs, we obtain
\begin{align*}
    \Pr\left( \sup_{\*x \in \cX} \left| \mu^{(j)}_t(\*x) \right| \geq \lambda \biggm\vert (\*x_i)_{i \in [t]} \right) 
    \leq 2 \Pr\left( \sup_{\*x \in \cX} \mu^{(j)}_t(\*x) \geq \lambda \biggm\vert (\*x_i)_{i \in [t]} \right)
    \leq 2 \Pr\left( \sup_{\*x \in \cX} \left| f^{(j)}_t(\*x) \right| \geq \lambda \right).
\end{align*}
Hence, from Lemma~\ref{assump:Bayesian_continuous}, we obtain the desired result.

We can obtain the result regarding $f^{(j)}(\*x) - \mu_t^{(j)}(\*x)$ in almost the same proof.
We can see that
\begin{align*}
    f^{(j)}(\*x) - \mu_t^{(j)}(\*x) \sim \cG \cP(0, \tilde{k}(\*x, \*z) - \tilde{k}_{\mu_t}(\*x, \*z)).
\end{align*}
Then, 
\begin{align*}
    &\EE\left[ \bigl( f^{(j)}(\*x) - f^{(j)}(\*z) \bigr)^2 \right] - \EE\left[ \left( f^{(j)}(\*x) - \mu_t^{(j)}(\*x) - \bigl( f^{(j)}(\*x) - \mu_t^{(j)}(\*x) \bigr) \right)^2 \biggm\vert (\*x_i)_{i \in [t]} \right] \\
    &= \tilde{k}_{\mu_t}(\*x, \*x) + \tilde{k}_{\mu_t}(\*z, \*z) - 2\tilde{k}_{\mu_t}(\*x, \*z)
    \geq 0.
\end{align*}
The remaining proof is the same as the case of $\mu_t^{(j)}$.

%%%%%%%%%%%%%%%%%%%%%%%%%%%%%%%%%%%%%%%%%%%%%%%%%%%%%%%%%%%%%%%%%%%%%%%%%%%
\subsection{Proof of Lemma~\ref{lem:UB_error_frequentist_continuous}}
\label{sec:proof_UB_error_frequentist_continuous}

As with the existing studies~\citep[e.g., ][]{Srinivas2010-Gaussian}, we consider the discretization of input space.
Let $\overline{\cX} \subset \cX$ be a finite set with each dimension equally divided into $\lceil \tau d r \rceil$, where $\tau > 0$.
Therefore, $|\overline{\cX}| = \lceil \tau d r \rceil^d$ and $\sup_{\*x \in \cX} \| \*x - [\*x] \|_1 \leq \frac{1}{\tau}$, where $[\*x]$ is the nearest input in $\overline{\cX}$, that is, $[\*x] = \argmin_{\tilde{\*x} \in \overline{\cX}} \|\tilde{\*x} - \*x \|_1$.
Note that we leverage $\overline{\cX}$ purely for the analysis, and $\overline{\cX}$ is not related to the algorithm.

From Assumption~\ref{assump:frequentist_continuous} and Lemma~\ref{lem:RKHS_lipschitz}, we see that $f$ is $B L_k$ Lipschitz continuous.
Furthermore, from Lemma~\ref{lem:lipschitz_posterior_mean}, $\mu_t$ is $L_k \left( B + \frac{R}{\sigma} \sqrt{ 2 \gamma_t + 2 \log \left( \frac{d}{\delta} \right)} \right)$ Lipschitz continuous with probability at least $1 - \delta$.
Combining the above, we see that $f - \mu_t$ is $L_k \left( 2B + \frac{R}{\sigma} \sqrt{ 2 \gamma_t + 2 \log \left( \frac{d}{\delta} \right)} \right)$ Lipschitz continuous with probability at least $1 - \delta$.

% From Lemma~\ref{lem:lipschitz_posterior_mean} and Assumprion~\ref{assump:frequentist}, we can see that
% \begin{align*}
%     \| f(\*x^*) - \mu_T(\*x^*) \|_{\cH_k} 
%     \leq \| f(\*x^*) \|_{\cH_k} + \| \mu_T(\*x^*) \|_{\cH_k}
%     \leq 2B + \frac{R}{\sigma} \sqrt{ 2T \log \left( \frac{2T}{\delta} \right)}.
% \end{align*}
% %
% Therefore, from Assumption~\ref{assump:frequentist_continuous} and Lemma~\ref{lem:RKHS_lipschitz}, $f(\*x^*) - \mu_T(\*x^*)$ is Lipschitz continuous with high probability.

From the above arguments, by combining Lemma~\ref{lem:bound_vakili} and the union bound, the following events hold simultaneously with probability at least $1 - \delta$:
\begin{enumerate}
    \item $f(\*x) - \mu_T(\*x)$ is $L_{\rm res}(T)$ Lipschitz continuous, where $L_{\rm res}(T) = L_k \left( 2B + \frac{R}{\sigma} \sqrt{ 2\gamma_T + 2 \log \left( \frac{2d}{\delta} \right)} \right)$.
    \item The confidence bounds on $\overline{\cX}$ hold; that is, 
    \begin{align*}
        \forall \*x \in \overline{\cX}, |f(\*x) - \mu_T(\*x)| \leq \beta_{\delta, \tau}^{1/2} \sigma_T (\*x),
    \end{align*}
    where $\beta_{\delta, \tau} = \left( B + \frac{R}{\sigma} \sqrt{ 2 d \log \left( \tau d r + 1 \right) + 2 \log \left( \frac{4}{\delta} \right)} \right)^2$.
\end{enumerate}

Then, we can obtain the upper bound as follows:
\begin{align*}
    E_T &= \max_{p \in \cP} \EE_{p(\*x^*)} \left[ \left( f(\*x^*) - \mu_T(\*x^*) \right)^2 \right] \\
    &\leq \max_{p \in \cP} \EE_{p(\*x^*)} \left[ \left( f([\*x^*]) - \mu_T([\*x^*]) + L_{\rm res}(T) \| \*x^* - [\*x^*] \|_1 \right)^2  \right] 
    && \left(\because \text{The above event~1} \right) \\
    &\leq \max_{p \in \cP} \EE_{p(\*x^*)} \left[ \left( f([\*x^*]) - \mu_T([\*x^*]) + \frac{L_{\rm res}(T)}{\tau} \right)^2  \right] 
    && \left(\because \text{The definition of $\overline{\cX}$} \right) \\
    &\leq \max_{p \in \cP} \EE_{p(\*x^*)} \left[ \left( \beta_{\delta, \tau}^{1/2} \sigma_T([\*x^*]) + \frac{L_{\rm res}(T)}{\tau} \right)^2  \right] 
    && \left(\because \text{The above event~2} \right) \\
    &\leq \max_{p \in \cP} \EE_{p(\*x^*)} \left[ \left( \beta_{\delta, \tau}^{1/2} \sigma_T(\*x^*) + \frac{\beta_{\delta, \tau}^{1/2} L_{\sigma} + L_{\rm res}(T)}{\tau} \right)^2  \right] 
    && \left(\because \text{Lemma~\ref{lem:Lipschitz_posterior_var}} \right) \\
    &\leq 2 \beta_{\delta, \tau} \max_{p \in \cP} \EE_{p(\*x^*)} \left[  \sigma_T^2(\*x^*) \right] + 2 \left( \frac{\beta_{\delta, \tau}^{1/2} L_{\sigma} + L_{\rm res}(T)}{\tau} \right)^2.
    && \left(\because (a + b)^2 / 2 \leq a^2 + b^2 \right)
\end{align*}
If we set $\tau = T$, noting that $L_{\rm res}(T) = \cO(\sqrt{\gamma_T})$ and $\beta_{\delta, \tau}^{1/2} = \cO(\log(T / \delta))$, we obtain the following:
\begin{align*}
    E_T 
    &\leq 2 \beta_{\delta, T} \max_{p \in \cP} \EE_{p(\*x^*)} \left[  \sigma_T^2(\*x^*) \right] 
    + \cO \left( \frac{\max\{\gamma_T, \log(T / \delta)\}}{T^2} \right).
\end{align*}
Although by setting $\tau = \Omega(T)$, we can make the second term small arbitrarily, $\beta_{\delta, T} = \Theta\left( \log( T / \delta) \right)$ and $\EE_{p(\*x^*)} \left[  \sigma_T^2(\*x^*) \right] = \Omega(1 / T)$~\citep[Lemma~4.2 in][]{takeno2024-posterior}.
Therefore, since the first term is $\Omega \left( \frac{\log( T / \delta)}{T} \right)$ and $\frac{\max\{\gamma_T, \log(T / \delta)\}}{T^2} = \cO\left( \frac{\log( 1 / \delta)}{T} \right)$ if $\gamma_T$ is sublinear, we do not set $\tau$ more large value for simplicity.

%%%%%%%%%%%%%%%%%%%%%%%%%%%%%%%%%%%%%%%%%%%%%%%%%%%%%%%%%%%%%%%%%%%%%%%%%%%
\subsection{Proof of Lemma~\ref{lem:UB_error_bayesian_continuous}}
\label{sec:proof_UB_error_bayesian_continuous}

As with the existing studies~\citep[e.g., ][]{Srinivas2010-Gaussian}, we consider the discretization of input space.
Let $\overline{\cX} \subset \cX$ be a finite set with each dimension equally divided into $\lceil \tau d r \rceil$, where $\tau > 0$.
Therefore, $|\overline{\cX}| = \lceil \tau d r \rceil^d$ and $\sup_{\*x \in \cX} \| \*x - [\*x] \|_1 \leq \frac{1}{\tau}$, where $[\*x]$ is the nearest input in $\overline{\cX}$, that is, $[\*x] = \argmin_{\tilde{\*x} \in \overline{\cX}} \|\tilde{\*x} - \*x \|_1$.
Note that we leverage $\overline{\cX}$ purely for the analysis, and $\overline{\cX}$ is not related to the algorithm.

In addition, from Lemma~\ref{lem:bayesian_lipschitz_posterior_mean}, the following inequality holds with probability at least $1 - \delta$:
\begin{align*}
    \sup_{j \in d} \sup_{\*x \in \cX} 
    \left| \frac{\partial r_t(\*u)}{\partial u_j} \Big|_{\*u = \*x} \right|
    \leq b \sqrt{\log (2ad / \delta)},
\end{align*}
which implies that $L_{\rm res}$, the Lipschitz constant of $r_t(\*x) = f(\*x) - \mu_T(\*x)$, can be bounded from above.

Then, by combining the above argument, Lemma~\ref{lem:bound_srinivas}, and the union bound, the following events hold simultaneously with probability at least $1 - \delta$:
\begin{enumerate}
    \item $f(\*x) - \mu_T(\*x)$ is $L_{\rm res}$ Lipschitz continuous, where $L_{\rm res} =b \sqrt{\log (4ad / \delta)}$.
    \item The confidence bounds on $\overline{\cX}$ hold; that is, 
    \begin{align*}
        \forall \*x \in \overline{\cX}, |f(\*x) - \mu_T(\*x)| \leq \beta_{\delta, \tau}^{1/2} \sigma_T (\*x),
    \end{align*}
    where $\beta_{\delta, \tau} = 2d \log (\tau d r + 1) + 2 \log (2 / \delta)$.
\end{enumerate}

Hence, we can obtain the upper bound as follows:
\begin{align*}
    E_T &= \max_{p \in \cP} \EE_{p(\*x^*)} \left[ \left( f(\*x^*) - \mu_T(\*x^*) \right)^2 \right] \\
    &\leq \max_{p \in \cP} \EE_{p(\*x^*)} \left[ \left( f([\*x^*]) - \mu_T([\*x^*]) + L_{\rm res} \| \*x^* - [\*x^*] \|_1 \right)^2  \right] 
    && \left(\because \text{The above event~1} \right) \\
    &\leq \max_{p \in \cP} \EE_{p(\*x^*)} \left[ \left( f([\*x^*]) - \mu_T([\*x^*]) + \frac{L_{\rm res}}{\tau} \right)^2  \right] 
    && \left(\because \text{The definition of $\overline{\cX}$} \right) \\
    &\leq \max_{p \in \cP} \EE_{p(\*x^*)} \left[ \left( \beta_{\delta, \tau}^{1/2} \sigma_T([\*x^*]) + \frac{L_{\rm res}}{\tau} \right)^2  \right] 
    && \left(\because \text{The above event~2} \right) \\
    &\leq \max_{p \in \cP} \EE_{p(\*x^*)} \left[ \left( \beta_{\delta, \tau}^{1/2} \sigma_T(\*x^*) + \frac{\beta_{\delta, \tau}^{1/2} L_{\sigma} + L_{\rm res}}{\tau} \right)^2  \right] 
    && \left(\because \text{Lemma~\ref{lem:Lipschitz_posterior_var}} \right) \\
    &\leq 2 \beta_{\delta, \tau} \max_{p \in \cP} \EE_{p(\*x^*)} \left[  \sigma_T^2(\*x^*) \right] + 2 \left( \frac{\beta_{\delta, \tau}^{1/2} L_{\sigma} + L_{\rm res}}{\tau} \right)^2.
    && \left(\because (a + b)^2 / 2 \leq a^2 + b^2 \right)
\end{align*}
Then, by setting $\tau = T$, we can see that
\begin{align*}
    E_T 
    &\leq 2 \beta_{\delta, T} \max_{p \in \cP} \EE_{p(\*x^*)} \left[  \sigma_T^2(\*x^*) \right] 
    + \cO\left( \frac{\log(T / \delta)}{T^2} \right).
\end{align*}

Although by setting $\tau = \Omega(T)$, we can make the second term small arbitrarily, we do not do so since the first term is dominant compared with $\cO\left( \frac{\log(T / \delta)}{T^2} \right)$ term.

%%%%%%%%%%%%%%%%%%%%%%%%%%%%%%%%%%%%%%%%%%%%%%%%%%%%%%%%%%%%%%%%%%%%%%%%%%%
\subsection{Proof of Lemma~\ref{lem:UB_absolute_error}}
\label{sec:UB_absolute_error_proof}

Since the maximum $\tilde{p}_T(\*x^{*}) = \argmax_{p \in \cP} \EE_{p(\*x^{*})} \left[ |f(\*x^{*}) - \mu_T(\*x^{*})| \right] $ exists, we obtain
\begin{align*}
    \max_{p \in \cP} \EE_{p(\*x^{*})} \left[ |f(\*x^{*}) - \mu_T(\*x^{*})| \right] 
    &= \EE_{\tilde{p}_T(\*x^{*})} \left[ |f(\*x^{*}) - \mu_T(\*x^{*})| \right], \\
    &\leq \sqrt{ \EE_{\tilde{p}_T(\*x^{*})} \left[ (f(\*x^{*}) - \mu_T(\*x^{*}))^2 \right]}, \\
    &\leq \sqrt{ \max_{p \in \cP} \EE_{p(\*x^{*})} \left[ (f(\*x^{*}) - \mu_T(\*x^{*}))^2 \right]},
    % &\coloneqq \max_{p \in \cP} \EE_{p(\*x^{*})} \left[ ( f(\*x^{*}) - \mu_T(\*x^{*}) )^2 \right]
\end{align*}
where we used Jensen's inequality.

%%%%%%%%%%%%%%%%%%%%%%%%%%%%%%%%%%%%%%%%%%%%%%%%%%%%%%%%%%%%%%%%%%%%%%%%%%%
\subsection{Proof of Lemma~\ref{lem:UB_entropy}}
\label{sec:UB_entropy_proof}

Since the maximum $\overline{p}_T(\*x^{*}) = \argmax_{p \in \cP} \EE_{p(\*x^{*})} \left[ H[f(\*x) \mid \cD_t] \right] $ exists, we obtain
\begin{align*}
    \max_{p \in \cP} \EE_{p(\*x)}[ H[f(\*x) \mid \cD_t] ]
    &= \frac{1}{2} \EE_{\overline{p}_T(\*x^{*})}[ \log(2\pi e \sigma_{T}^2(\*x)) ] \\
    &\leq \frac{1}{2} \log(2\pi e \EE_{\overline{p}_T(\*x^{*})}[  \sigma_{T}^2(\*x) ]) && (\because \text{Jensen's inequality}) \\
    &\leq \frac{1}{2} \log \left( 2\pi e \max_{p \in \cP} \EE_{p (\*x)}[ \sigma_{T}^2(\*x) ] \right).
    % &\leq \frac{1}{2} \log\left( \frac{2\pi e}{T} \left( 2 C_1 \gamma_T + 4\log\frac{1}{\delta} + 8 \log(4) + 1 \right) \right) && \left(\because  \text{Theorem~\ref{theo:error_convergence_RS} and \ref{theo:error_convergence_greedy}}\right).
\end{align*}

%%%%%%%%%%%%%%%%%%%%%%%%%%%%%%%%%%%%%%%%%%%%%%%%%%%%%%%%%%%%%%%%%%%%%%%%%%%
\section{Proofs for Section~\ref{sec:proposed}}
\label{sec:proposed_proof}

\subsection{Proof of Theorem~\ref{theo:error_convergence_RS}}

From the definition, $\sigma_t^2(\*x)$ is monotonically decreasing along with $\cD_{t-1} \subset \cD_t$.
Therefore, for all $t \leq T$ and $\*x_1, \dots, \*x_T$, 
\begin{align*}
    \max_{p \in \cP} \EE_{p(\*x^*)}[\sigma_{T}^2(\*x^*) \mid \*x_1, \dots, \*x_T]
    \leq 
    \max_{p \in \cP} \EE_{p(\*x^*)}[\sigma_{t}^2(\*x^*) \mid \*x_1, \dots, \*x_t].
\end{align*}
Note that $\*x_1, \dots, \*x_T$ are random variables due to the randomness of the algorithm.
Hence, we obtain
\begin{align*}
    \max_{p \in \cP} \EE_{p(\*x^*)}[\sigma_{T}^2(\*x^*) \mid \*x_1, \dots, \*x_T]
    &\leq 
    \frac{1}{T} \sum_{t=1}^T
    \max_{p \in \cP} \EE_{p(\*x^*)}[\sigma_{t-1}^2(\*x^*) \mid \*x_1, \dots, \*x_{t-1}] \\
    &\leq
    \frac{1}{T} \sum_{t=1}^T
    \EE_{p_t(\*x_t)}[\sigma_{t-1}^2(\*x_t) \mid \*x_1, \dots, \*x_{t-1}]. && (\because \text{Definition of $p_t$})
\end{align*}
Then, we apply the following lemma \citep[Lemma 3 in][]{kirschner2018-information}:
\begin{lemma}
    Let $Y_t$ be any non-negative stochastic process adapted to a filtration $\{ \cF_t \}$, and define $m_t = \EE[Y_t \mid \cF_{t-1}]$.
    Further assume that $Y_t \leq b_t$ for a fixed, non-decreasing sequence $(b_t)_{t \geq 1}$.
    Then, if $b_T \geq 1$, with probability at least $1 - \delta$ for any $T \geq 1$, it holds that,
    \begin{align*}
        \sum_{t=1}^T m_t \leq 2 \sum_{t=1}^T Y_t + 4 b_T \log \frac{1}{\delta} + 8b_T \log(4 b_T) + 1.
    \end{align*}
\end{lemma}
The random variable $\EE_{p_t(\*x)}[\sigma_{t}^2(\*x) \mid \*x_1, \dots, \*x_t]$ satisfies the condition of this lemma by setting $b_t = 1$ for all $t \in [T]$.
Therefore, with probability at least $1 - \delta$,
\begin{align*}
    \max_{p \in \cP} \EE_{p(\*x^*)}[\sigma_{T}^2(\*x^*) \mid \*x_1, \dots, \*x_T]
    &\leq 
    \frac{1}{T} \left( 
    2 \sum_{t=1}^T \sigma_{t-1}^2(\*x_t) + 4\log\frac{1}{\delta} + 8 \log(4) + 1
    \right) \\
    &\leq 
    \frac{1}{T} \left( 
    2 C_1 \gamma_T + 4\log\frac{1}{\delta} + 8 \log(4) + 1
    \right).
\end{align*}
Here, we use $\sum_{t=1}^T \sigma_{t-1}^2(\*x_t) \leq C_1 \gamma_T$ \citep[Lemma~5.2 in][]{Srinivas2010-Gaussian}.

%%%%%%%%%%%%%%%%%%%%%%%%%%%%%%%%%%%%%%%%%%%%%%%%%%%%%%%%%%%%%%%%%
\subsection{Proof of Theorem~\ref{theo:error_convergence_greedy}}

From the definition, $\sigma_t^2(\*x)$ is monotonically decreasing along with $\cD_{t-1} \subset \cD_t$.
Therefore, for all $t \leq T$ and $\*x_1, \dots, \*x_T$, 
\begin{align*}
    \max_{p \in \cP} \EE_{p(\*x^*)}[\sigma_{T}^2(\*x^*) \mid \*x_1, \dots, \*x_T]
    \leq 
    \max_{p \in \cP} \EE_{p(\*x^*)}[\sigma_{t}^2(\*x^*) \mid \*x_1, \dots, \*x_t].
\end{align*}
Hence, we obtain
\begin{align*}
    \max_{p \in \cP} \EE_{p(\*x^*)}[\sigma_{T}^2(\*x^*) \mid \*x_1, \dots, \*x_T]
    &\leq 
    \frac{1}{T} \sum_{t=1}^T
    \max_{p \in \cP} \EE_{p(\*x^*)}[\sigma_{t-1}^2(\*x^*) \mid \*x_1, \dots, \*x_{t-1}] \\
    &\leq
    \frac{1}{T} \sum_{t=1}^T \sigma_{t-1}^2(\*x_t) && (\because \text{Definition of $\cX_t$}) \\
    &\leq \frac{C_1 \gamma_T}{T}. && (\because \text{Lemma~5.2 in \citep{Srinivas2010-Gaussian}})
\end{align*}

%%%%%%%%%%%%%%%%%%%%%%%%%%%%%%%%%%%%%%%%%%%%%%%%%%%%%%%%%%%%%%%%%%%%%%%%%%%
\section{Other Experimental Settings and Results}
\label{sec:exp_settings}

\subsection{Results for Variance}
\label{sec:exp_variance}

\begin{figure*}[t]
    \centering
    \includegraphics[width=\linewidth]{fig/experiments/legend.pdf}\\
    \includegraphics[width=0.24\linewidth]{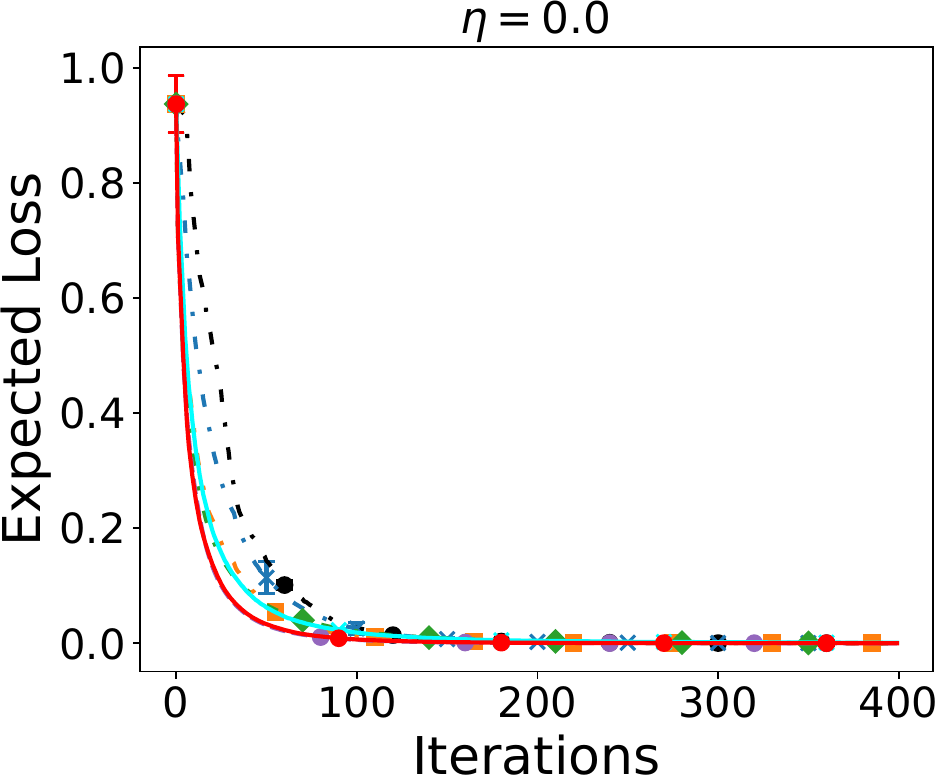}
    \includegraphics[width=0.24\linewidth]{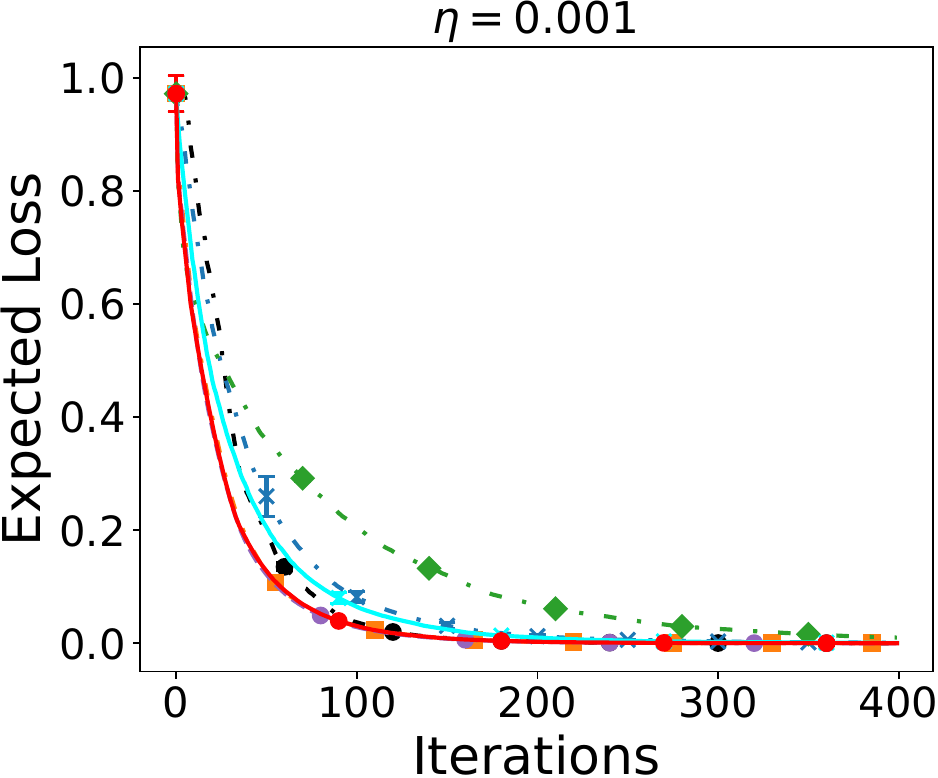}
    \includegraphics[width=0.24\linewidth]{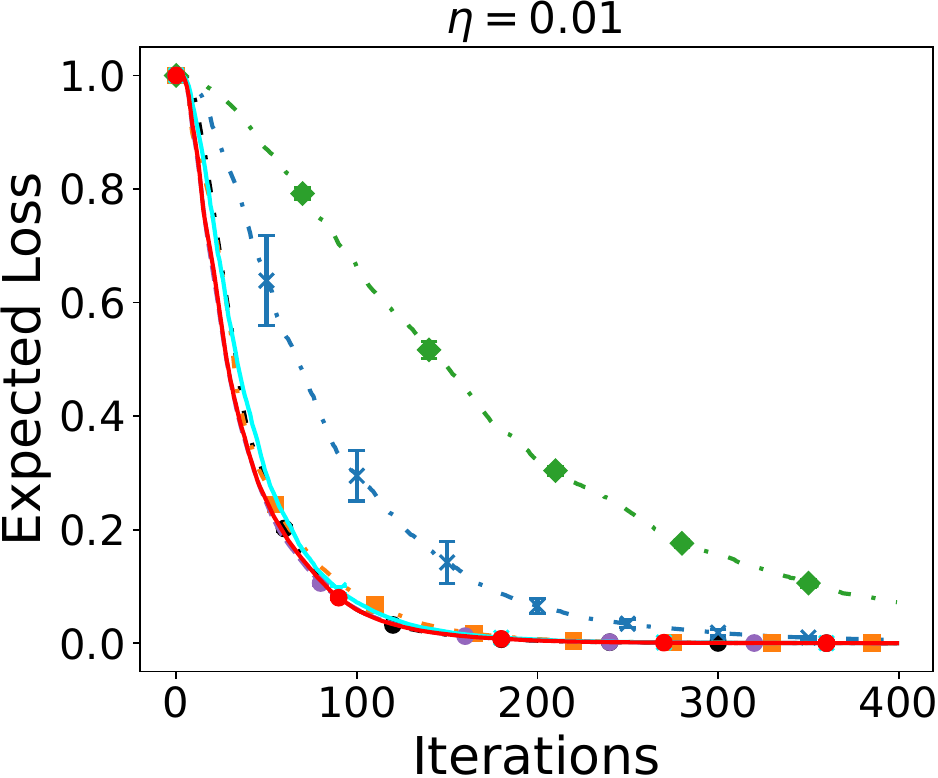}
    \includegraphics[width=0.24\linewidth]{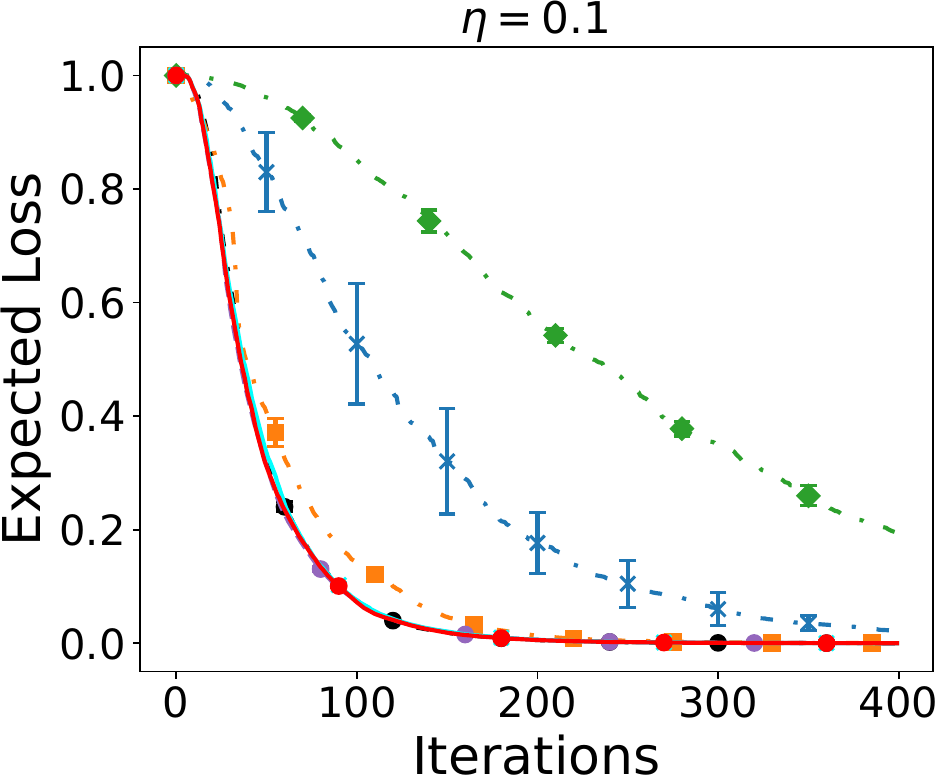}\\
    \includegraphics[width=0.24\linewidth]{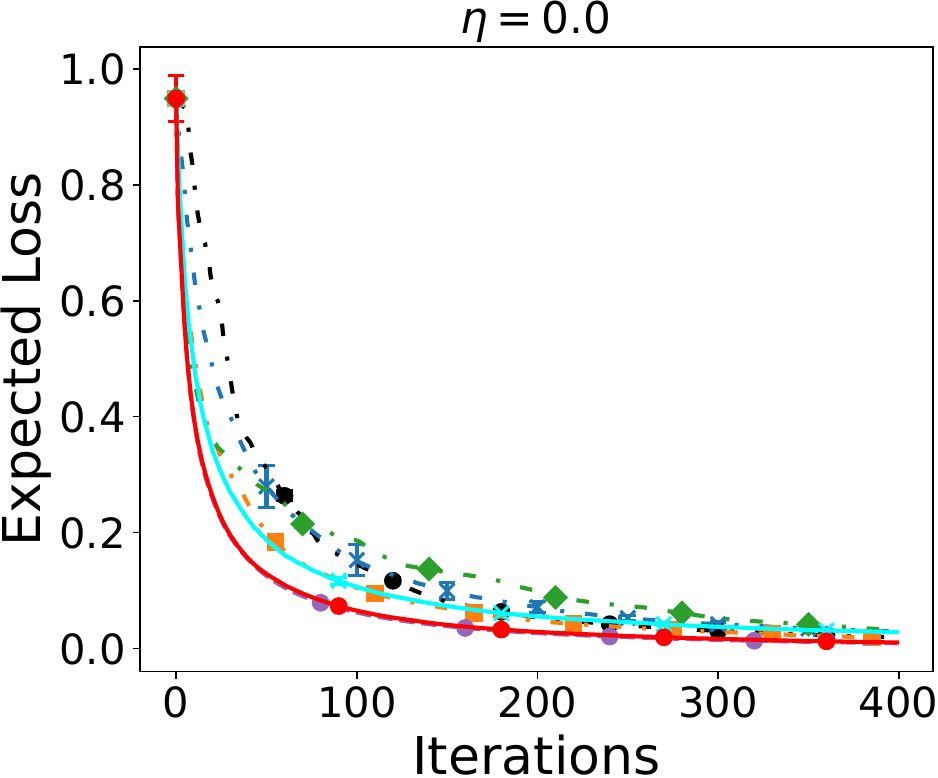}
    \includegraphics[width=0.24\linewidth]{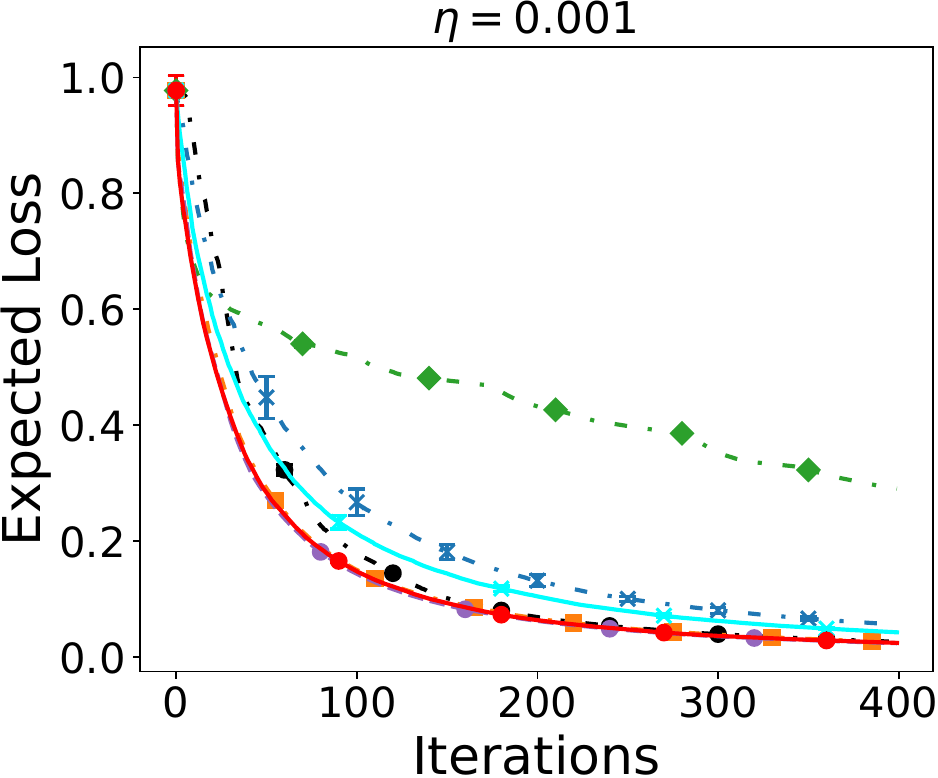}
    \includegraphics[width=0.24\linewidth]{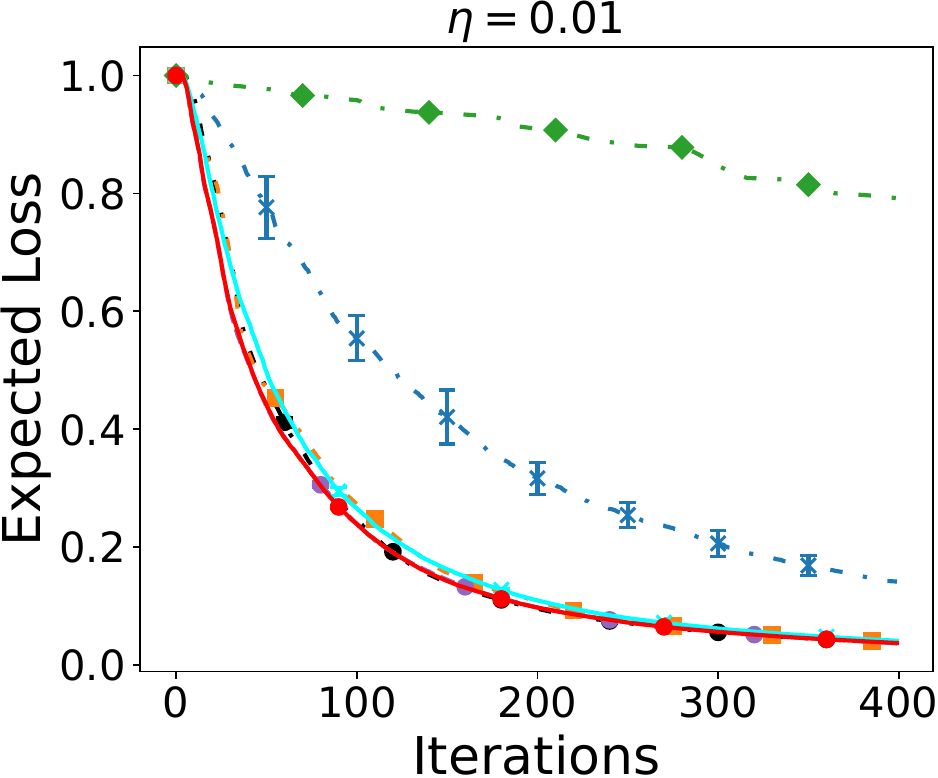}
    \includegraphics[width=0.24\linewidth]{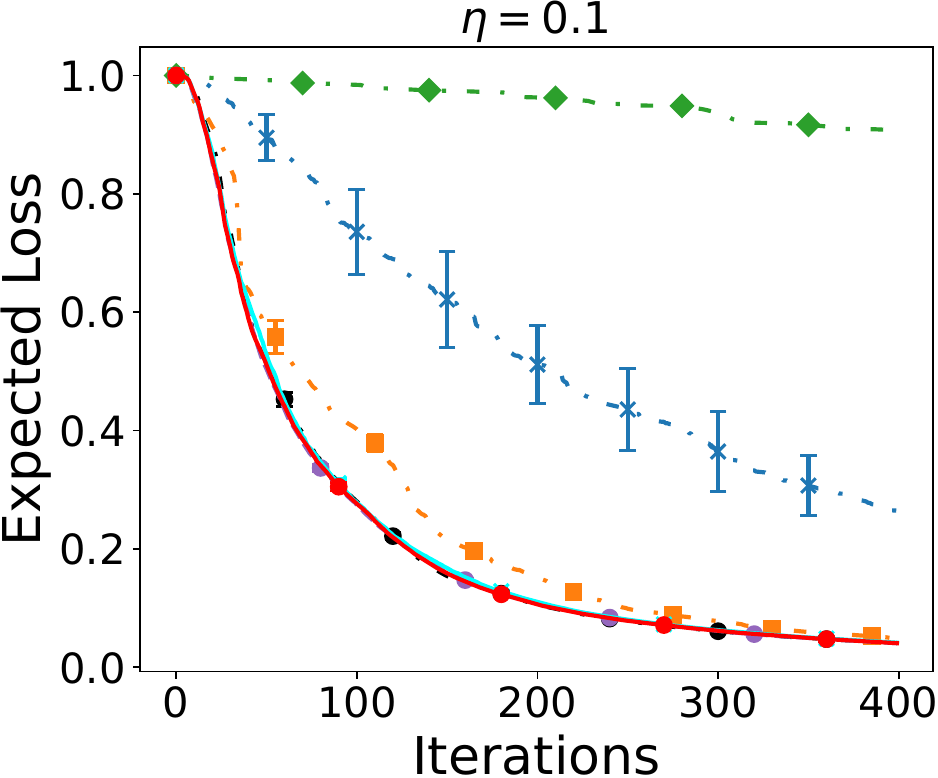}
    \caption{
        Result of $\max_{p \in \cP} \EE_{p(\*x^*)} \left[ \sigma_t^2 (\*x^*) \right]$ in the synthetic data experiments with $\eta=0, 0.001, 0.01, 0.1$.
        The horizontal and vertical axes show the number of iterations and $\max_{p \in \cP} \EE_{p(\*x^*)} \left[ \sigma_t^2 (\*x^*) \right]$, respectively.
        The error bar shows mean and standard errors for 20 random trials regarding the random initial point (and the algorithm's randomness).
        The top and bottom rows represent the results of the GPR model with SE and Mat\'ern kernels, respectively.
        %
        % 最悪期待損失を用いて評価を行った場合の人工データ実験結果.
        % %
        % 横軸は反復数, 縦軸は最悪期待損失を示す.
        % %
        % ランダムに初期点を変更したことによる20回試行の実験結果の平均と標準誤差を示す.
        % %
        % 上段はRBFカーネル, 下段はMaternカーネルをGP回帰のカーネルとして用いた場合の結果を示す.
        % %
        % 参照分布は平均が${\* 0}$, 共分散行列が$0.2*{\*I}_3$の三次元正規分布とする.
        % %
        % ここで, ${\*I}_3$は三次元の単位行列である.
        % %
        % また, 参照分布からの最大距離$\epsilon=0, 0.001, 0.01, 0.1$とする.
        %
    }
    \label{fig:syn_loss_results}
\end{figure*}

Figure~\ref{fig:syn_loss_results} shows the result of $\max_{p \in \cP} \EE_{p(\*x^*)} \left[ \sigma_t^2 (\*x^*) \right]$, which suggests that the proposed methods effectively minimize $\max_{p \in \cP} \EE_{p(\*x^*)} \left[ \sigma_t^2 (\*x^*) \right]$.

\subsection{Details on Implementation of EPIG}
\label{sec:EPIG}

EPIG is defined as follows~\citep{bickford2023-prediction}:
\begin{align*}
    \*x_t = \argmax_{\*x \in \cX} \EE_{p(\*x^*)}\left[ H[ y_{\*x^*} \mid \cD_{t-1}, \*x] - H[ y_{\*x^*} \mid \cD_{t-1}] \right].
\end{align*}
Although \citet{bickford2023-prediction} have discussed the efficient computation for EPIG, in the regression problem, the EPIG can be computed analytically except for the expectation over $p(\*x^*)$ as follows:
\begin{align*}
    \*x_t = \argmax_{\*x \in \cX} \EE_{p(\*x^*)}\left[ \log \left( \frac{\sigma_{t-1}^2(\*x, \*x^*) }{ \sqrt{\sigma_{t-1}^2(\*x) + \sigma^2} \sqrt{\sigma_{t-1}^2(\*x^*) + \sigma^2} } \right) \right],
\end{align*}
where $\sigma_{t-1}^2(\*x, \*x^*)$ is the posterior covariance between $\*x$ and $\*x^*$.
Since we focus on the discrete input domain in the experiments, the expectation over $p(\*x^*)$ can also be computed analytically.
We used the above equation for the implementation.

\subsection{Details on Real-World Datasets}
\label{sec:datasets_detail}

The King County house sales dataset is a dataset used to predict house prices in King County by 7-dimensional features, such as the area and the number of rooms. 
This dataset has been used for testing the regression~\citep{park2020additive}, and a similar dataset has also been used for the AL studies~\citep{park2020robust}.
Although this dataset includes 20000 data points, we used a random sample of 1000 data points for simplicity.

% King County House Sales Datasetは回帰問題の対象として使用されてきたデータセット\citep{park2020additive}であり, 既存の能動学習の研究においても類似したデータセットが用いられている\citep{park2020robust}.
%
% データセットには20000個のデータが含まれているが, 本実験では扱いやすさのため1000個をランダムに選択して使用する.
%
% データの入力次元は7次元である.

Red wine quality dataset~\citep{wine_quality_186} is a dataset used to predict the quality of wines from the wine ingredients expressed by 11-dimensional features.
This dataset includes 1600 data points and has been used for the regression problem~\citep{cortez2009using}.

% Red Wine Quality Dataset\citep{wine_quality_186}はワインの成分からワインの品質を予測することを目的とするデータセットである.
% %
% Red Wine Quality Datasetは回帰問題の対象として使用されてきたデータセットである\citep{cortez2009using}. 
% %
% データセットには1600個のデータが含まれており, データの入力次元は11次元である.

Auto MPG dataset~\citep{auto_mpg_9} is a dataset used to predict automobile fuel efficiency from 6-dimensional features, such as the weight of the automobile and engine horsepower.
Auto MPG dataset has been used for the AL research~\citep {park2020robust}.
This dataset includes 399 data points.

% Auto MPG Dataset\citep{auto_mpg_9}は自動車の重量やエンジンの馬力から自動車の燃費を予測することを目的としたデータセットである.
% %
% Auto MPG Datasetは既存の能動学習の研究で使用されているデータセットである\citep{park2020robust}.
% %
% データセットには399個のデータセットが含まれており, データの入力次元は6次元である.

%%%%%%%%%%%%%%%%%%%%%%%%%%%%%%%%%%%%%%%%%%%%%%%%%%%%%%%%%%%%%%%%%%%%%%%%%%%%%%%%%%%%%%%%%%%%%%
\end{document}